%% file: ldSVM.tex
\documentclass[letterpaper]{article}
\usepackage[margin=1in,dvips]{geometry}
\usepackage{graphicx,psfrag,amsmath,amsthm,amssymb}
\usepackage{natbib}
\usepackage{url,color}
\usepackage{algorithm,algorithmic}

\input{MACPdef}
\input{MACPdef-ams}

\graphicspath{{grf/}}

\bibpunct[, ]{(}{)}{;}{a}{,}{,} % Should be the natbib default but it isn't!

\title{The role of dimensionality reduction in linear classification}
\author{
  Weiran Wang \hspace{5ex} Miguel \'A. Carreira-Perpi\~n\'an \\
  Electrical Engineering and Computer Science, University of California, Merced \\
  {\url{http://eecs.ucmerced.edu}}
}
\date{May 25, 2014}

\AtBeginDvi{}

\begin{document}

\maketitle

\begin{abstract}

  Dimensionality reduction (DR) is often used as a preprocessing step in classification, but usually one first fixes the DR mapping, possibly using label information, and then learns a classifier (a filter approach). Best performance would be obtained by optimizing the classification error jointly over DR mapping and classifier (a wrapper approach), but this is a difficult nonconvex problem, particularly with nonlinear DR. Using the method of auxiliary coordinates, we give a simple, efficient algorithm to train a combination of nonlinear DR and a classifier, and apply it to a RBF mapping with a linear SVM. This alternates steps where we train the RBF mapping and a linear SVM as usual regression and classification, respectively, with a closed-form step that coordinates both. The resulting nonlinear low-dimensional classifier achieves classification errors competitive with the state-of-the-art but is fast at training and testing, and allows the user to trade off runtime for classification accuracy easily. We then study the role of nonlinear DR in linear classification, and the interplay between the DR mapping, the number of latent dimensions and the number of classes. When trained jointly, the DR mapping takes an extreme role in eliminating variation: it tends to collapse classes in latent space, erasing all manifold structure, and lay out class centroids so they are linearly separable with maximum margin.

\end{abstract}

\section{Introduction}

Dimensionality reduction (DR) is a common preprocessing step for classification and other tasks. Learning a classifier on low-dimensional inputs is fast (though learning the DR itself may be costly). More importantly, DR can help learn a better classifier, particularly when the data does have a low-dimensional structure, and with small datasets, where DR has a regularizing effect that can help avoid overfitting. The reason is that DR can remove two types of ``noise'' from the input: (1) independent random noise, which is uncorrelated with the input and the label, and mostly perturbs points away from the data manifold. Simply running PCA, or other unsupervised DR algorithm, with an adequate number of components, can achieve this to some extent. (2) Unwanted degrees of freedom, which are possibly nonlinear, along which the input changes but the label does not. This more radical form of denoising requires the DR to be informed by the labels, of course, and is commonly called supervised DR.

Call \F\ the DR mapping, which takes an input $\x \in \bbR^D$ and projects it to $L < D$ dimensions, and \g\ the classifier, which applies to the low-dimensional vector $\F(\x)$ and produces a label $y$, so that the overall classifier is $\g \circ \F$. The great majority of supervised DR algorithms are ``filter'' approaches \citep{KohaviJohn98a,GuyonElisseef03a}, where one first learns \F\ from the training set of pairs $(\x_n,y_n)$, fixes it, and then train \g\ on the pairs $(\F(\x_n),y_n)$, using a standard classification algorithm as if the inputs were $\F(\x)$. An example of supervised DR is linear discriminant analysis (LDA), which learns the best linear DR \F\ in the sense of minimal intra-class scatter and maximal across-class scatter. \emph{The key in filter approaches is the design of a proxy objective function over \F\ that leads to learning a good overall classifier $\g \circ \F$}. Although the particulars differ among existing supervised DR methods (e.g.\ \citealp{Belhum_97a,GloberRoweis06a}), usually they encourage \F\ to separate inputs or manifolds having different labels from each other. While this makes intuitive sense, and filter methods (and even PCA) can often do a reasonable job, it is clear that the best \g\ and \F\ are not obtained by optimizing a proxy function over \F\ and then having \g\ minimize the classification error (our real objective), but by \emph{jointly} minimizing the latter over \g\ and \F\ (the ``wrapper'' approach). However, filter approaches (particularly using a linear DR) are far more popular than wrapper ones. With filters, the classifier is learned as usual once \F\ has been learned. With wrappers, learning \F\ and \g\ involve a considerably more difficult, nonconvex optimization, particularly with nonlinear DR, having many more parameters that are coupled with each other. At this point, an important question arises: \emph{what is the real role of (nonlinear) DR in classification, and how does it depend on the choice of mapping \F\ and of latent dimensionality $L$?} Guided by this overall goal, the contributions of this paper are as follows. (1) We propose a simple, efficient, scalable and generic way of jointly optimizing the classifier's loss over $(\F,\g)$, and in this paper we apply it to the case where \F\ is a RBF network and \g\ a linear SVM. (2) Armed with this algorithm, we study the role of nonlinear DR in the classifier's performance and the latent space representation, and find lessons that apply to filter design. (3) We obtain a nonlinear low-dimensional SVM classifier that achieves state-of-the-art performance while being fast at test time.

A shorter version of this work appears in a conference paper \citep{WangCarreir14a}.

\section{Joint optimization of mapping and classifier using auxiliary coordinates}
\label{s:mac}

We describe first the approach for binary classification and focus on the case where \g\ is a linear SVM. We give the multiclass case in section~\ref{s:multiclass}. Given a training set of $N$ input patterns $\x_n \in \bbR^D$ and corresponding labels $y_n \in \{-1,+1\}$, $n=1,\dots,N$, we want to learn a nonlinear low-dimensional classifier $\g \circ \F$ that optimizes the following objective:
\begin{gather}
  \label{e:svm-nested}
  \min_{\F,\g,\bxi}{ \lambda R(\F) + \frac{1}{2} \norm{\w}^2 + C \sum^N_{n=1}{\xi_n} } \\
  \text{ s.t.\ } \left\{ y_n (\w^T \F(\x_n) + b) \ge 1-\xi_n,\ \xi_n \ge 0 \right\}^N_{n=1}.\notag
\end{gather}
This is the usual linear SVM objective of finding a separating hyperplane with maximum margin, but with inputs given by \F, which has a regularization term $\lambda R(\F)$, where $\g = \{\w,b\}$ are the weights and bias of the linear SVM \g, and with a slack variable $\xi_n$ per point $n$, where $C$ is a penalty parameter for the slackness. The difficulty is that the constraints are heavily nonconvex because of the nonlinear mapping \F. However, the problem can be significantly simplified if we use the recently introduced \emph{method of auxiliary coordinates (MAC)} \citep{CarreirWang12a,CarreirWang14a} for nested systems. The idea is to introduce auxiliary variables that break nested functional dependences $\g(\F(\cdot))$ into simpler shallow mappings $\g(\z)$ and $\F(\cdot)$. In our case, we introduce one auxiliary vector per input pattern and define the following constrained problem, which can be proven \citep{CarreirWang12a} to be equivalent to~\eqref{e:svm-nested}:
\begin{gather}
  \label{e:svm-constrained}
  \min_{\F,\g,\bxi,\Z}{ \lambda R(\F) + \frac{1}{2} \norm{\w}^2 + C \sum^N_{n=1}{\xi_n} } \\
  \text{ s.t.\ } \left\{ y_n (\w^T \z_n + b) \ge 1-\xi_n,\ \xi_n \ge 0,\ \z_n = \F(\x_n) \right\}^N_{n=1}.\notag
\end{gather}
This seems like a notational trick, but now we solve this with a quadratic-penalty method \citep{NocedalWright06a}. We optimize the following problem for fixed penalty parameter $\mu>0$ and drive $\mu \rightarrow \infty$:
\begin{gather}
  \min_{\F,\g,\bxi,\Z}{ \lambda R(\F) + \frac{1}{2} \norm{\w}^2 + C \sum^N_{n=1}{\xi_n} + \frac{\mu}{2} \sum_{n=1}^N{\norm{\z_n-\F(\x_n)}^2} }\notag \\
  \label{e:svm-qp}
  \text{ s.t.\ } \left\{y_n (\w^T \z_n + b) \ge 1-\xi_n,\ \xi_n \ge 0 \right\}^N_{n=1}.
\end{gather}
This defines a continuous path $(\F^*(\mu),\g^*(\mu),\Z^*(\mu))$ which, under mild assumptions, converges to a minimum of the constrained problem~\eqref{e:svm-constrained}, and thus to a minimum of the original problem~\eqref{e:svm-nested} \citep{CarreirWang12a}. Although problem~\eqref{e:svm-qp} has more parameters, all the terms are simple and partially separable. The auxiliary vector $\z_n$ can be interpreted as a target in latent space for the mapping \F, but these targets are themselves coordinated with the classifier. Using alternating optimization of~\eqref{e:svm-qp} over $(\F,\g,\Z)$ results in very simple, convex steps. The $(\F,\g)$ step is a usual RBF regression and linear SVM classification done independently from each other reusing existing, well-developed algorithms. The \Z-step has a closed-form solution for each $\z_n$ separately. We describe the steps next. The complete alternating optimization procedure is given in Algorithm~\ref{a:drsvm}.

\renewcommand{\algorithmicrequire}{\textbf{Input:}}
\begin{algorithm}[t]
\caption{Nonlinear low-dimensional SVM using the method of auxiliary coordinates (MAC).}
\label{a:drsvm}
\begin{algorithmic}[1]
\REQUIRE dataset of points \X\ and their labels \Y
\STATE initialize \Z
\STATE fit \g\ to $(\Z,\Y)$ by SVM training
\STATE fit \F\ to $(\X,\Z)$ by regression
\FOR{$\mu = \mu_0 < \mu_1 < \dots < \mu_{\infty}$}
\REPEAT
\STATE optimize over \Z\ (algorithm~\ref{a:drsvm-optZ})
\STATE fit \g\ to $(\Z,\Y)$ by SVM training
\STATE fit \F\ to $(\X,\Z)$ by regression
\UNTIL stop
\ENDFOR
\RETURN \F\ and \g\
\end{algorithmic}
\end{algorithm}

\subsection{The \g\ step}
\label{s:g-step}

For fixed $(\F,\Z)$, optimizing over $\w,b,\bxi$ is just training an ordinary linear SVM with low-dimensional inputs \Z. Much work exists on fast, scalable SVM training. We use LIBSVM~\citep{ChangLin11a}. With $K$ classes, each SVM can be trained independently of the others.

\subsection{The \F\ step}
\label{s:F-step}

For fixed $(\g,\Z)$, optimizing over \F\ is just a regularized regression with inputs \X\ and low-dimensional outputs \Z. So far the approach is generic over \F, which could be a deep net or a Gaussian process, for example, and we would use its corresponding training method within this step. However, we now focus on a special case which results in a particularly efficient step over \F, and which includes linear DR as a particular case. We use radial basis function (RBF) networks, which are universal function approximators, $\F(\x) = \W\bPhi(\x)$, with $M \ll N$ Gaussian RBFs $\phi_m(\x) = \smash{\exp}{(-\frac{1}{2}\smash{\norm{(\x-\c_m)/\sigma}}^2)}$, and $R(\F) = \smash{\smash{\norm{\W}}^2}$ is a quadratic regularizer on the weights. As commonly done in practice \citep{Bishop06a}, we determine the centers $\c_m$ by $k$-means on \X, and then the weights \W\ have a unique solution given by a linear system. The total cost is $\calO(M(L+D))$ in memory and $\calO(NM(M+D))$ in training time, mainly driven by setting up the linear system for \W\ (involving the Gram matrix $\phi_m(\x_n)$); solving it exactly is a negligible $\calO(\smash{M^3})$ since $M \ll N$ in practice, which can be reduced to $\calO(\smash{M^2})$ by using warm-starts or caching its Cholesky factor. Note we only need to run $k$-means and factorize the linear system once and for all, since its input \X\ does not change. Thus, the \F-step simply involves a linear system for \W.

\subsection{The \Z\ step}
\label{s:Z-step}

For fixed $(\F,\g)$, eq.~\eqref{e:svm-qp} decouples on each $n$, so instead of one large problem on $NL$ parameters, we have $N$ independent small problems each on $L$ parameters, of the form (omitting subindex $n$):
\begin{gather}
  \label{e:svm-optZ}
  \min_{\z,\xi}{ \norm{\z-\F(\x)}^2 + c \xi } \\
  \text{ s.t.\ } y(\w^T \z + b) \ge 1-\xi,\ \xi \ge 0 \qquad \z \in \bbR^L \notag
\end{gather}
where we have also included the slack variable, since we find this speeds up the alternating optimization. This is a convex quadratic program, whose closed-form solution is $\z = \F(\x) + \gamma y \w$, where $\gamma$ is a scalar which takes one of three possible values, and costs $\calO(L)$.

\begin{figure}[t]
  \centering
  \psfrag{0}[t][]{$0$}
  \psfrag{z}[t][]{$z$}
  \psfrag{xi}[][][1][270]{$\xi$}
  \begin{tabular}{@{}c@{\hspace{0\linewidth}}c@{\hspace{0\linewidth}}c@{}}
    \psfrag{l1=0,  l2=1}[][]{$\lambda_1=0$}
    \psfrag{zopt=fx}[][]{$\z_{\text{opt}}=\F(\x)$\hspace{1.5em}}
    \includegraphics[width=0.35\linewidth]{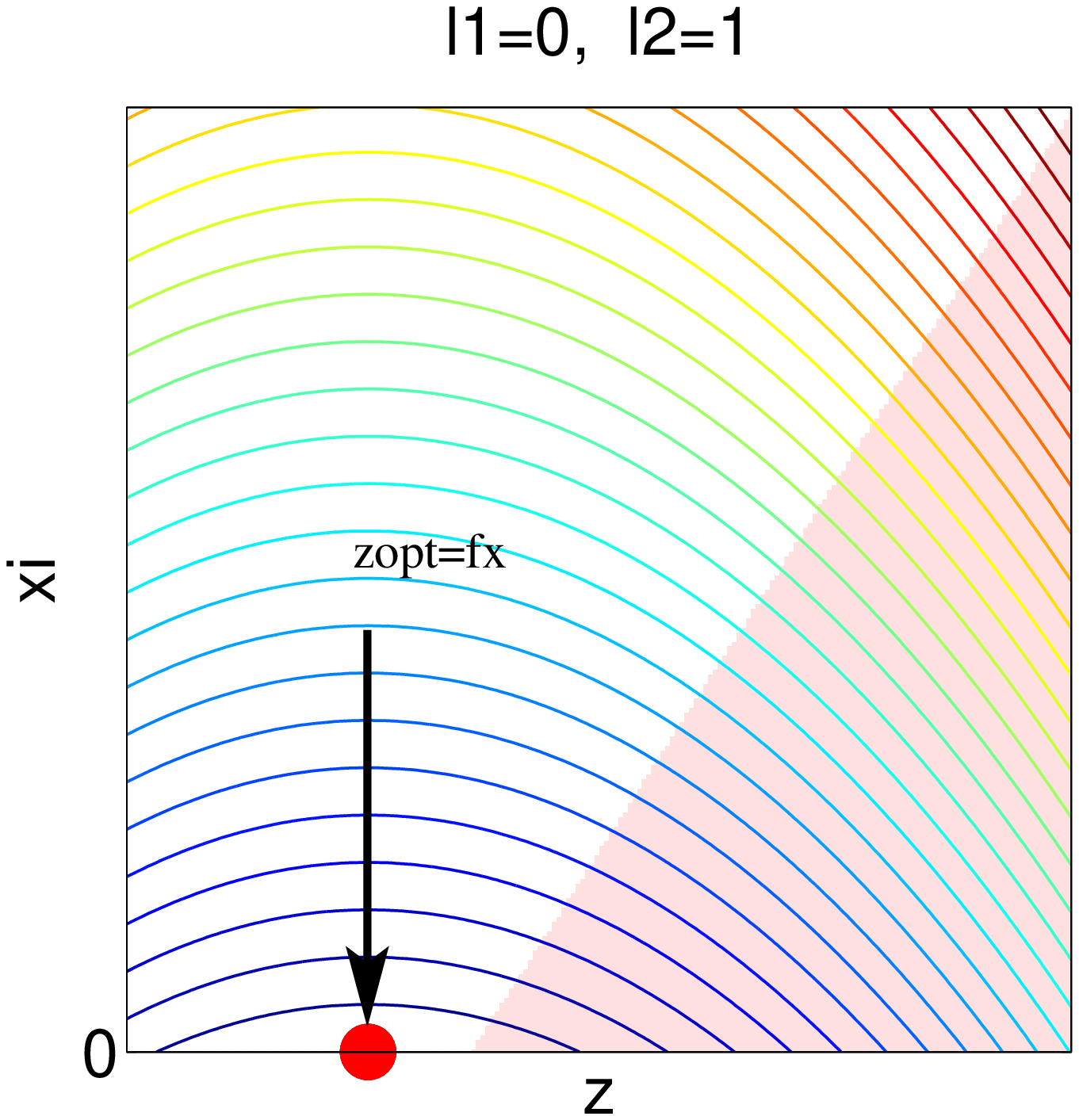} &
    \psfrag{l1=0.7549,  l2=0.2451}[][]{$0<\lambda_1<c$}
    \psfrag{zopt}[][]{ $\z_{\text{opt}}$}
    \psfrag{fx}[][]{ $\F(\x)$}
    \includegraphics[width=0.325\linewidth]{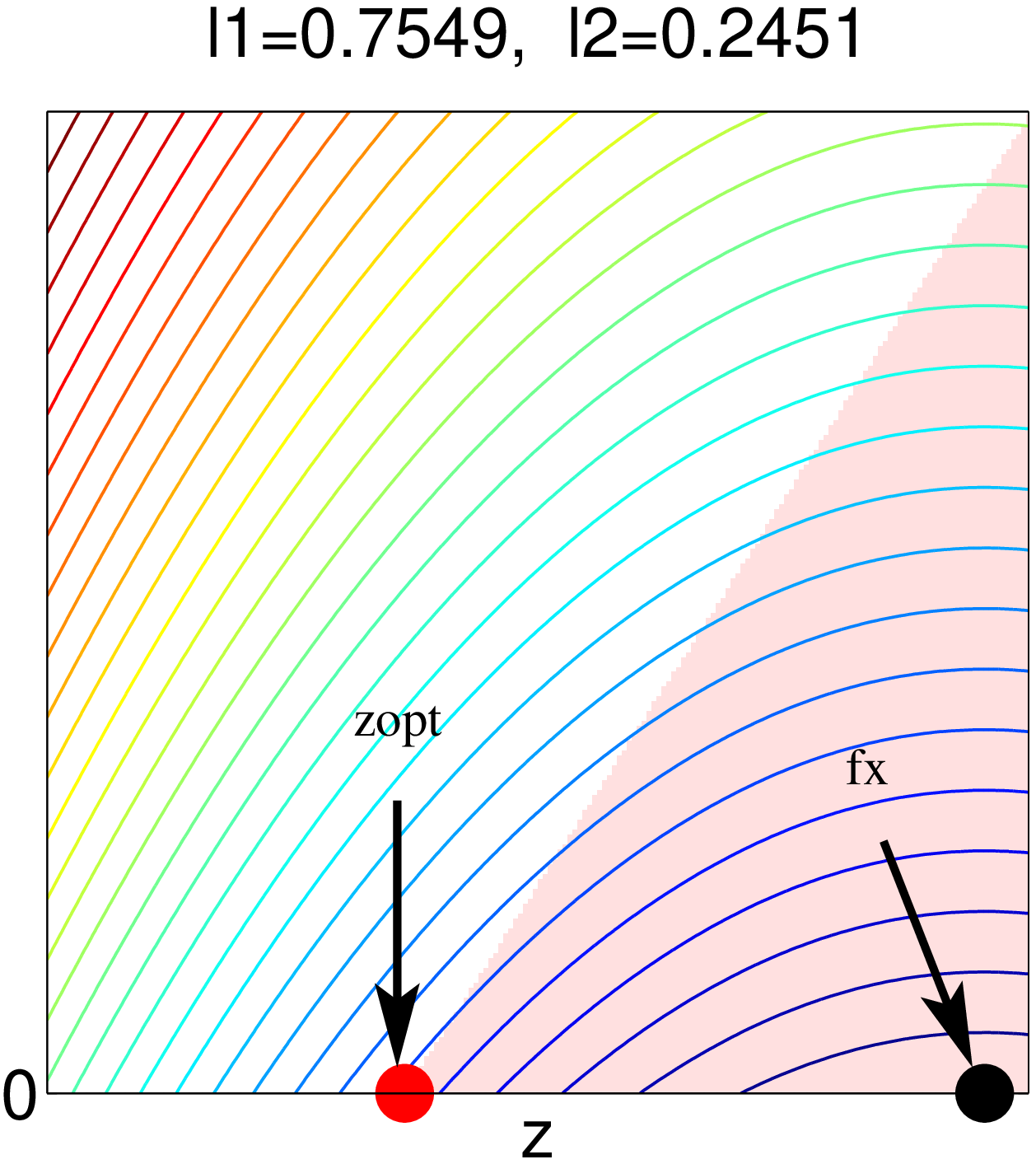} &
    \psfrag{l1=1,  l2=0}[][]{$\lambda_1=c$}
    \psfrag{zopt}[][]{ $\z_{\text{opt}}$}
    \psfrag{fx}[][]{ $\F(\x)$}
    \includegraphics[width=0.325\linewidth]{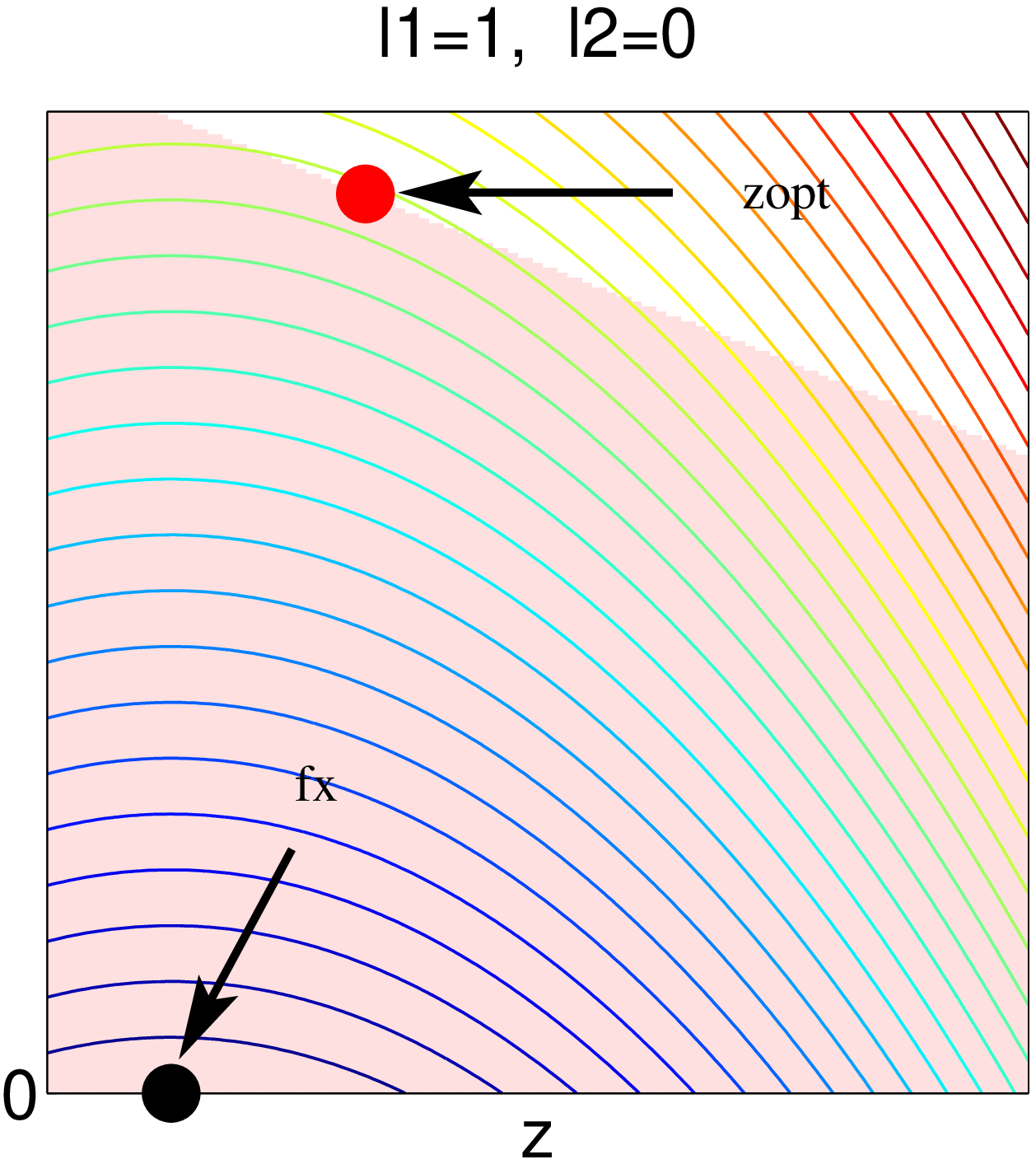}
  \end{tabular}
  \caption{Three possible solutions of the quadratic program in the \Z\ step. The contours of the objective function are shown in color (increasing from blue to red) and the area not shaded denotes the feasible set. In each plot, the origin is in the bottom left corner, the black dot denotes $\F(\x)$ and the red dot denotes the final solution \z.}
  \label{f:drsvm-Z}
\end{figure}

This can be seen using the KKT theorem. The Lagrangian of \eqref{e:svm-optZ} is
\begin{equation}
  \label{e:Lagrangian}
  \calL(z,\xi,\lambda_1,\lambda_2) = \norm{\z-\F(\x)}^2+c\xi-\lambda_1 (y(\w^T \z + b)+\xi-1)-\lambda_2 \xi
\end{equation}
where $\lambda_1$ and $\lambda_2$ are Lagrange multipliers for the two inequality constraints. Its KKT system is
\begin{align*}
  \nabla \calL_{\z} = \0 \;\Longrightarrow\; \z = \F(\x)+\frac{\lambda_1}{2}y\w \\
  \nabla \calL_{\xi} = 0 \;\Longrightarrow\; \lambda_1 + \lambda_2 = c \\
  y(\w^T \z + b) \ge 1-\xi, \quad \xi \ge 0 \\
  \lambda_1 \ge 0, \quad \lambda_2 \ge 0 \\
  \lambda_1 (y(\w^T \z + b)+\xi-1) = 0, \quad \lambda_2 \xi = 0.
\end{align*}
From the first equation, the optimal \z\ lies on a line which passes through $\F(\x)$ and is parallel to the normal direction of \g. If $\lambda_1=0$, we have $\lambda_2=c$, $\xi=0$, and the KKT system reduces to $\z=\F(\x)$, $y(\w^T \z+b)\ge 1$. We have three cases.
\begin{description}
\item[Case 1] This means, if $y(\w^T \F(\x)+b)\ge 1$ is satisfied (i.e., $\F(\x)$ is classified correctly with a margin of $1$), we should simply set $\z=\F(\x)$ and get an objective function value of $0$ (the ideal case). This situation is demonstrated in fig.~\ref{f:drsvm-Z} (left plot). \\
  Otherwise, we must have $\lambda_1>0$. The KKT system reduces to (using the relation $\lambda_2=c-\lambda_1$):
  \begin{align*}
    \z=\F(\x) + \frac{\lambda_1}{2} y\w , \quad  y(\w^T \z+b)+\xi = 1 \\
    \xi \ge 0, \quad  0<\lambda_1\le c, \quad (c-\lambda_1)\xi=0 
  \end{align*}
  and from the first two equations we get
  \begin{align*}
    \xi = 1-y(\w^T \F(\x) + b)-\frac{\lambda_1}{2}\w^T \w \ge 0 \\
    \Longrightarrow \lambda_1\le \frac{2(1-y(\w^T \F(\x) + b))}{\w^T \w}.
  \end{align*}
  Substituting the above KKT conditions into the Lagrangian $\calL$, we can  express the dual of \eqref{e:svm-optZ} using only $\lambda_1$ as
  \begin{gather*}
    \max_{\lambda_1}{ -\frac{1}{4} (\w^T\w) \lambda_1^2 + (1-y(\w^T \F(\x) + b))\lambda_1 } \\
    \text{s.t.\ } \lambda_1\le \min \left(\frac{2(1-y(\w^T \F(\x) + b))}{\w^T \w},c \right).
  \end{gather*}
  The objective function is a concave parabola with maximum achieved at $\tilde{\lambda}_1 = \frac{2(1-y(\w^T \F(\x) + b))}{\w^T \w}$.
\item[Case 2] Thus if $\tilde{\lambda}_1 < c$, the optimum of \eqref{e:svm-optZ} is achieved by $\lambda_1=\tilde{\lambda}_1$, $\lambda_2=c-\tilde{\lambda}_1$, and $\z=\F(\x)+\frac{\tilde{\lambda}_1}{2} y \w$.
\item[Case 3] Otherwise the optimum of \eqref{e:svm-optZ} is achieved by $\lambda_1=c$, $\lambda_2=0$, and $\z=\F(\x)+\frac{c}{2} y \w$. 
\end{description}
By now, we have found the solutions to \eqref{e:svm-optZ} completely with a cost of $\calO(L)$. We give illustrations of the three possible cases in figure~\ref{f:drsvm-Z}. This procedure is summarized in Algorithm~\ref{a:drsvm-optZ}.

\begin{algorithm}[t]
\caption{\Z\ step optimization over $\{\z,\xi\}$.}
\label{a:drsvm-optZ}
\begin{algorithmic}[1]
\REQUIRE $\w$, $b$ and $c$ from \g, and $\F(\x)$ from \F
\STATE $m=y \, (\w^T \F(\x) + b)$
\IF{$m\ge1$}
\STATE $\z=\F(\x)$, $\xi=0$, $\lambda_1=0$
\ELSE 
\STATE $\lambda_1 = 2(1-m)/\w^T \w$
\IF{$\lambda_1<c$}
\STATE $\z=\F(\x)+\frac{\lambda_1}{2} y \w$
\ELSE 
\STATE $\z=\F(\x)+\frac{c}{2} y \w$
\ENDIF
\STATE $\xi=1-y(\w^T \z+b)$
\ENDIF
\RETURN \z\ and $\xi$
\end{algorithmic}
\end{algorithm}

\subsection{Formulation for the multiclass problem}
\label{s:multiclass}

Let $K$ be the number of classes. There are several ways to construct a classifier for $K$ classes. We use the one-vs-all scheme, which has been shown to perform as well as any other variants of multiclass SVM \citep{RifkinKlautau04a}, and which gives rise to a simpler \Z\ step. In the one-vs-all scheme we have $K$ binary SVMs, each of which is trained to classify whether a point belongs to some class or not. The decision function on a test point \x\ is $\arg\max{(\w^T_1\x,\dots,\w^T_K\x)}$, i.e., the final label is determined by the SVM with the largest decision value. The objective in \eqref{e:svm-nested} is replaced by the sum of the $K$ SVMs' objective functions. In the \Z-step, we solve for each point a quadratic program of the following form (omitting subindex $n$):
\begin{gather*}
  \label{e:svm-optZm}
  \min_{\z,\{\xi^k\}^K_{k=1}}{ \norm{\z-\F(\x)}^2 + \sum_{k=1}^K C^k \xi^k } \\
  \text{s.t.\ } y^k((\w^k)^T \z + b^k) \ge 1-\xi^k,\ \xi^k \ge 0, \qquad\text{for } k=1,\dots,K
\end{gather*}
where $C^k$ is the $k$th SVM's penalty parameter for the hinge loss and $\xi^k$ is the slack variable of the $k$th SVM (associated with the point in consideration). This quadratic program contains $L+K$ variables and $2K$ inequality constraints. Since the size of this problem is typically not large, we use an active set algorithm for solving it (as implemented in Matlab's Optimization Toolbox). For the binary case, there exists only one SVM and the problem has a closed-form solution, as shown before.

It is also possible to use the one-vs-one scheme. With $K$ classes, we have $\frac{K(K-1)}{2}$ binary SVMs, each trained on each pair of classes. The decision function on a test point is given by majority vote (i.e., the classifier that wins more times). This has a lower training time but a higher test time. Also, the one-vs-one scheme involves a little more bookkeeping in the \Z\ step than the one-vs-all scheme. To see this, take for example $K=10$ classes. The one-vs-all scheme uses 10 SVMs, and, in the \Z\ step, each data point is involved in all 10 SVMs, either as positive or negative example. The one-vs-one scheme uses 45 SVMs, and each point is involved in only 9 of them.

\subsection{Summary and practicalities}

Jointly optimizing the classification error over \g\ and \F\ becomes iterating simple convex subproblems: RBF regression, linear SVM and a closed-form update for \Z. Remarkably, we do not require any involved gradient computation, but simply reuse existing techniques for training \F\ and \g\ efficiently. Thus, although the problem~\eqref{e:svm-nested} is nonconvex and has multiple local optima, the algorithm is deterministic given its initialization. We run the algorithm from an initial \Z. We observe that, given reasonable hyperparameters, even random values work well (section~\ref{s:roleDR} shows how to construct a near-optimal initial \Z). In the first 1--2 iterations, the $\z_n$ quickly reorganize in latent space, and they only get refined afterwards so as to enlarge the margin and reduce the bias. We use a initial value of $\mu = 2$ for the quadratic penalty parameter and increase it times $1.5$ when the alternating scheme converges for fixed $\mu$. We use early stopping, by exiting the iteration when the error in a validation set does not change or goes up. This helps to improve generalization and is faster: we observe a few iterations suffice, because each step updates very large, decoupled blocks of variables, and we need not drive $\mu \rightarrow \infty$ or achieve convergence. The hyperparameters are the usual ones: $M$, $\sigma$, $\lambda$ for the RBF mapping \F, and $C$ for the SVM \g, and can be determined by cross-validation.

Our algorithm affords massive parallelization and is suitable for large scale learning. The \F-step (a regression problem) decouples over latent dimensions, the \g-step (one-versus-all SVM) decouples over classes, and the \Z-step decouples over training samples. Indeed, our experiments show linear speedups with multiple processors.

The form of our final classifier is $y = \g(\F(\x)) = \vv^T \bPhi(\x) + b = \smash{\sum^M_{m=1}{v_m \phi_m(\x)}} + b$ where $\vv = \W^T \w \in \bbR^M$, and the sign of $y$ gives the label. The number of parameters (weights and centers) is $M(D+L)+L = \calO(MD)$ and the runtime for a test point is $\calO(MD)$.

\section{Experiments}
\label{s:expts}

We explore three questions across a range of datasets: the role of dimension reduction on the classification error and the latent space representation; the performance of our classifier, compared to the state-of-the-art; and the training speed of our algorithm.

\subsection{The role of dimension reduction}
\label{s:roleDR}

\paragraph{The ideal nonlinear dimensionality reduction + linear classifier}

Given that the classifier \g\ is linear, consider the ideal case where \F\ is infinitely flexible and can represent any desirable mapping from $D$ to $L$ dimensions. Then, a perfect wrapper classifier $\g \circ \F$ can be achieved by having \F\ map all the inputs having the same label $k$ to a point $\bmu_k \in \bbR^L$, and then locating the $K$ class centroids $\bmu_1,\dots,\bmu_K$ in $\bbR^L$ such that they are linearly separable and have maximum margin. How this can be achieved depends on the dimension $L$. With $L=1$, only $K=2$ classes may be linearly separable. With $L=2$, all $K$ classes are linearly separable if placing the centroids on the vertices of a regular $K$-polygon; however, this leads to a small margin as $K$ grows. In $L>2$, this generalizes to placing the centroids maximally apart on a hypersphere. However, when $L \ge K-1$, the margin cannot be further improved, because the $K$ points span a space of $K-1$ dimensions, specifically a regular simplex, which provides linear separability and maximum margin. Is this ideal actually realized?

\begin{figure*}[t]
  \centering
  \begin{tabular}{@{}c@{\hspace{0\linewidth}}c@{\hspace{0\linewidth}}c@{\hspace{0\linewidth}}c@{\hspace{0\linewidth}}c@{\hspace{0\linewidth}}c@{\hspace{0\linewidth}}c@{\hspace{0\linewidth}}c@{}}
    \F\ linear & $M=4$ & $M=10$ & $M=20$ & $M=40$ & $M=100$ & $M=200$ & $M=2000$ \\
    \includegraphics[width=0.125\linewidth]{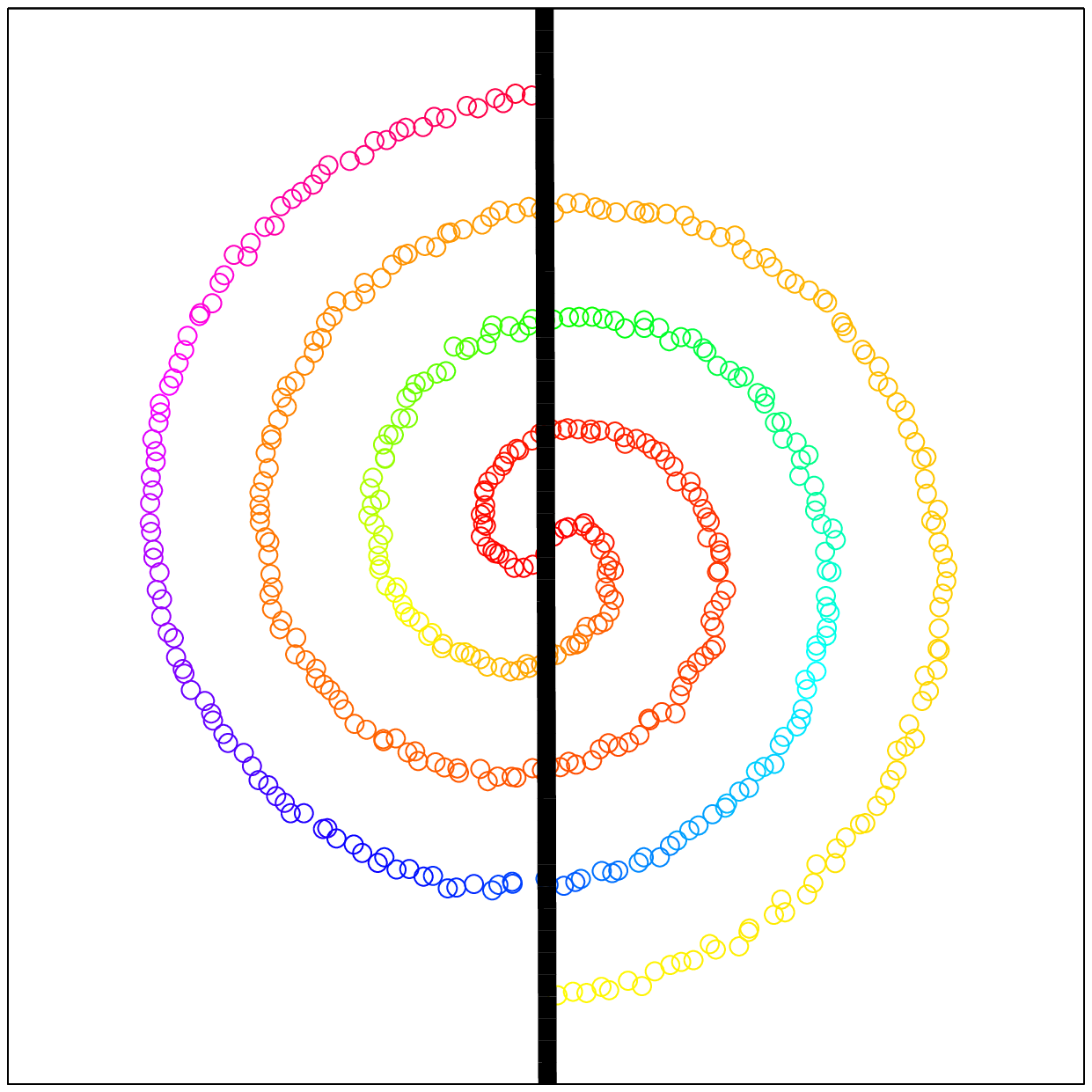} &
    \includegraphics[width=0.125\linewidth]{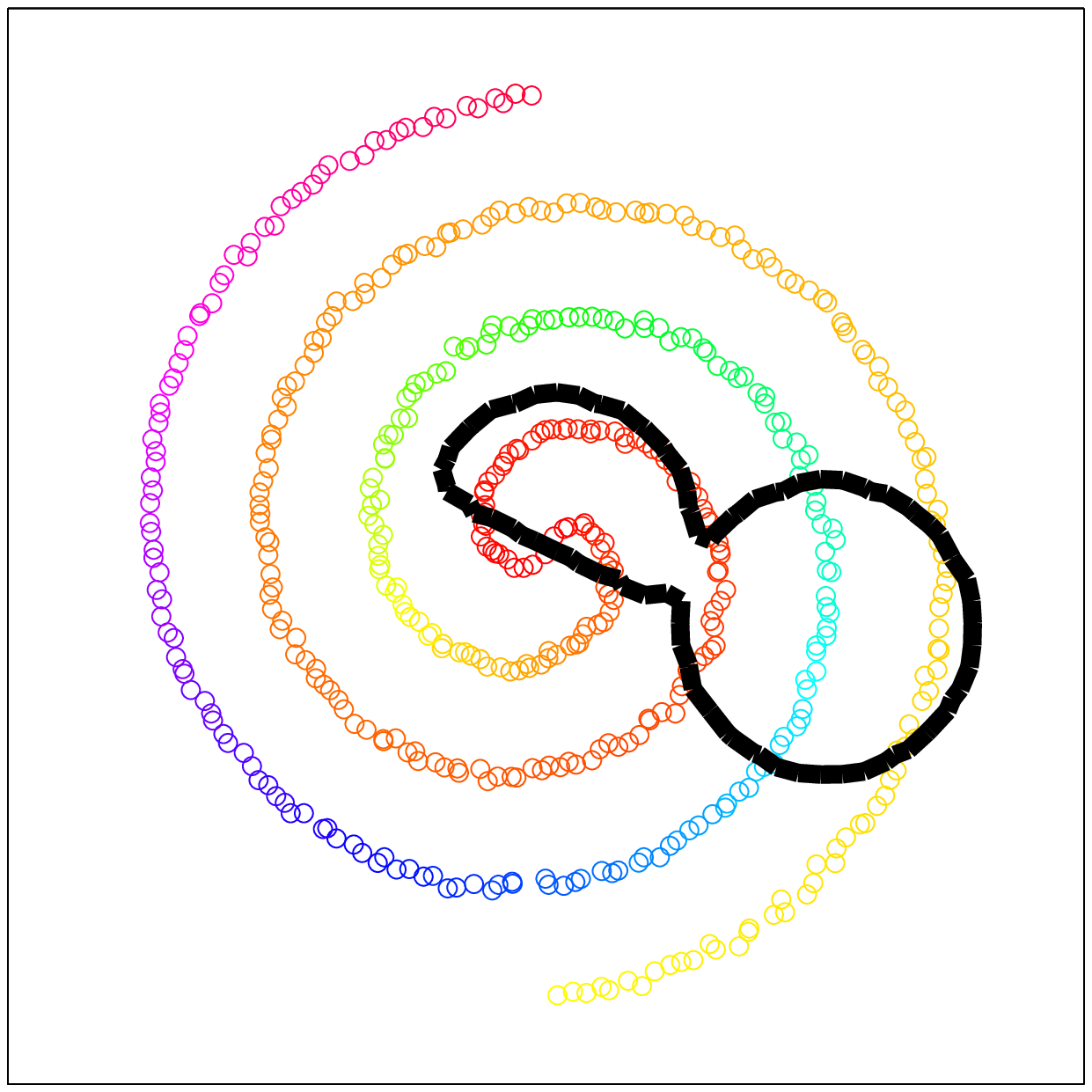} &
    \includegraphics[width=0.125\linewidth]{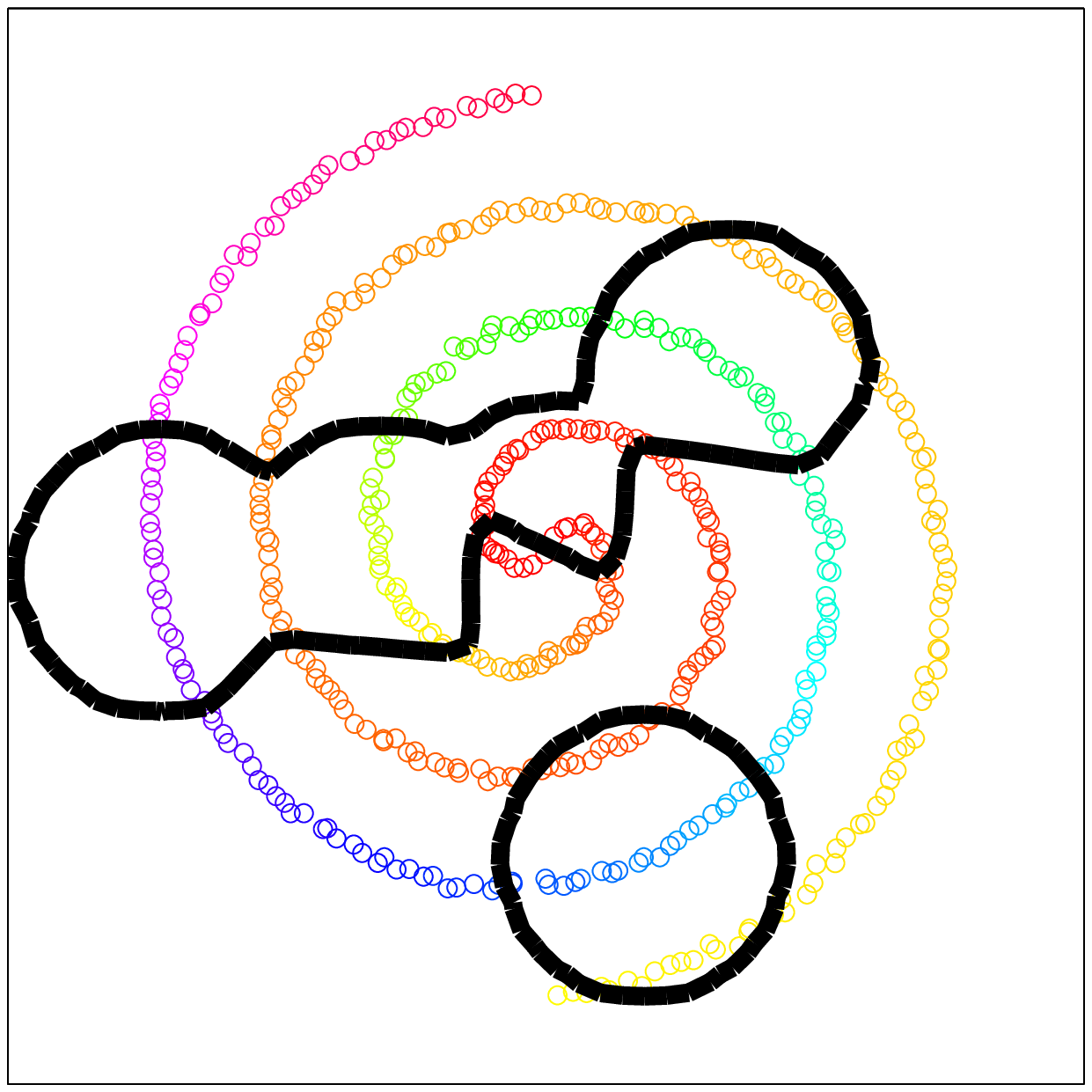} &
    \includegraphics[width=0.125\linewidth]{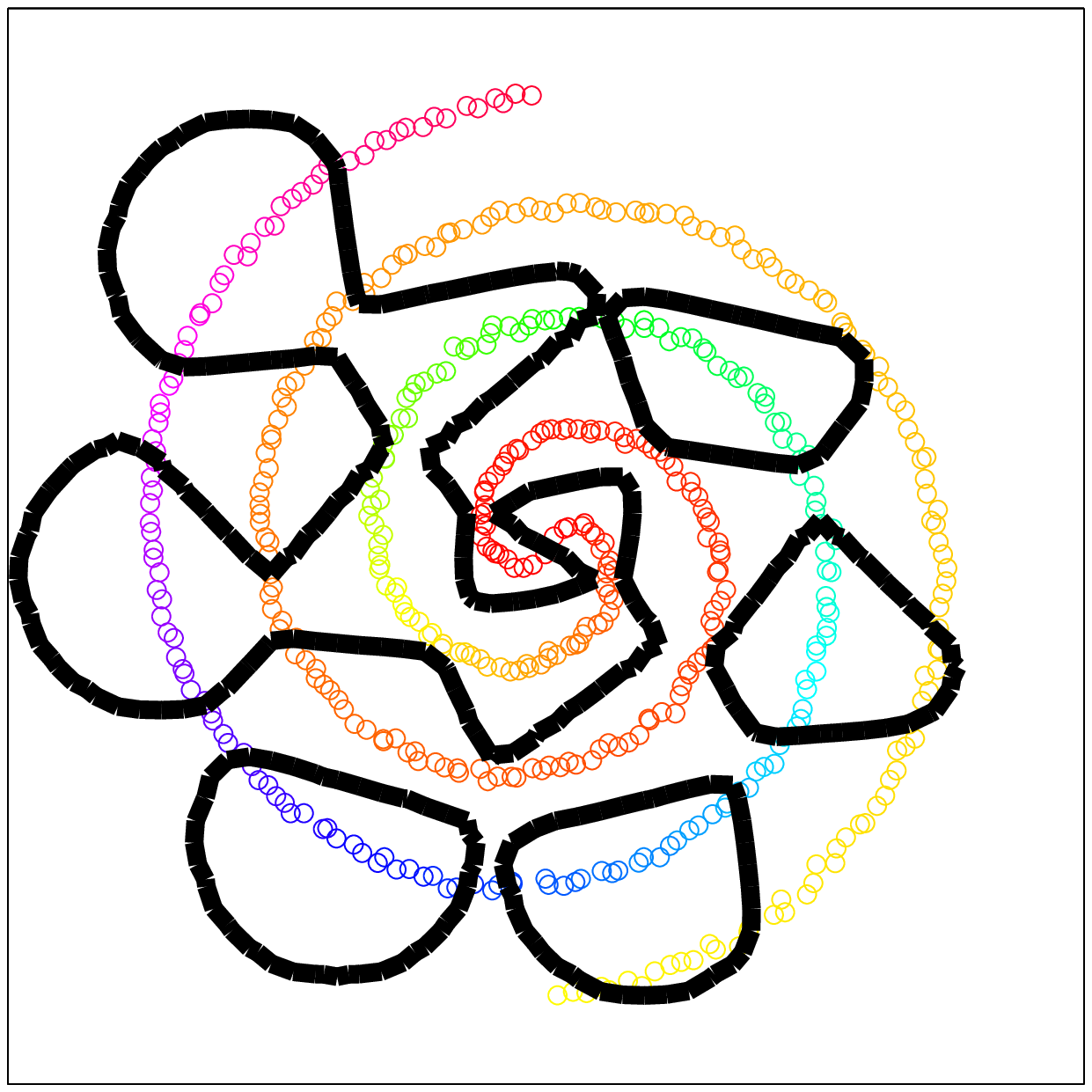} &
    \includegraphics[width=0.125\linewidth]{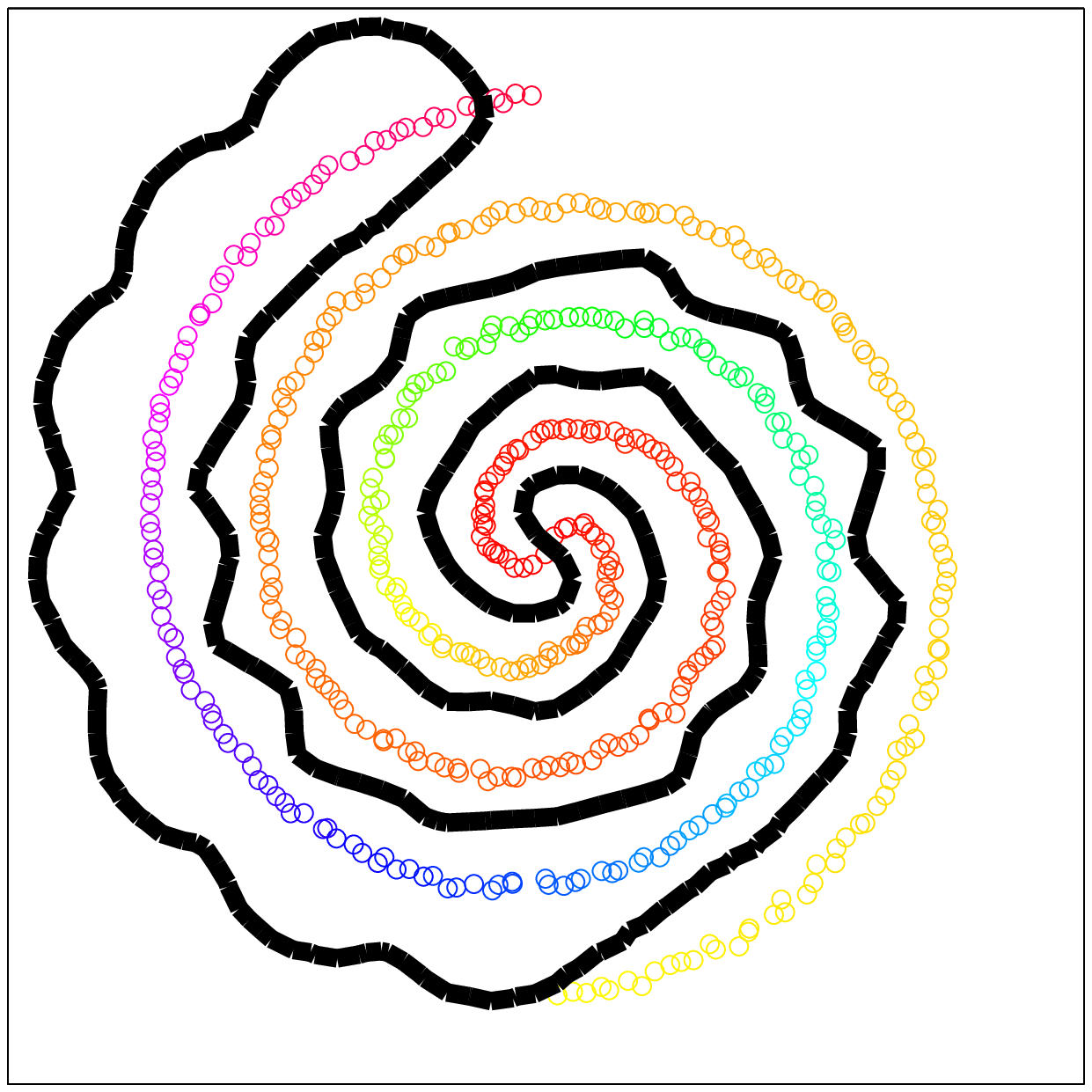} &
    \includegraphics[width=0.125\linewidth]{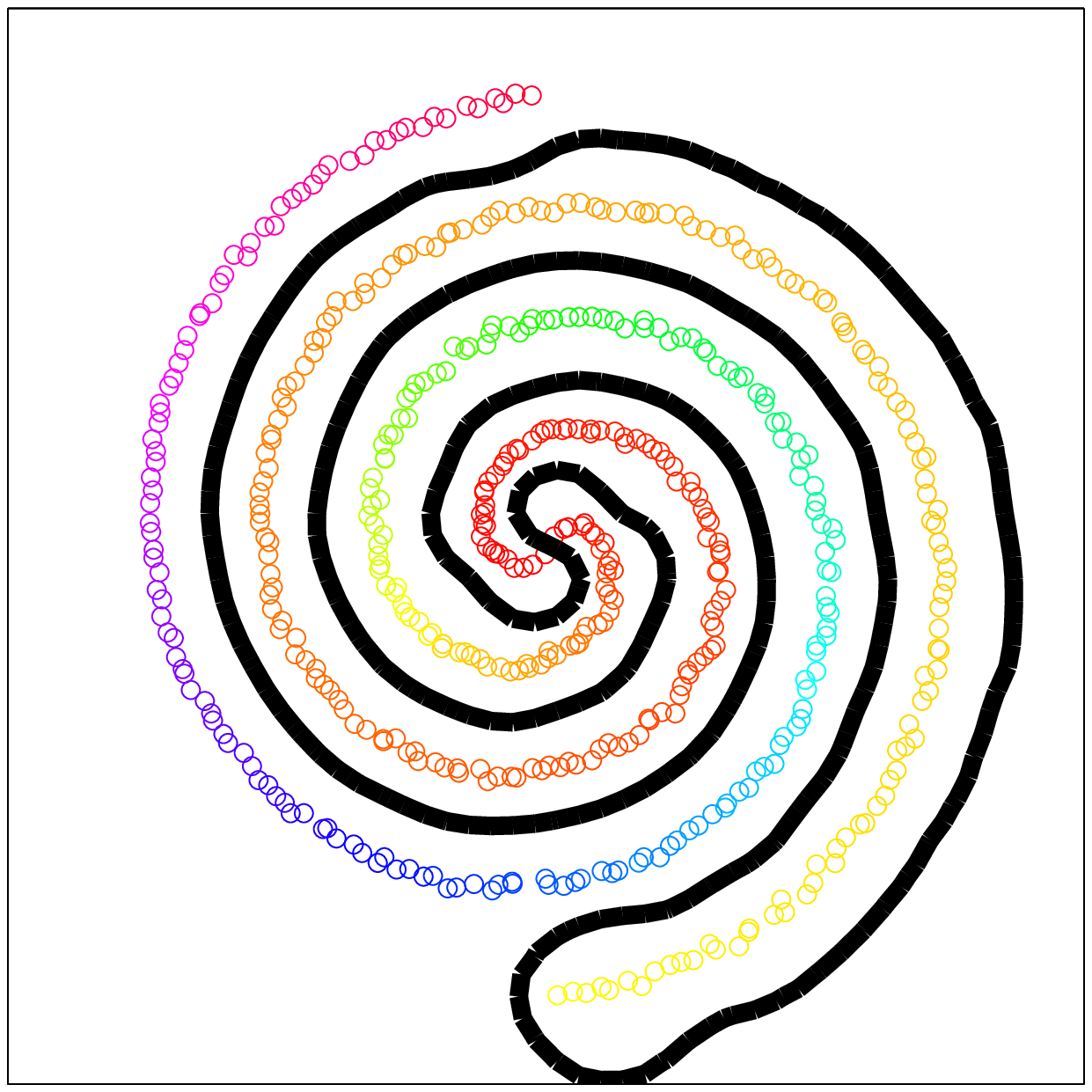} &
    \includegraphics[width=0.125\linewidth]{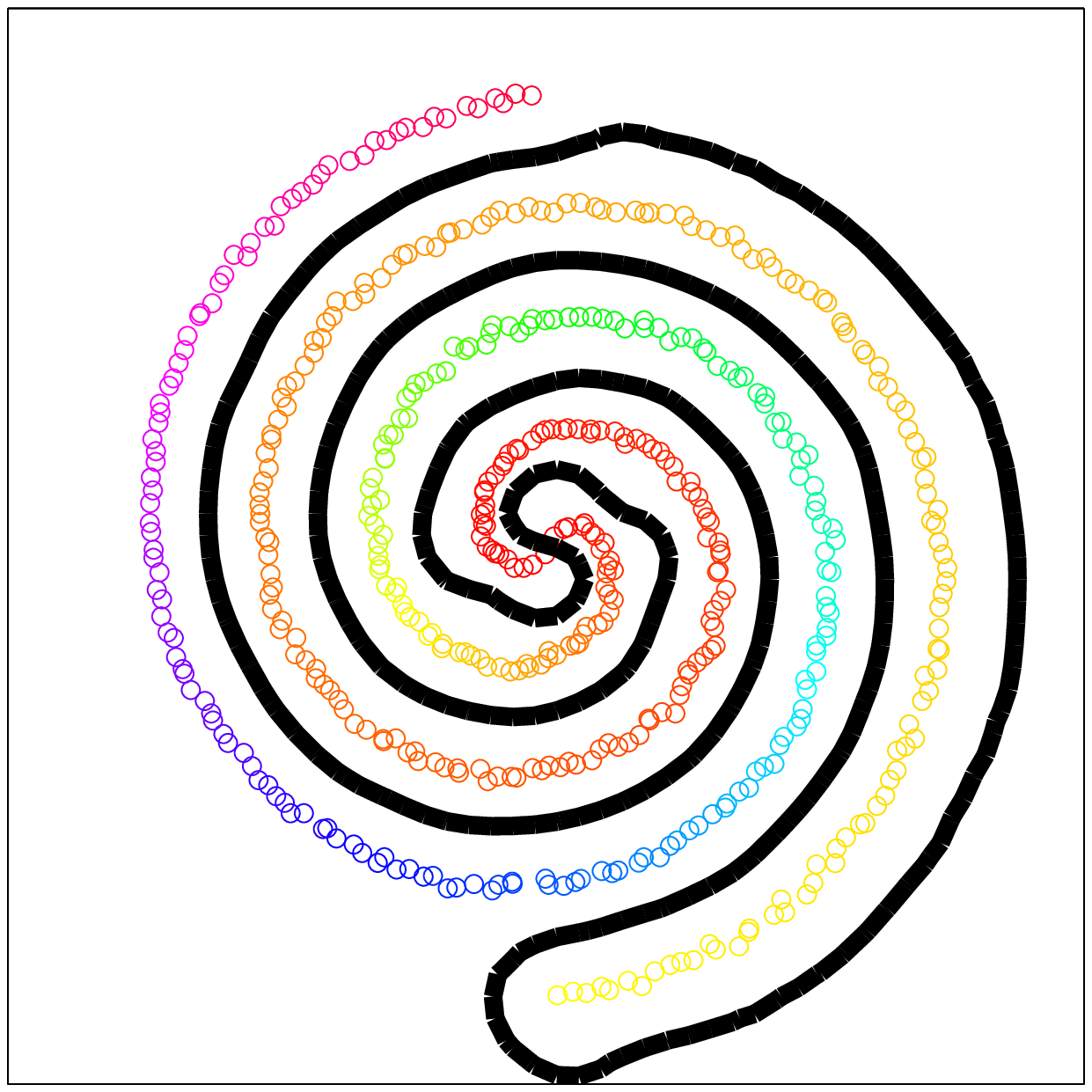} &
    \includegraphics[width=0.125\linewidth]{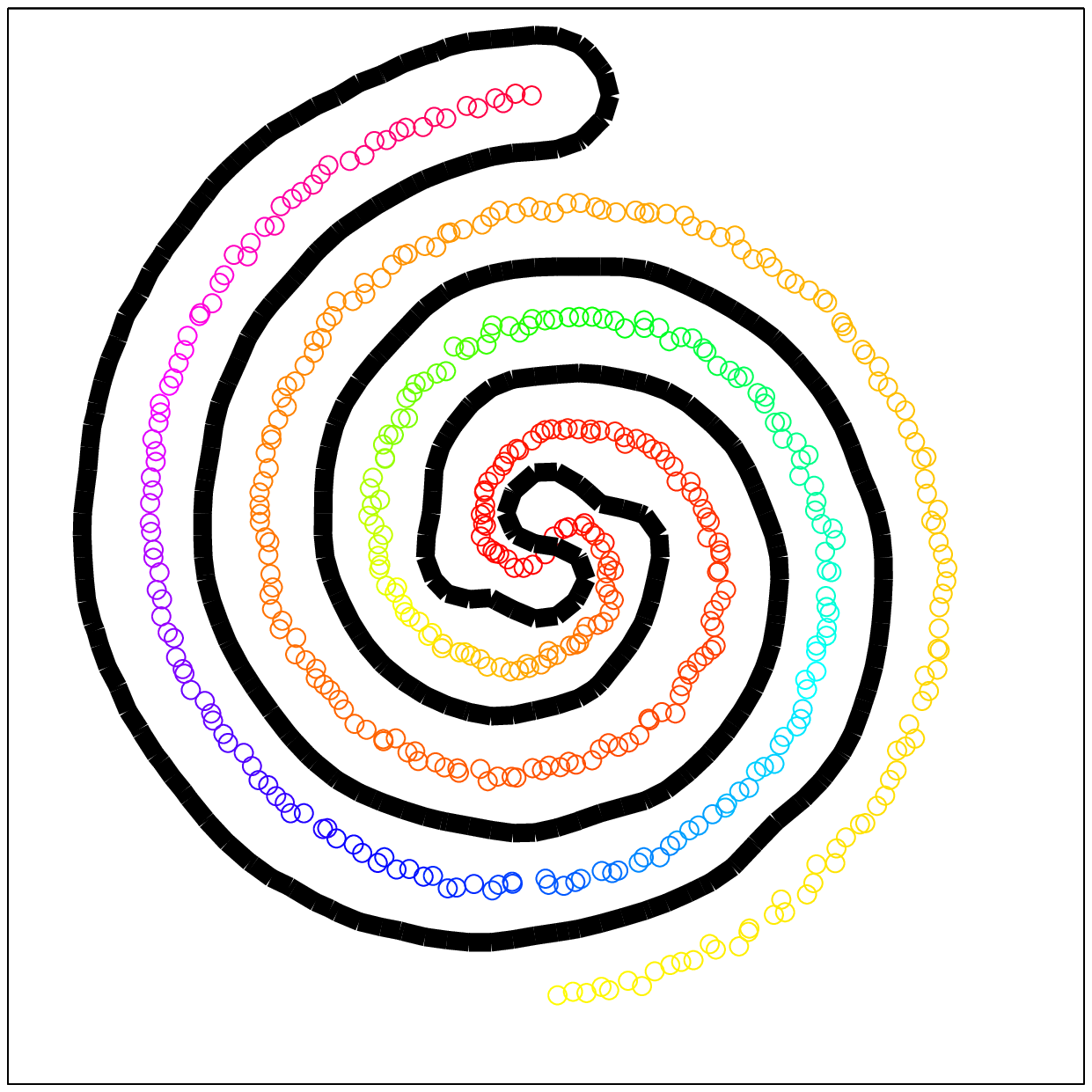} \\[-.6ex]
    \includegraphics[width=0.125\linewidth]{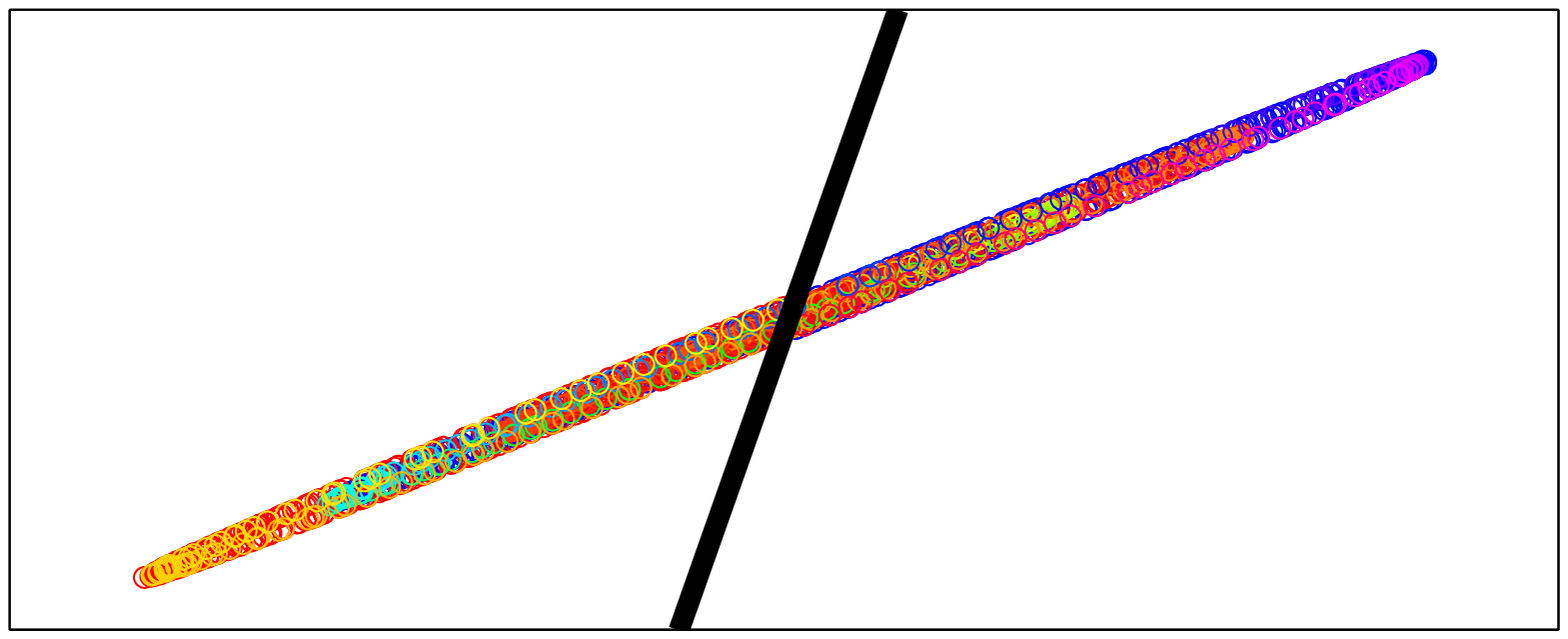} &
    \includegraphics[width=0.125\linewidth]{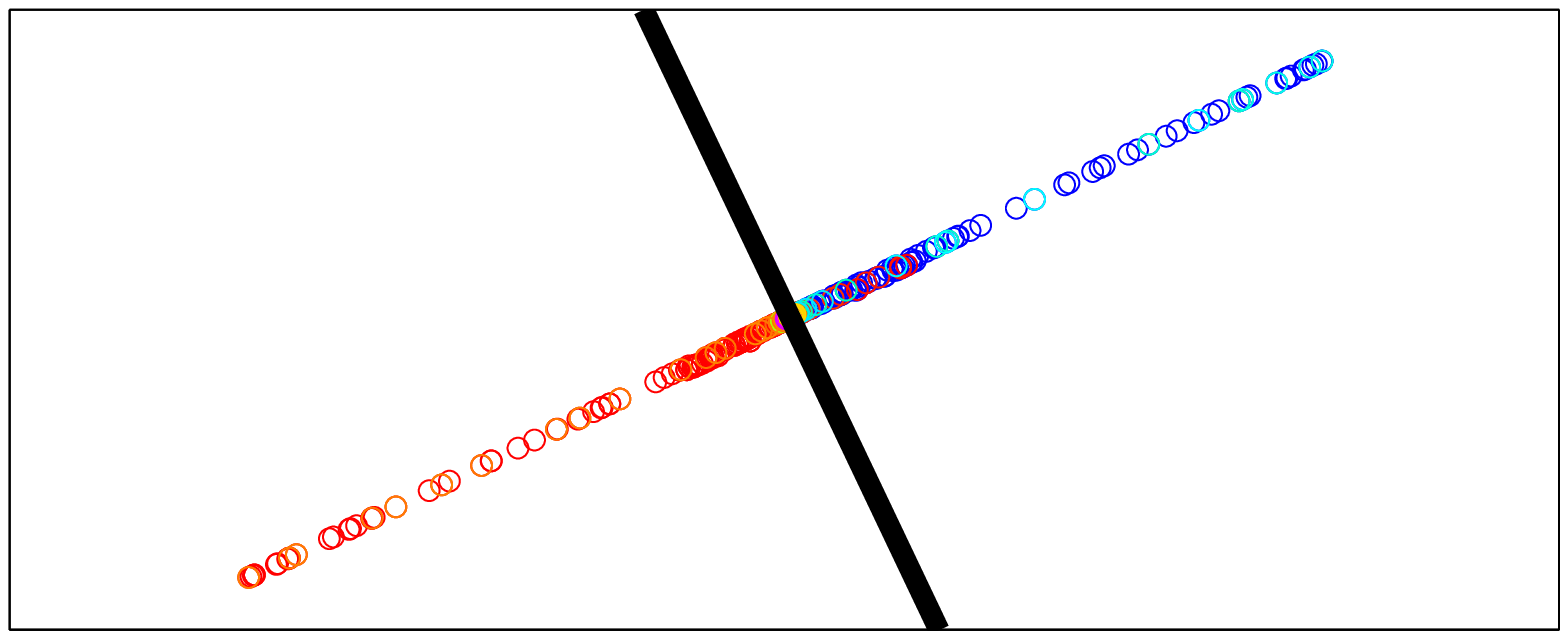} &
    \includegraphics[width=0.125\linewidth]{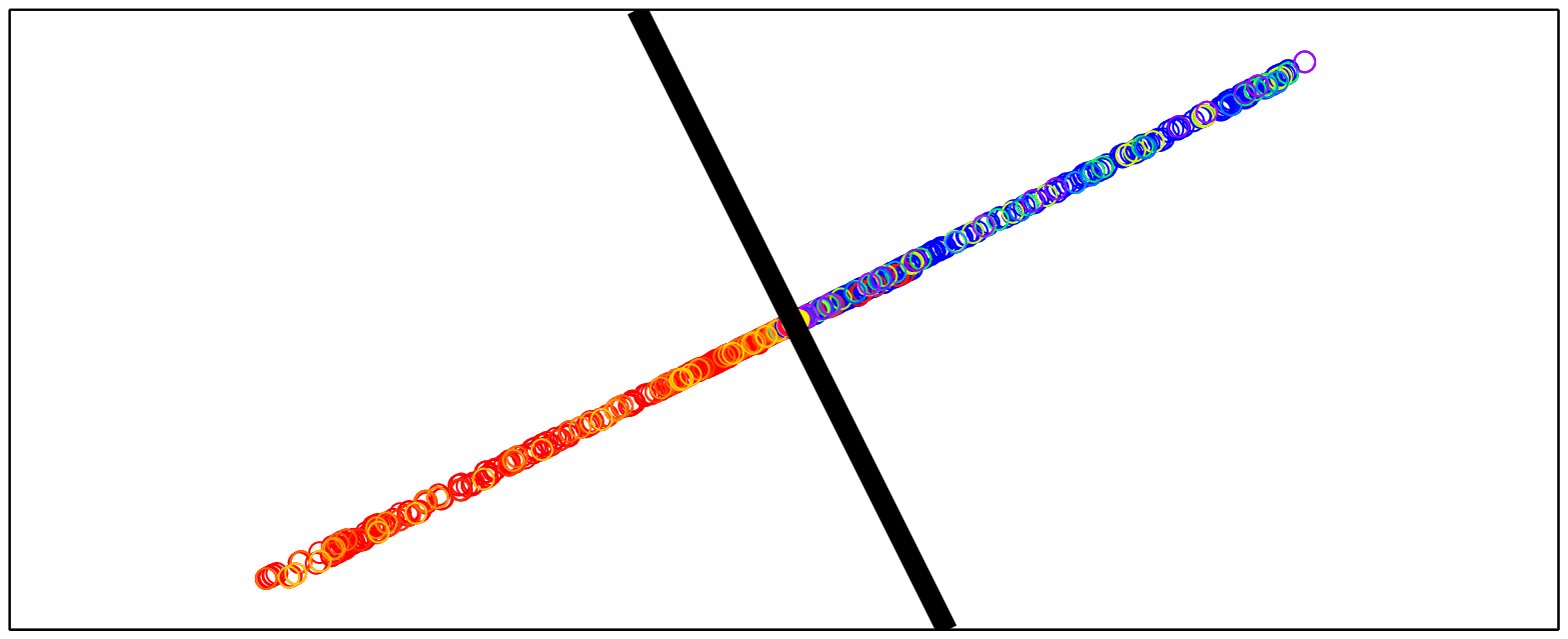} &
    \includegraphics[width=0.125\linewidth]{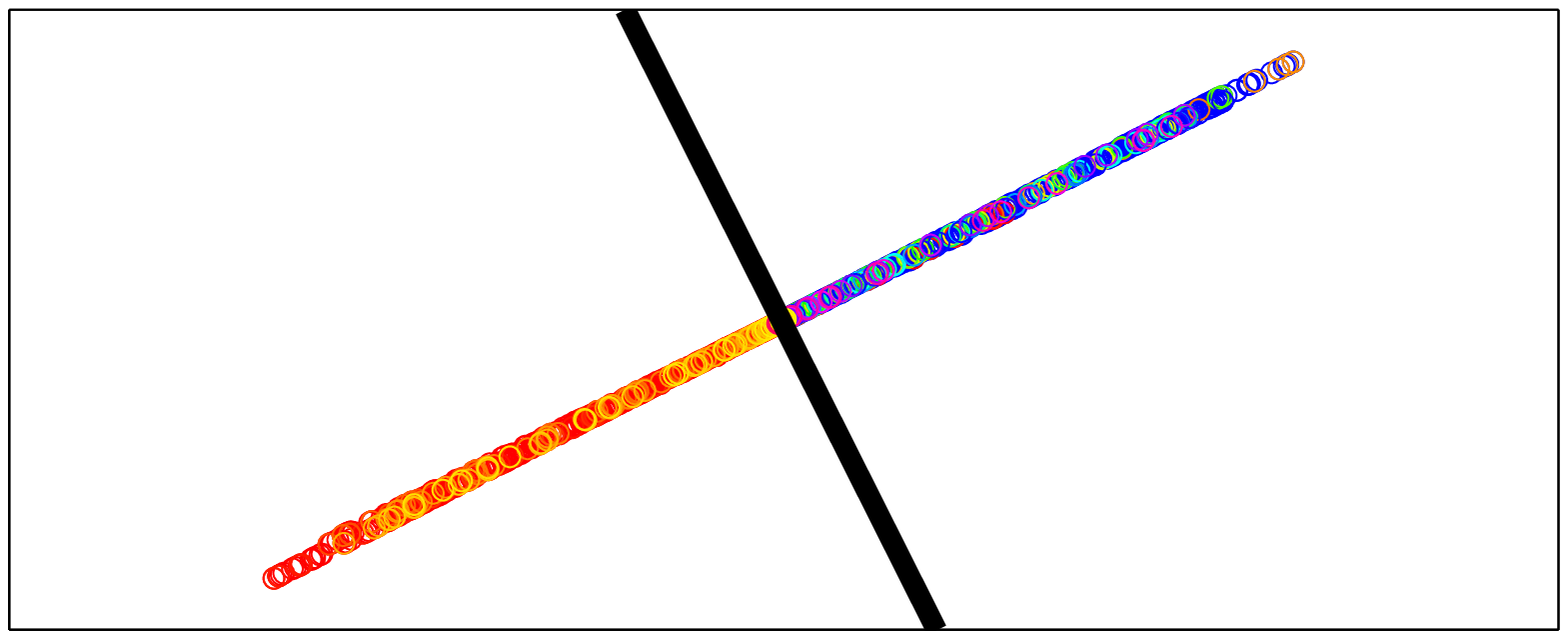} &
    \includegraphics[width=0.125\linewidth]{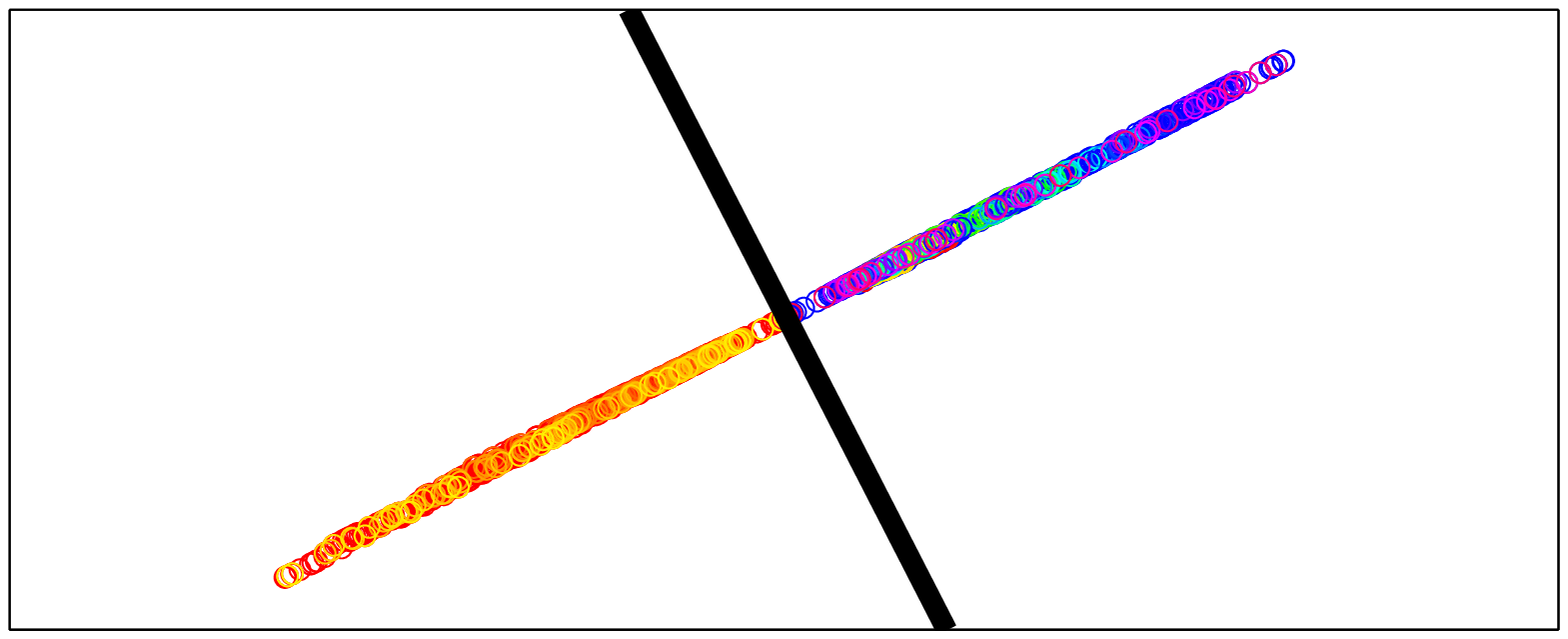} &
    \includegraphics[width=0.125\linewidth]{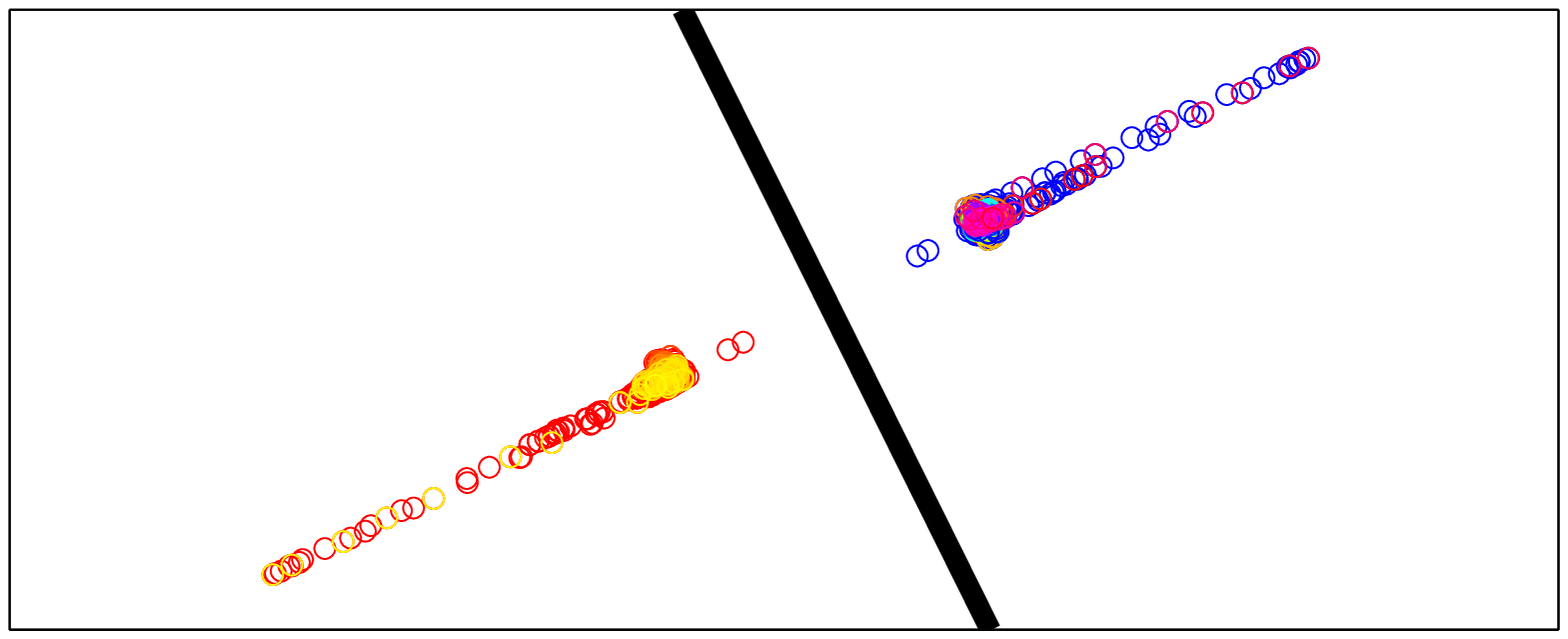} &
    \includegraphics[width=0.125\linewidth]{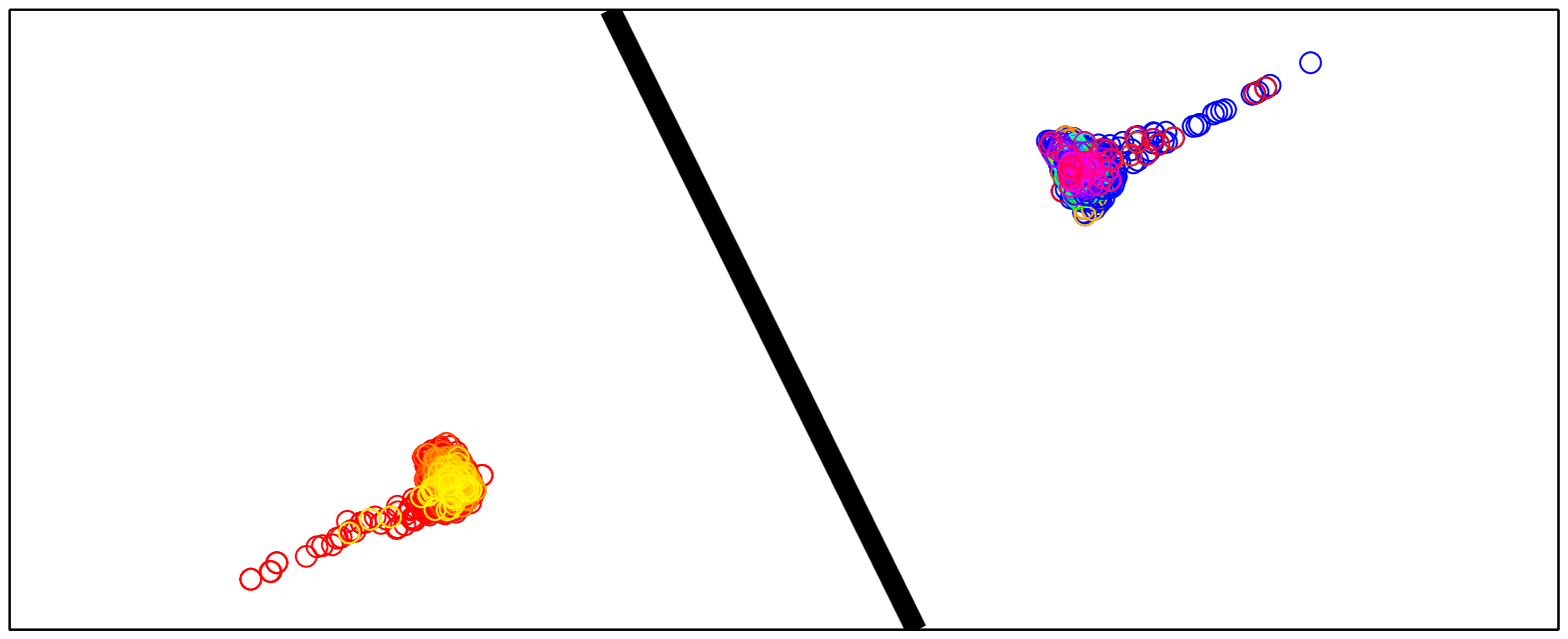} &
    \includegraphics[width=0.125\linewidth]{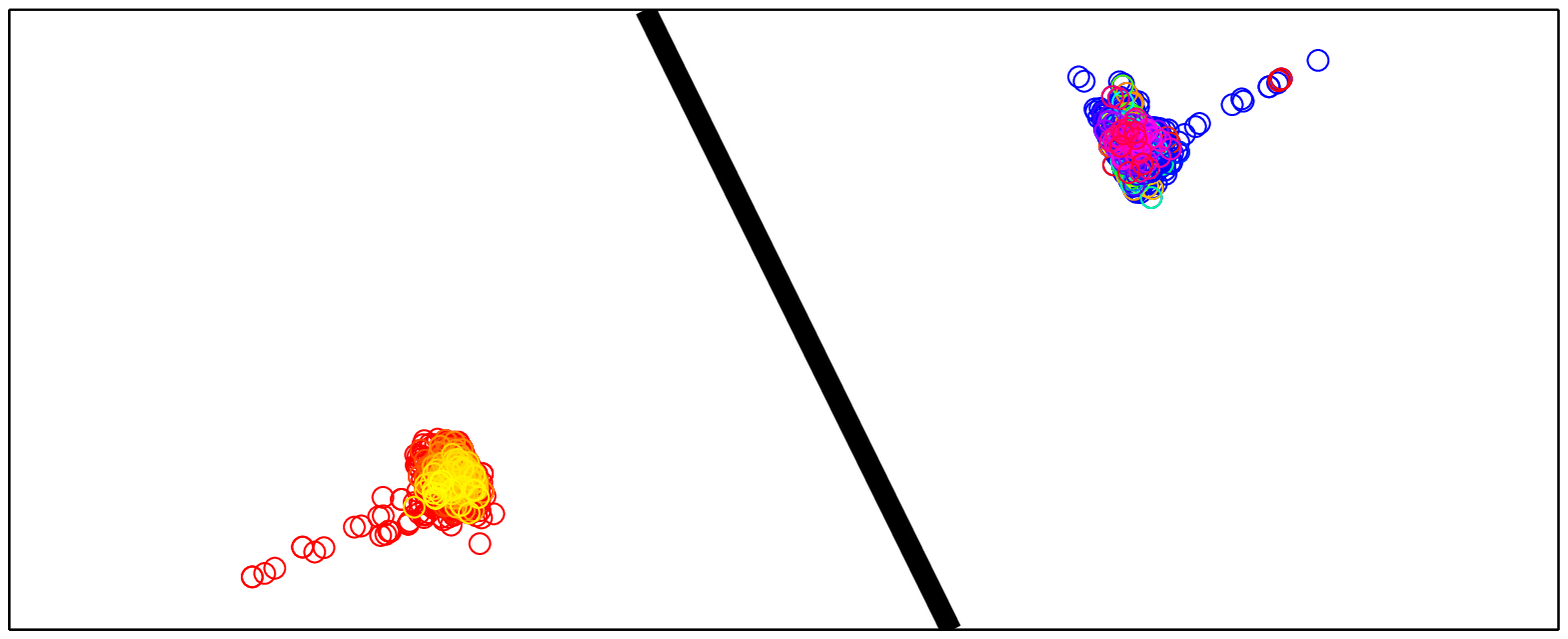} 
  \end{tabular}
  \caption{Results on the two spirals dataset. We vary the flexibility of \F\ to investigate the ideal dimensionality reduction for \g. \emph{Top}: dataset and decision boundary (black curves) achieved with different \F. \emph{Bottom}: representation achieved in the 2D latent space, i.e., the data projections $\F(\X)$ and the classification boundary of the SVM \g\ (black line).}
  \label{f:2spirals}
\end{figure*}

We investigate this question experimentally. The dataset in fig.~\ref{f:2spirals} contains two spirals, each has $1\,000$ samples and defines a class. We change the flexibility of \F\ to see what latent representations are obtained. We try a linear DR and a RBF DR with varying number of basis functions ($M$ centers uniformly sampled from the training set) while keeping the other hyperparameters fixed.

We observe the following from the projections $\F(\X)$ shown in fig.~\ref{f:2spirals}. (1) Since the dataset is not linearly separable, the two classes overlap severely in the latent space for linear \F\ (first column). (2) We see from the latent representations (shown in the second row) that, the more flexible \F\ is, the more classes collapse and training samples from different classes are pushed far apart (especially if using $M =$ all $2000$ training samples as RBFs centers). Thus, they are easily separated by \g\ with a big margin (so the nearest neighbor classifier would do very well here). Thus, the overall classifier $\g\circ\F$ is capable of solving linearly non-separable problems given sufficient BFs. (3) To achieve a perfect classification, we need only a few basis functions ($M=100$ more than suffice here). (4) The projections $\F(\x)$ approximately form a line, implying that $L=1$ is enough to separate two classes.

\paragraph{Relation between the latent dimensionality and the number of classes} 

\begin{figure*}[t!]
  \begin{center}
    \begin{tabular}{@{}c@{\hspace{0.009\linewidth}}c@{\hspace{0\linewidth}}c@{\hspace{0\linewidth}}c@{\hspace{0\linewidth}}c@{\hspace{0\linewidth}}c@{\hspace{0\linewidth}}c@{\hspace{0\linewidth}}c@{\hspace{0\linewidth}}c@{\hspace{0\linewidth}}c@{\hspace{0\linewidth}}c@{\hspace{0\linewidth}}c@{}}
      & data & $L=1$ & $L=2$ & $L=3$ & $L=4$ & $L=5$ & $L=6$ & $L=7$ & $L=8$ & $L=9$ & $L=10$ \\
      \rotatebox{90}{\hspace{1.5ex}$K=2$} &
      \includegraphics[width=0.089\linewidth]{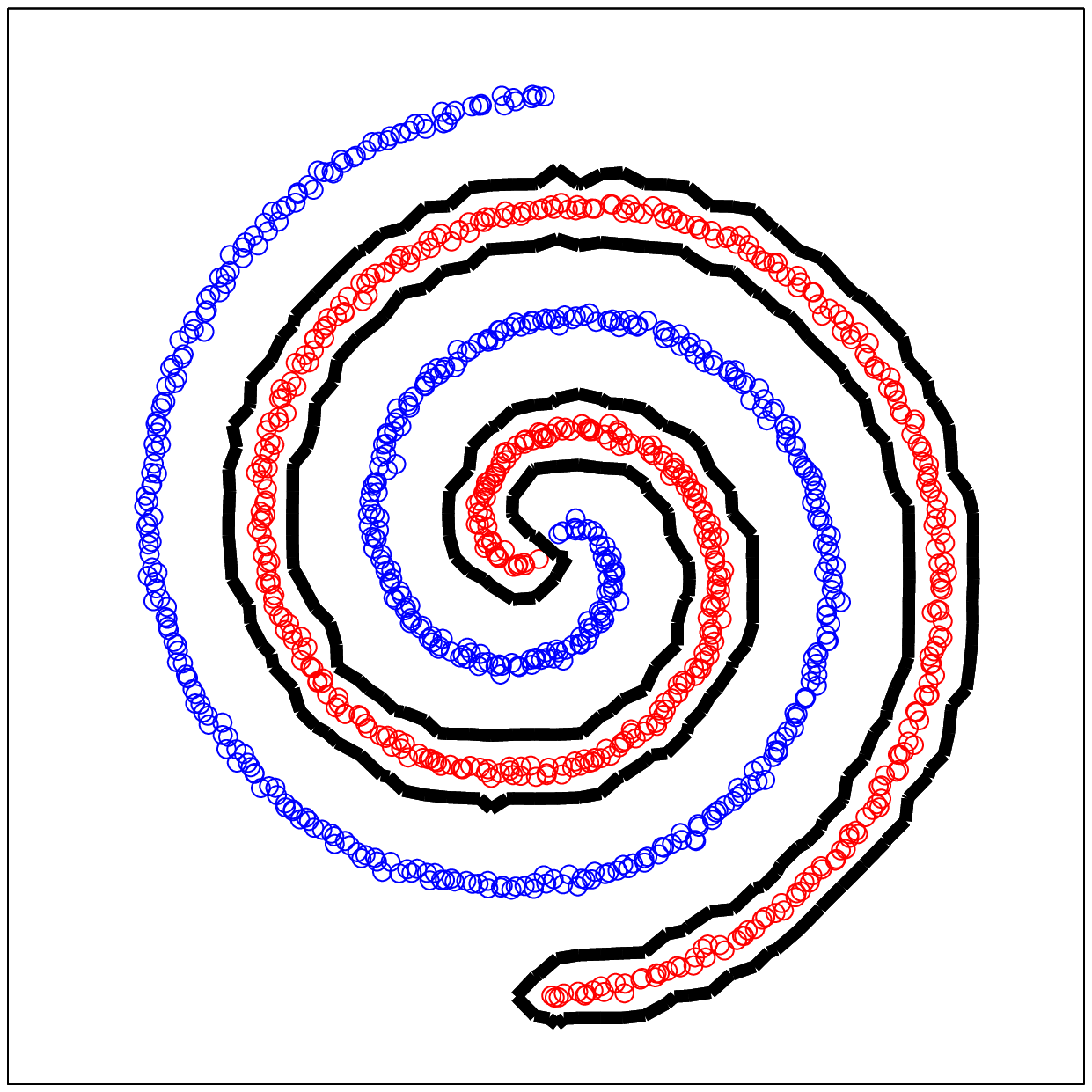} &
      \includegraphics[width=0.089\linewidth]{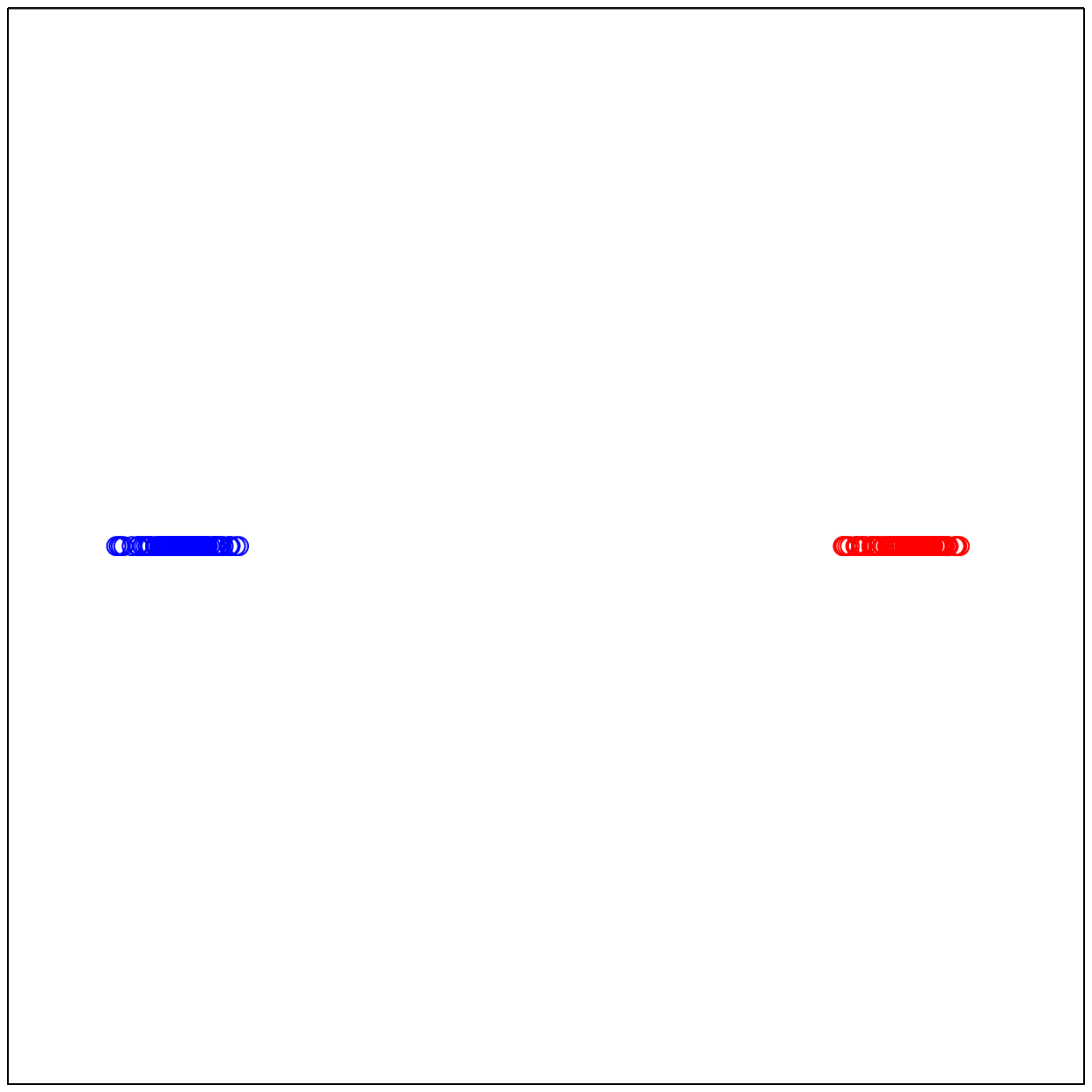} &
      \includegraphics[width=0.089\linewidth]{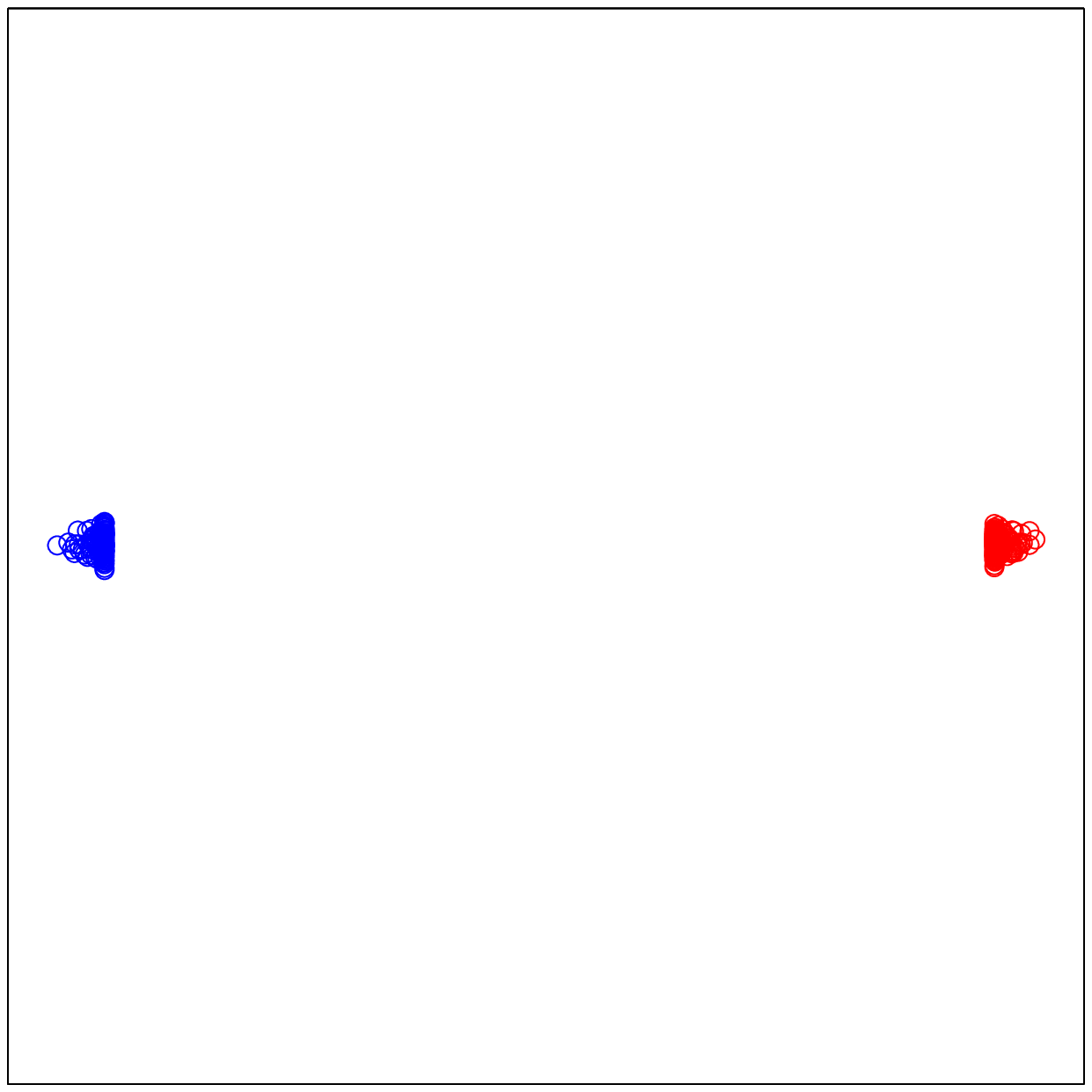} &
      \includegraphics[width=0.089\linewidth]{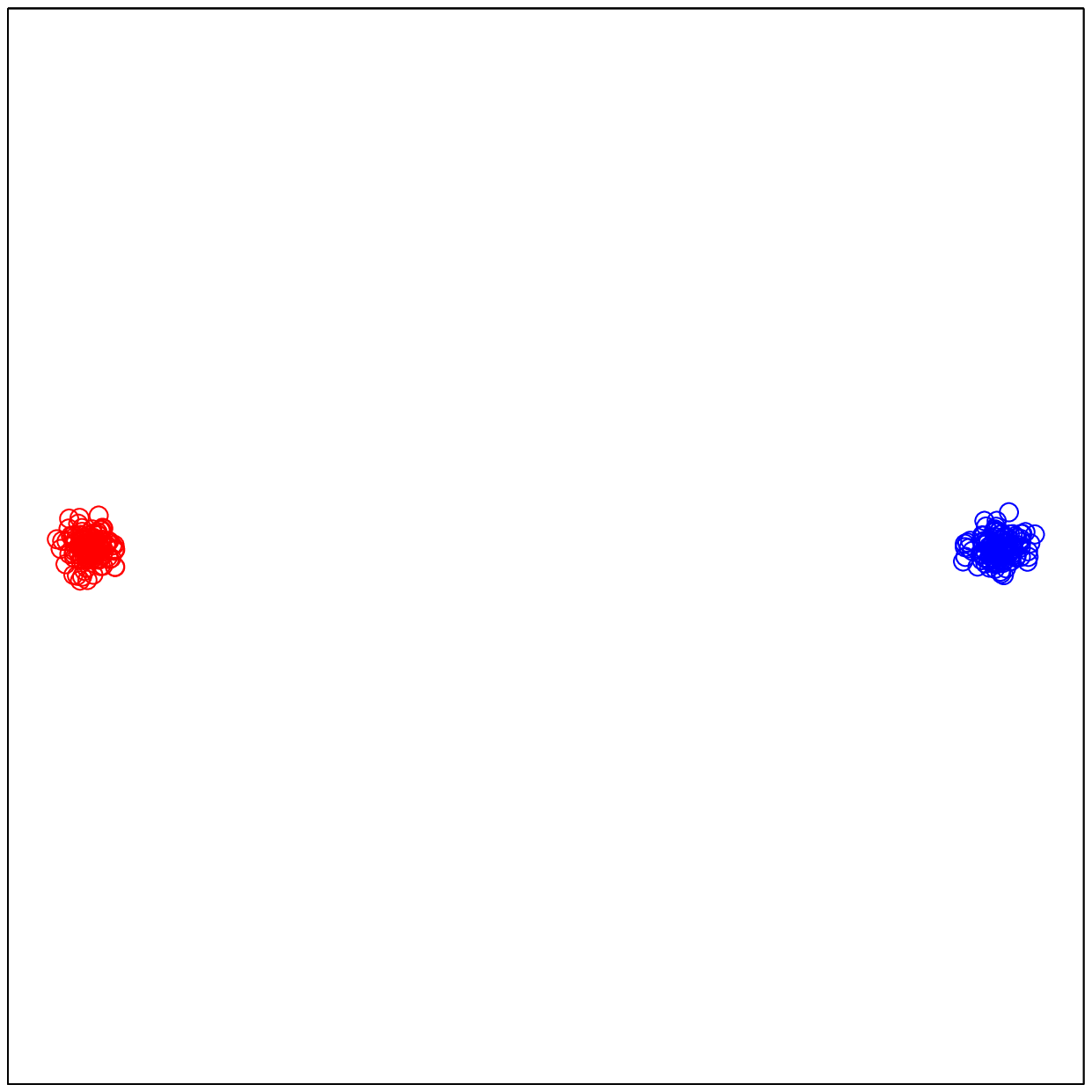} &
      \includegraphics[width=0.089\linewidth]{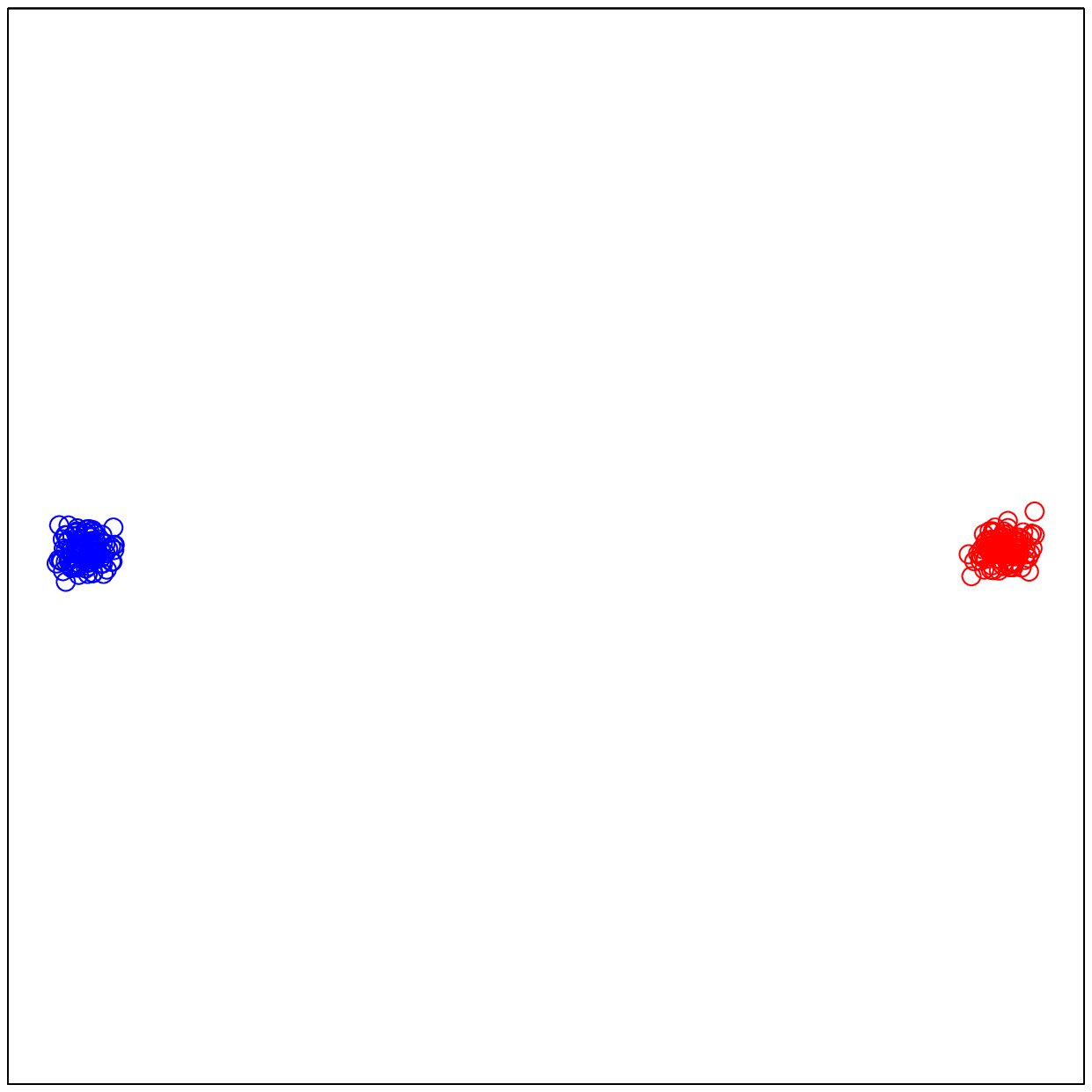} &
      \includegraphics[width=0.089\linewidth]{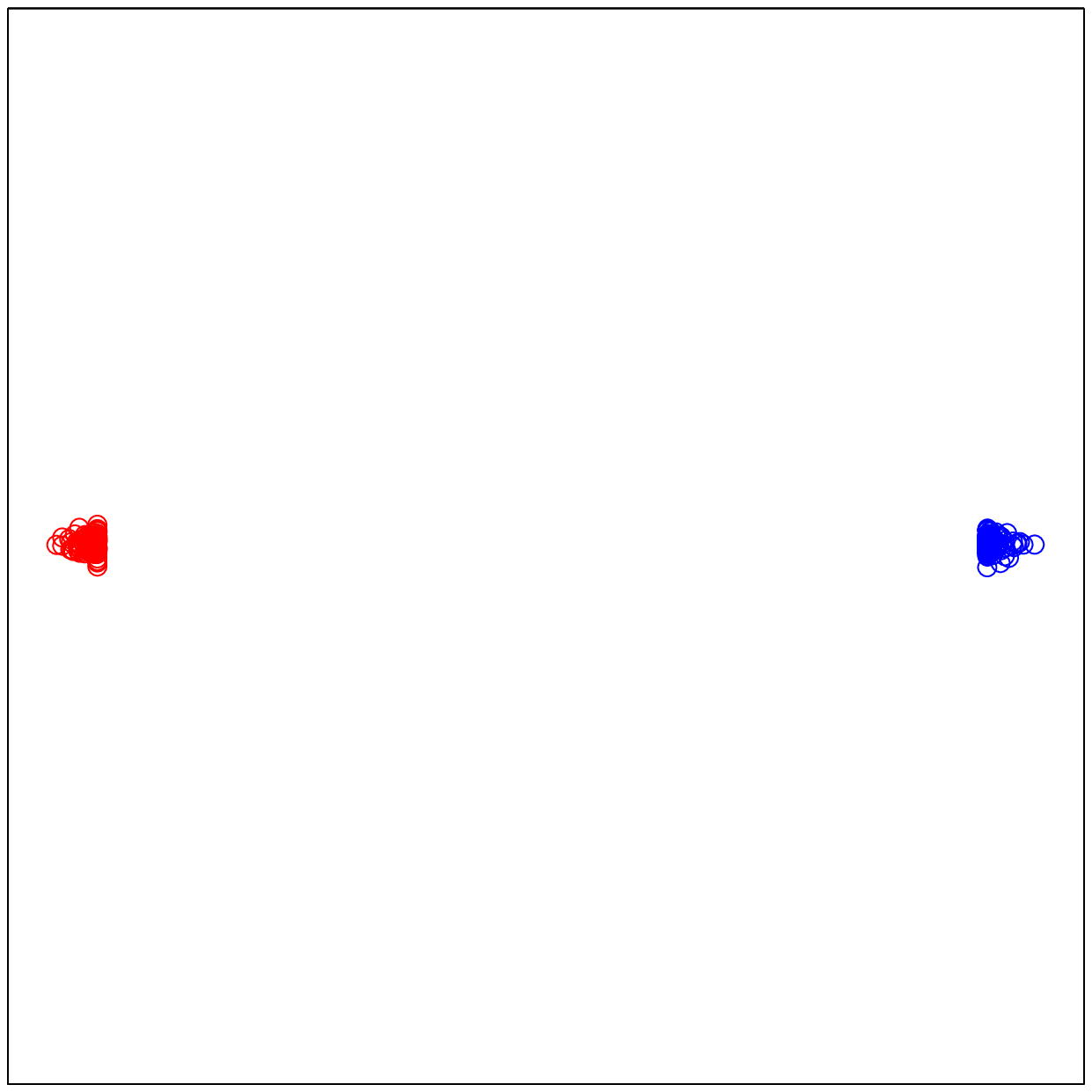} &
      \includegraphics[width=0.089\linewidth]{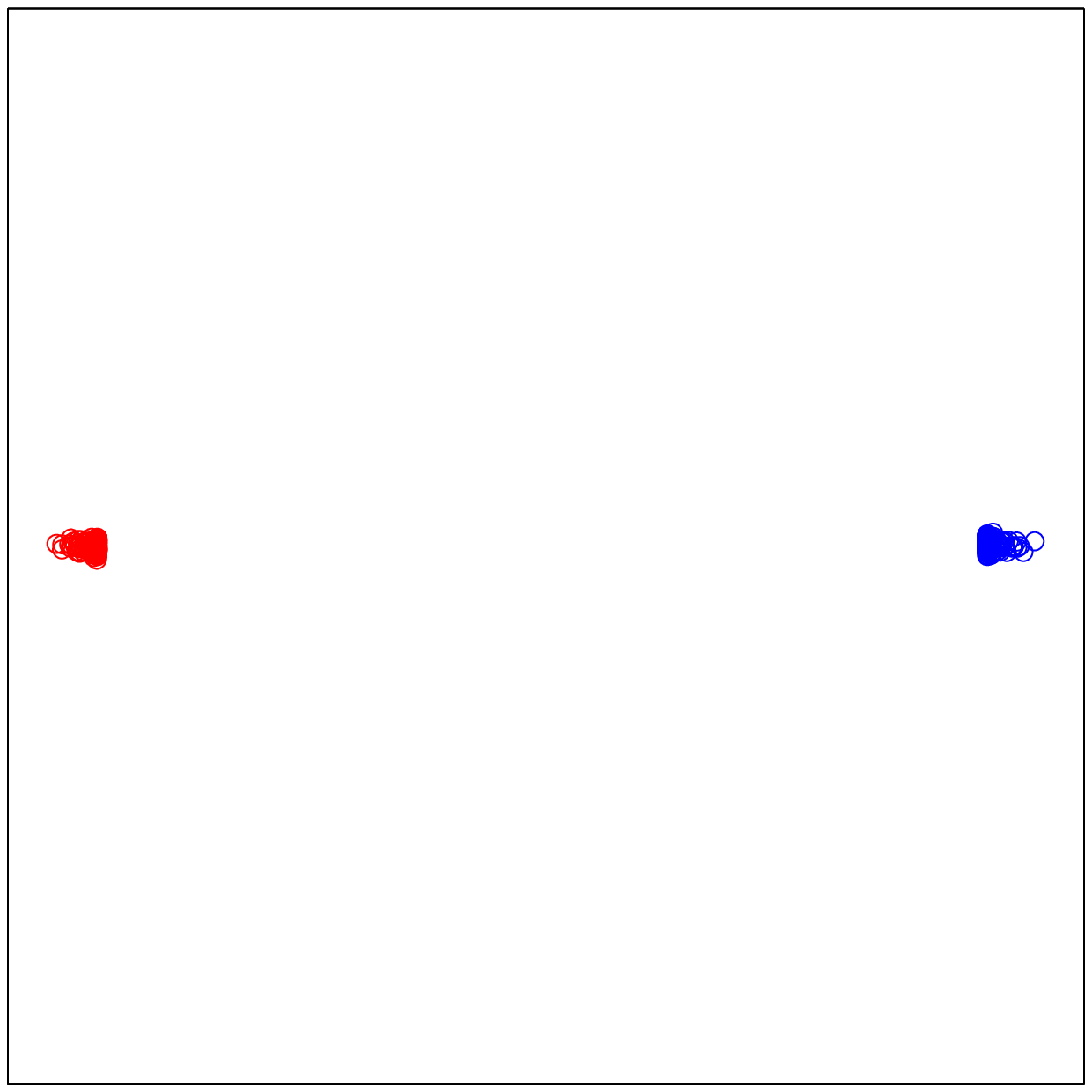} &
      \includegraphics[width=0.089\linewidth]{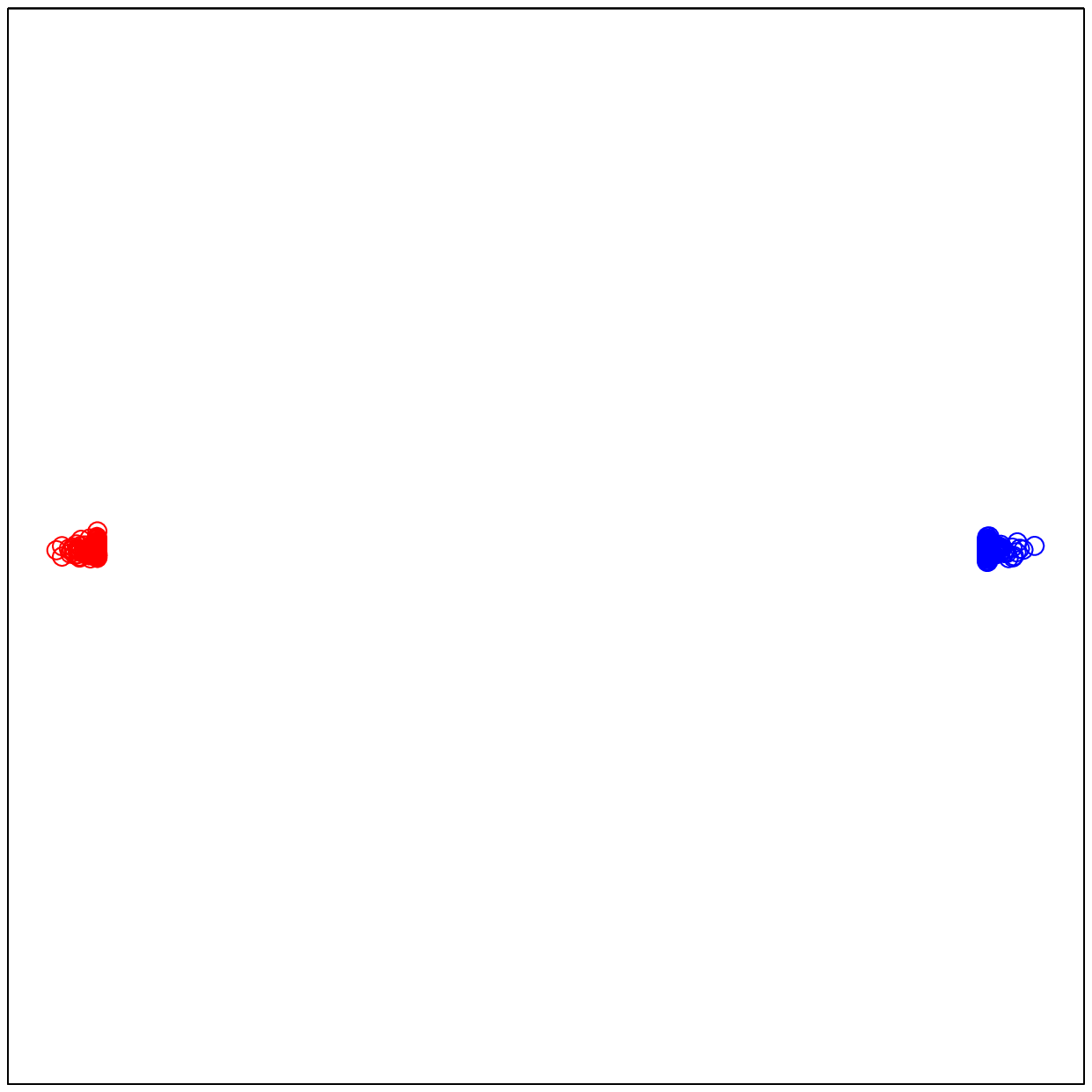} &
      \includegraphics[width=0.089\linewidth]{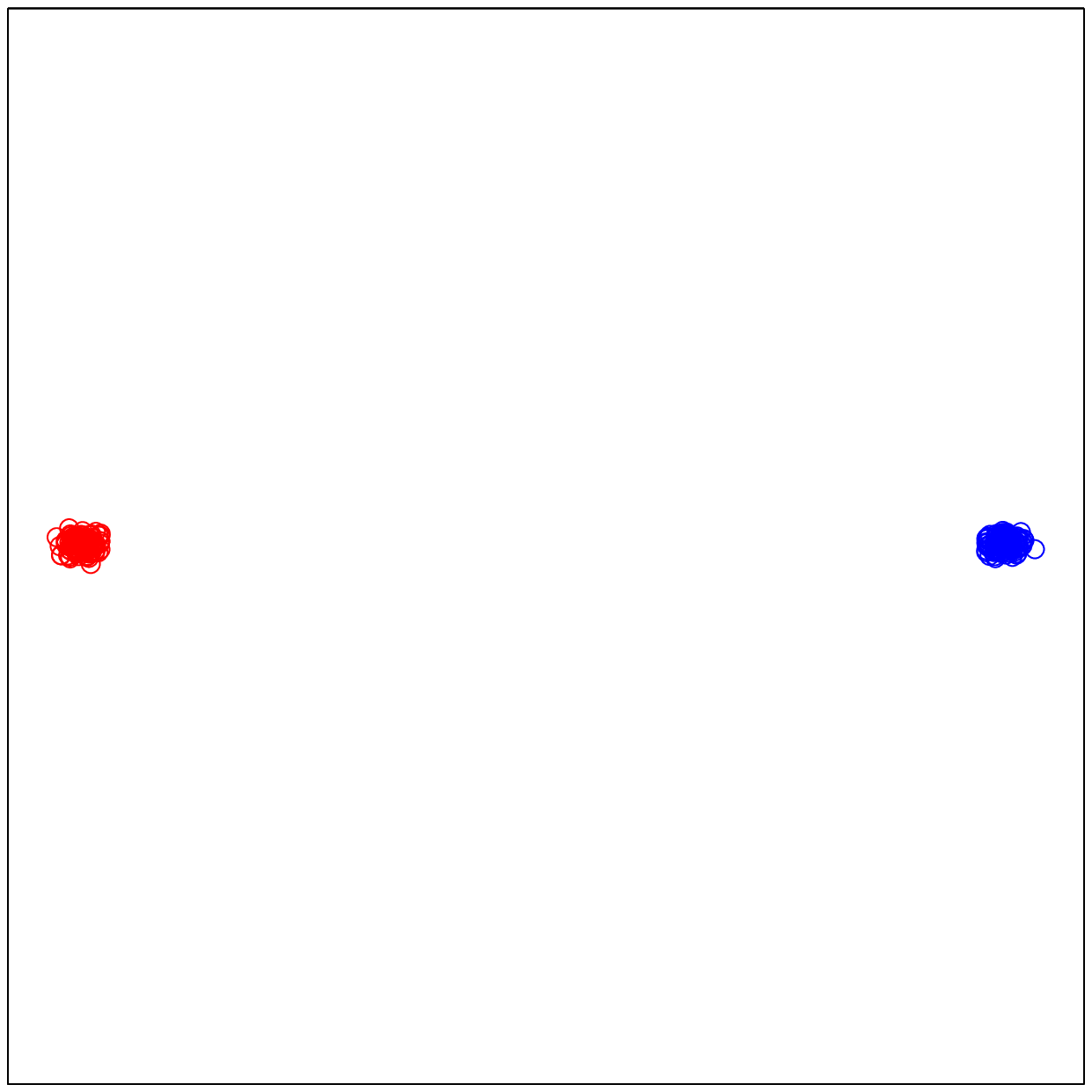} &
      \includegraphics[width=0.089\linewidth]{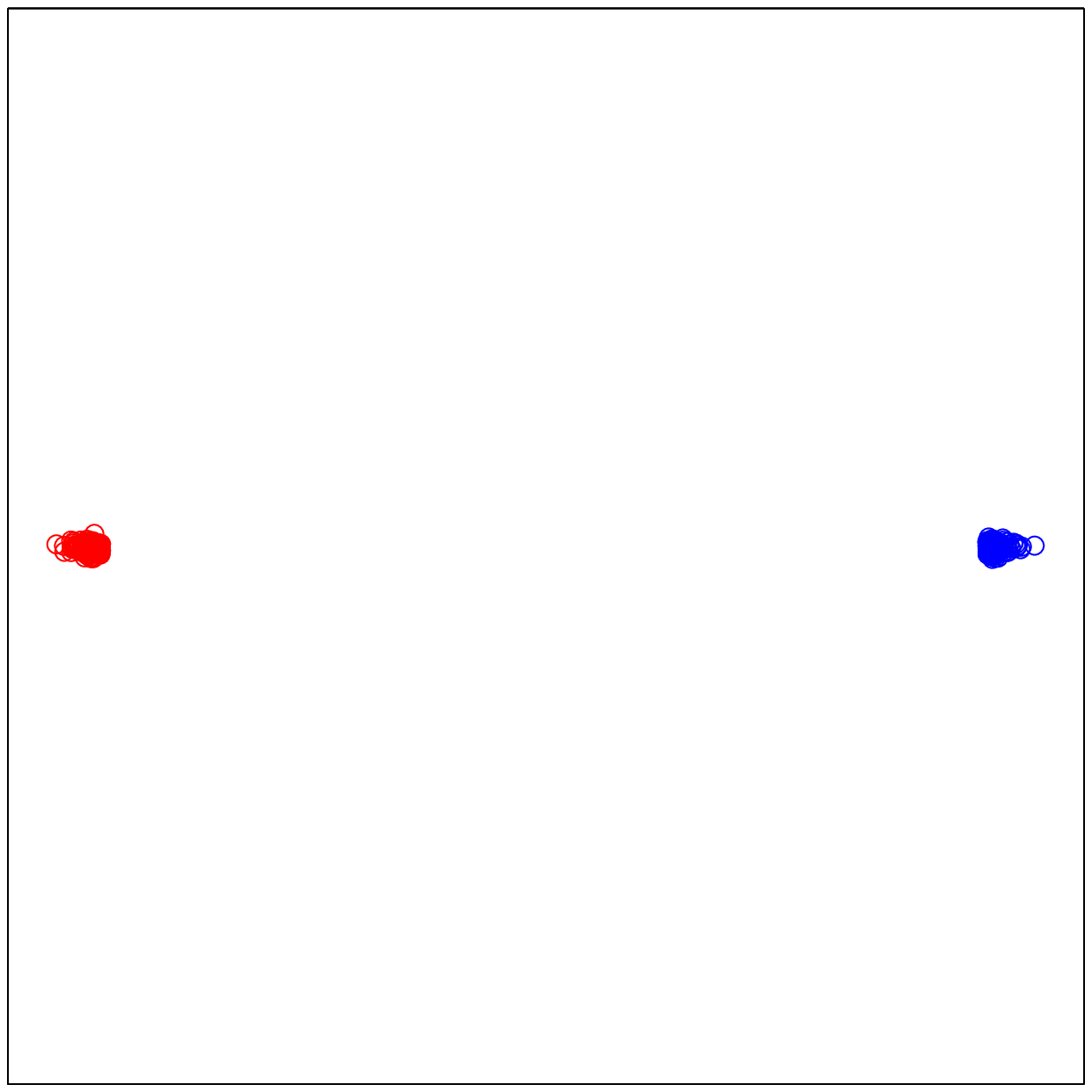} &
      \includegraphics[width=0.089\linewidth]{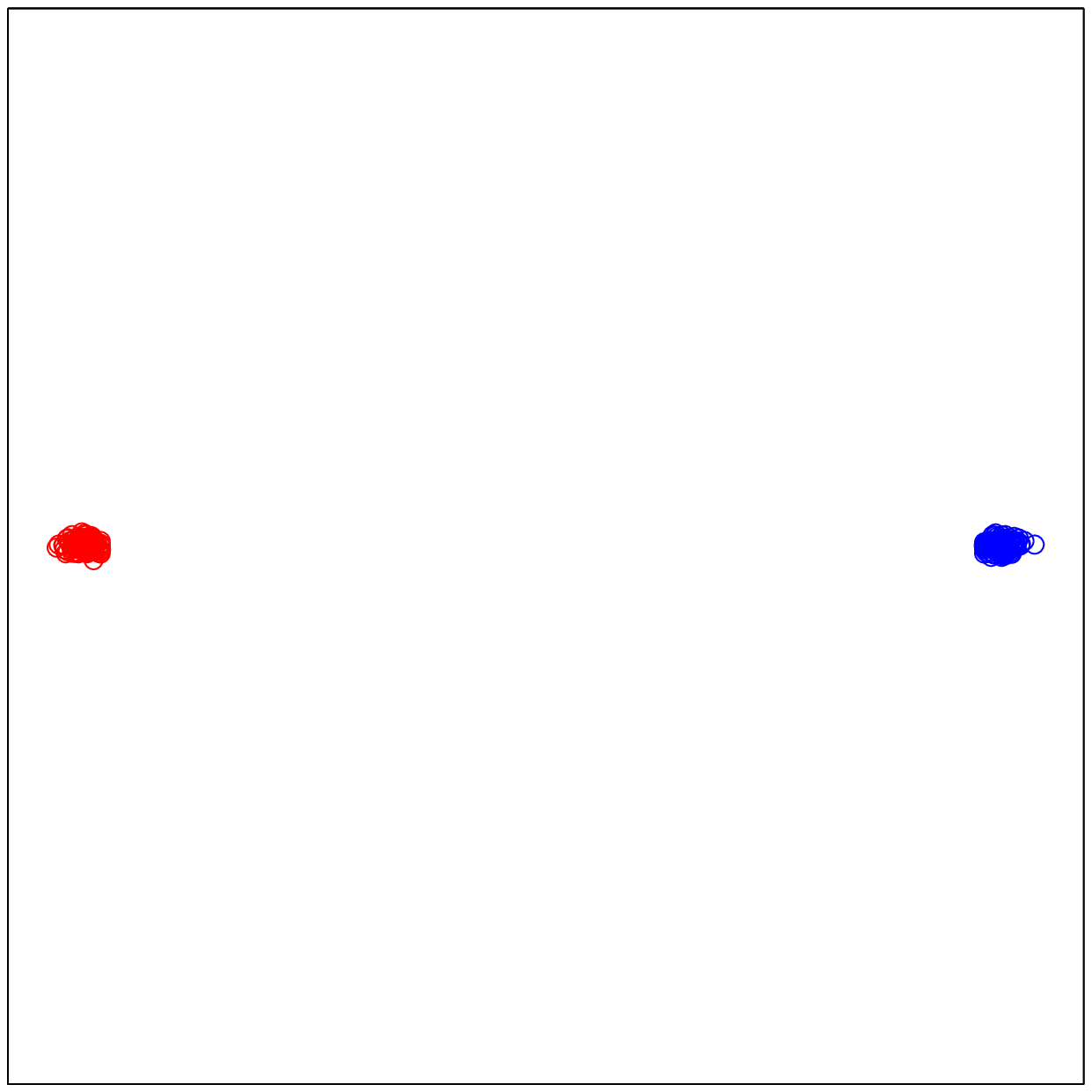} \\[-.6ex]
      \rotatebox{90}{\hspace{1.5ex}$K=3$} &
      \includegraphics[width=0.089\linewidth]{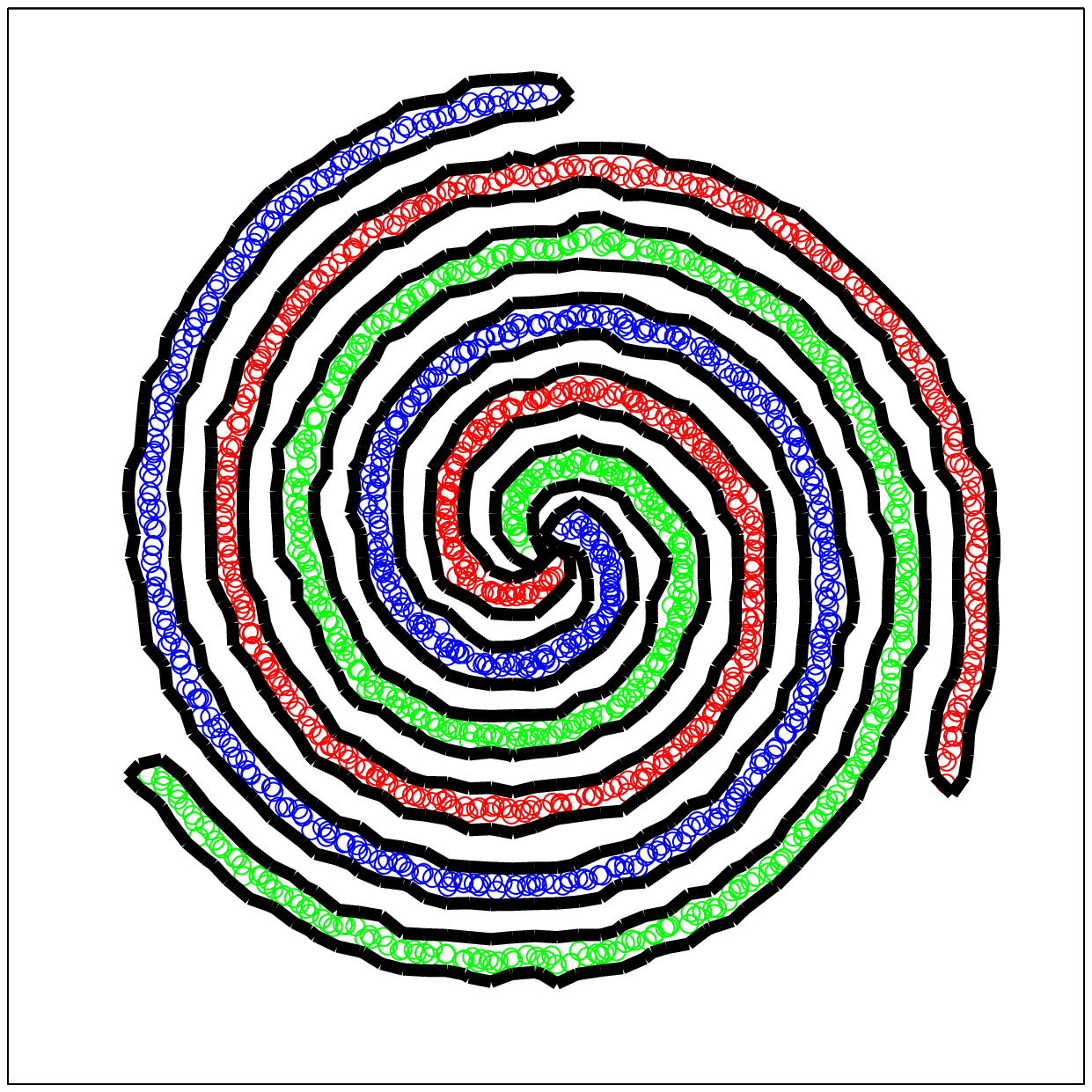} &
      \includegraphics[width=0.089\linewidth]{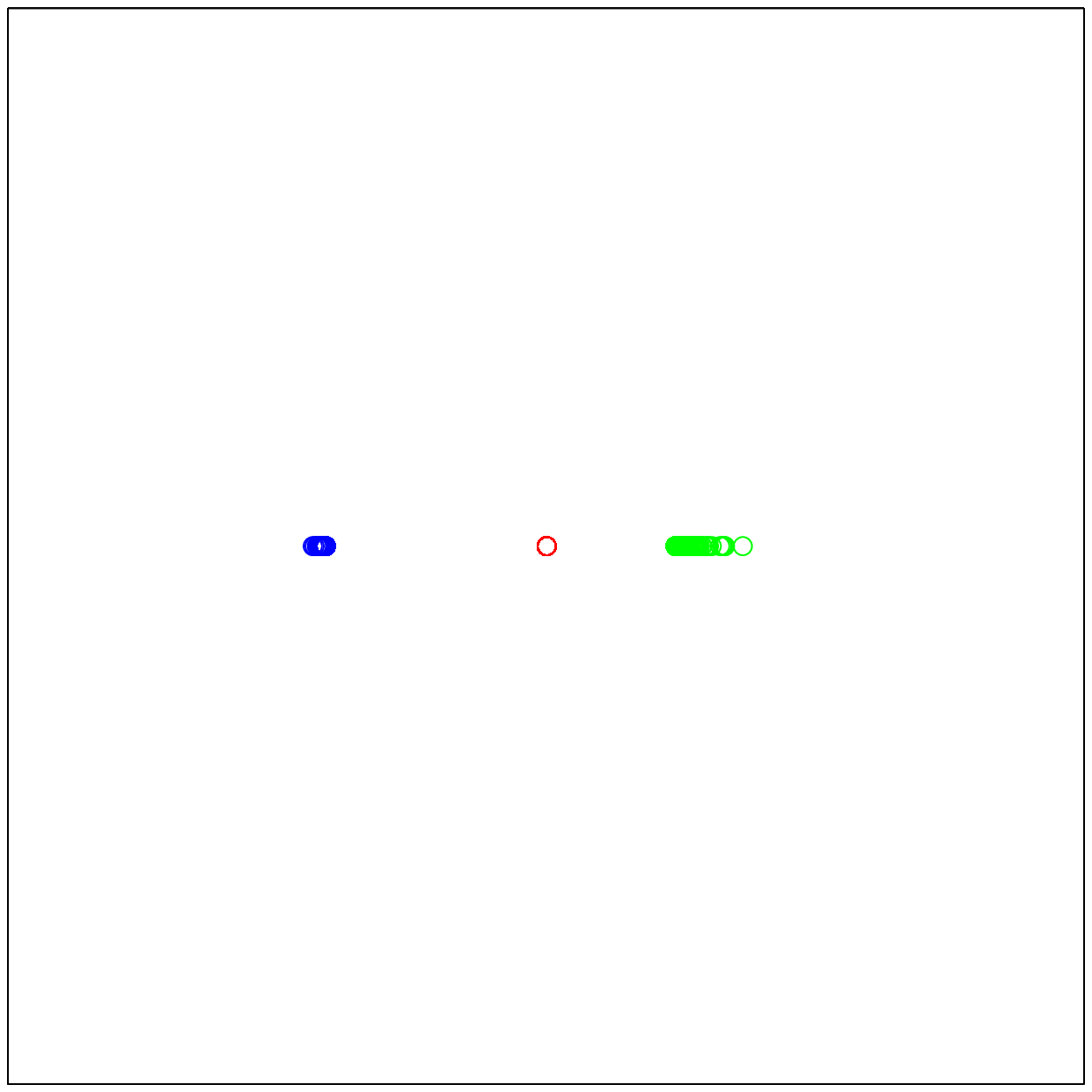} &
      \includegraphics[width=0.089\linewidth]{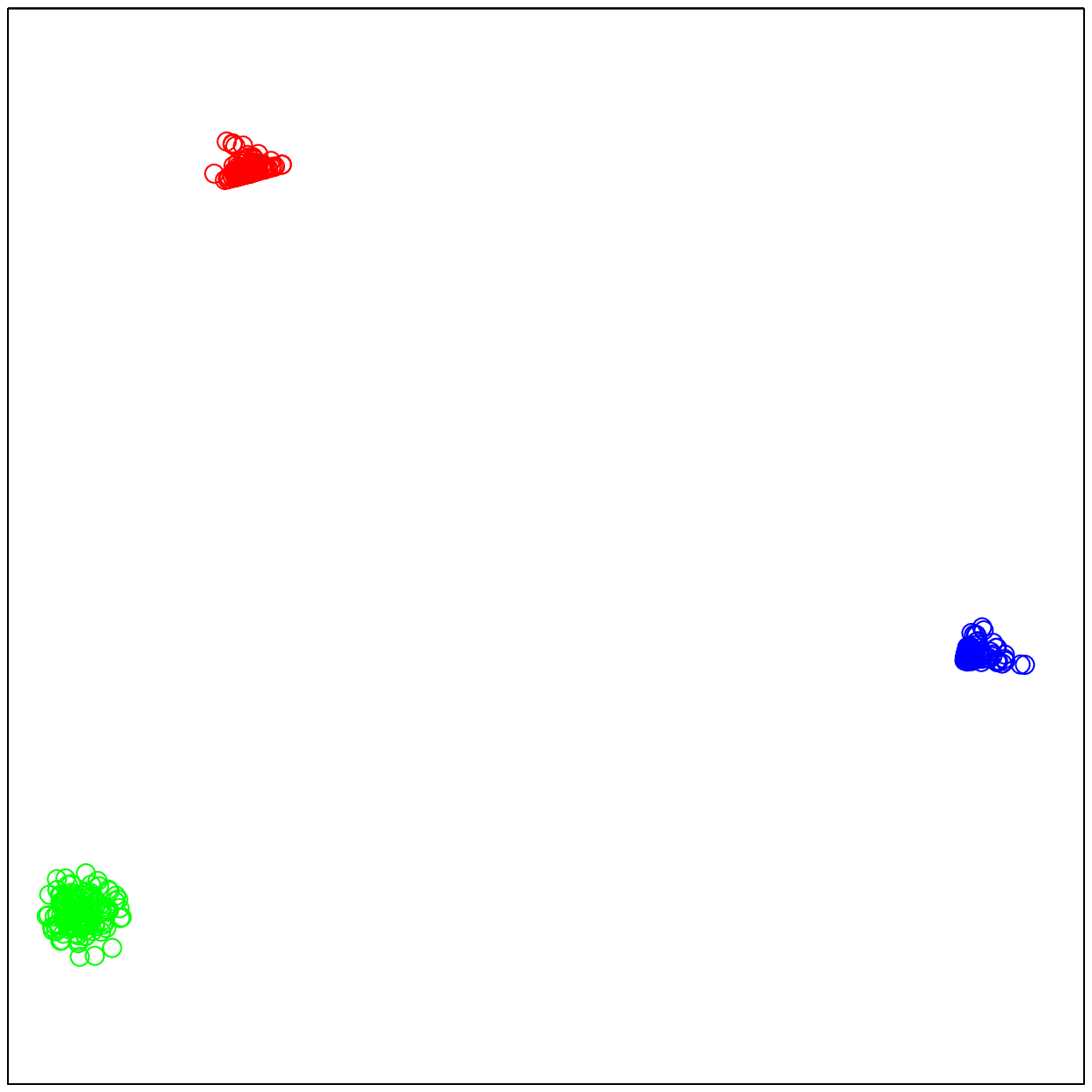} &
      \includegraphics[width=0.089\linewidth]{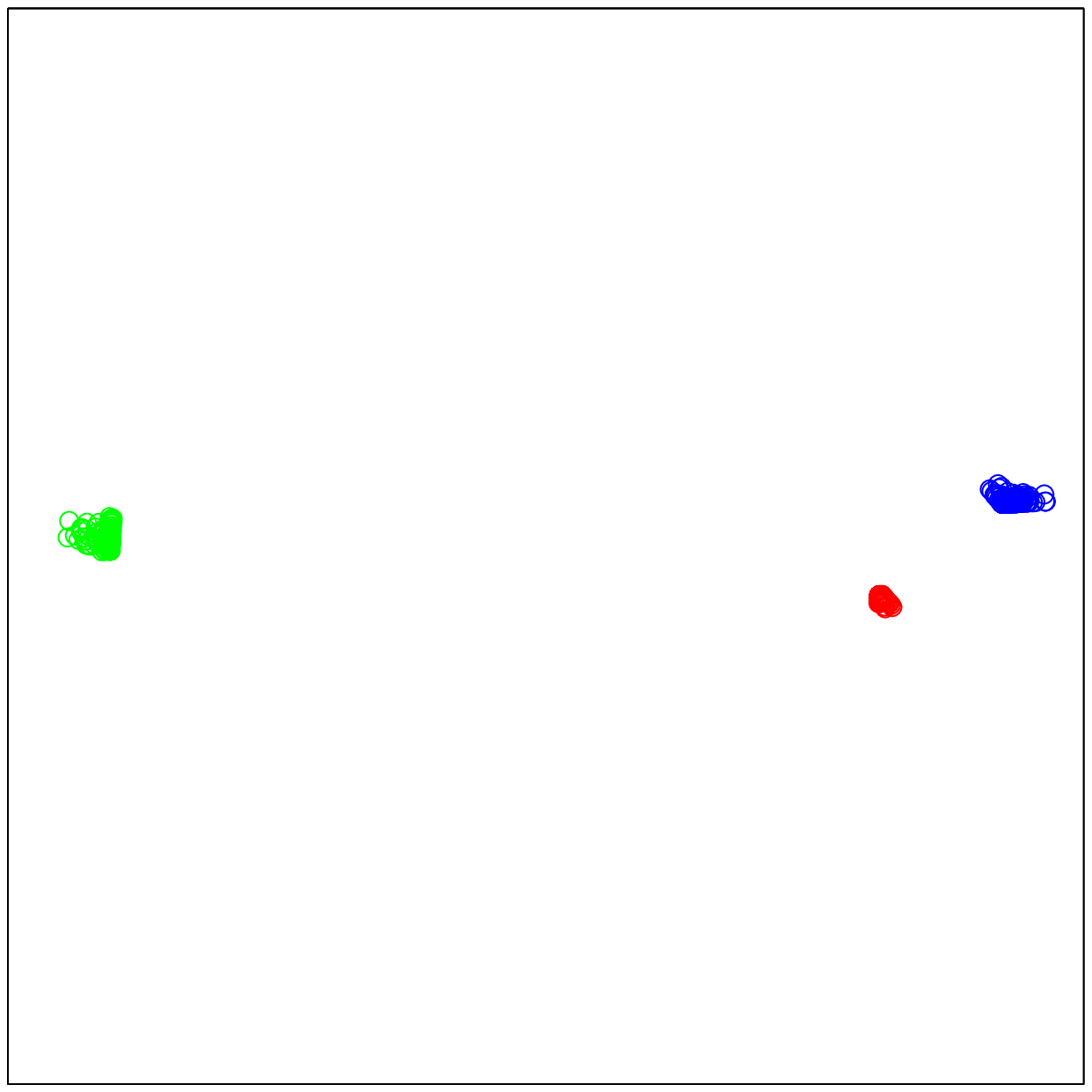} &
      \includegraphics[width=0.089\linewidth]{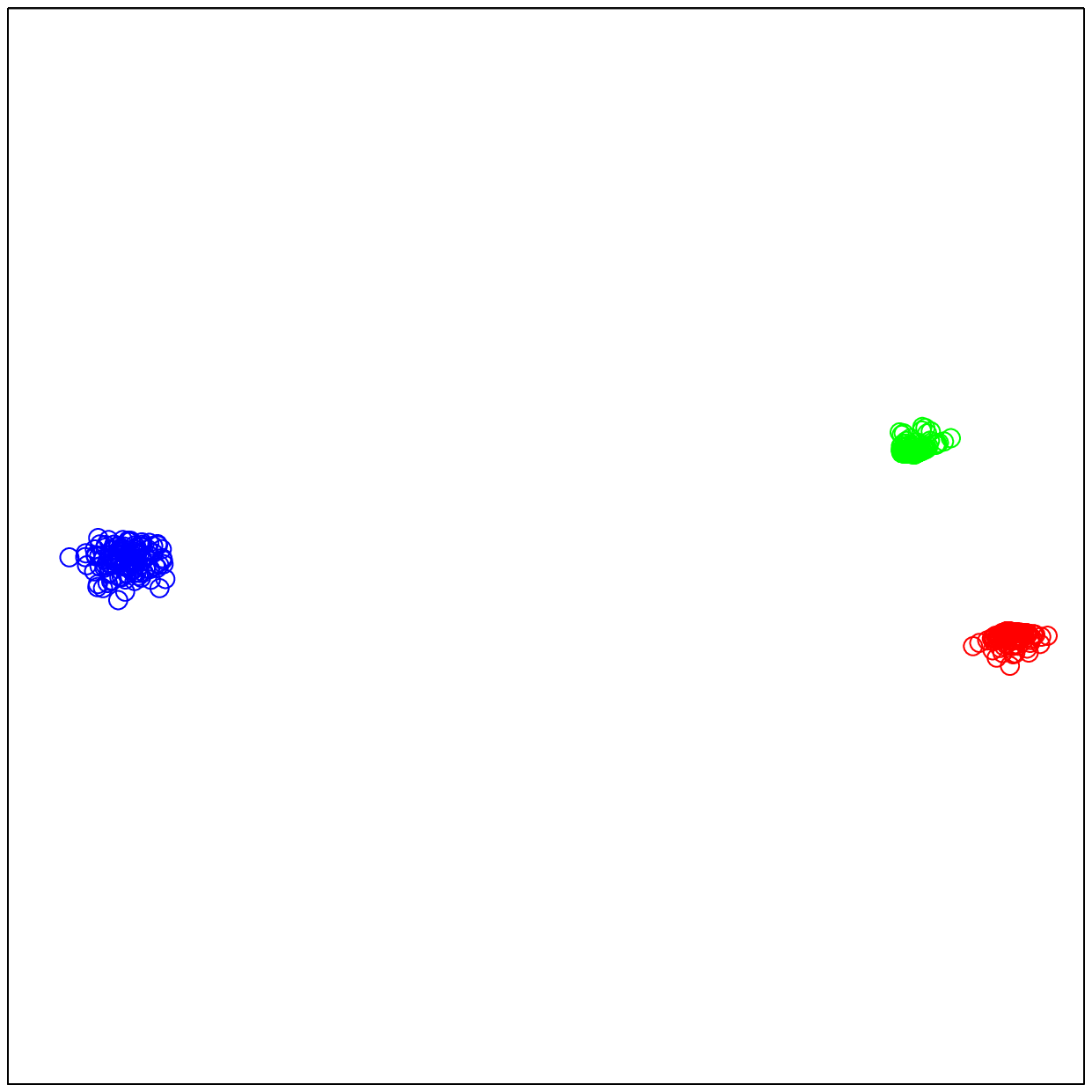} &
      \includegraphics[width=0.089\linewidth]{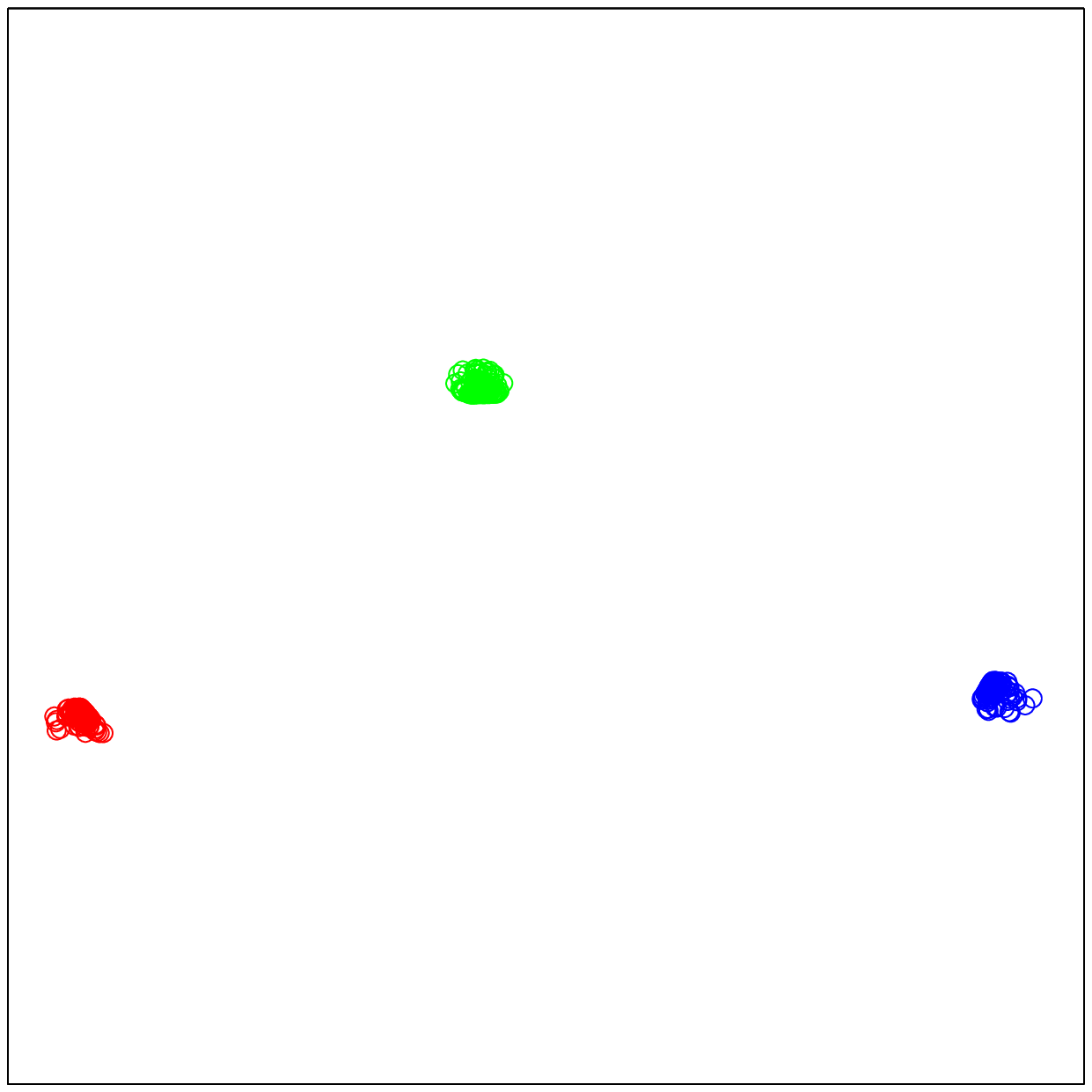} &
      \includegraphics[width=0.089\linewidth]{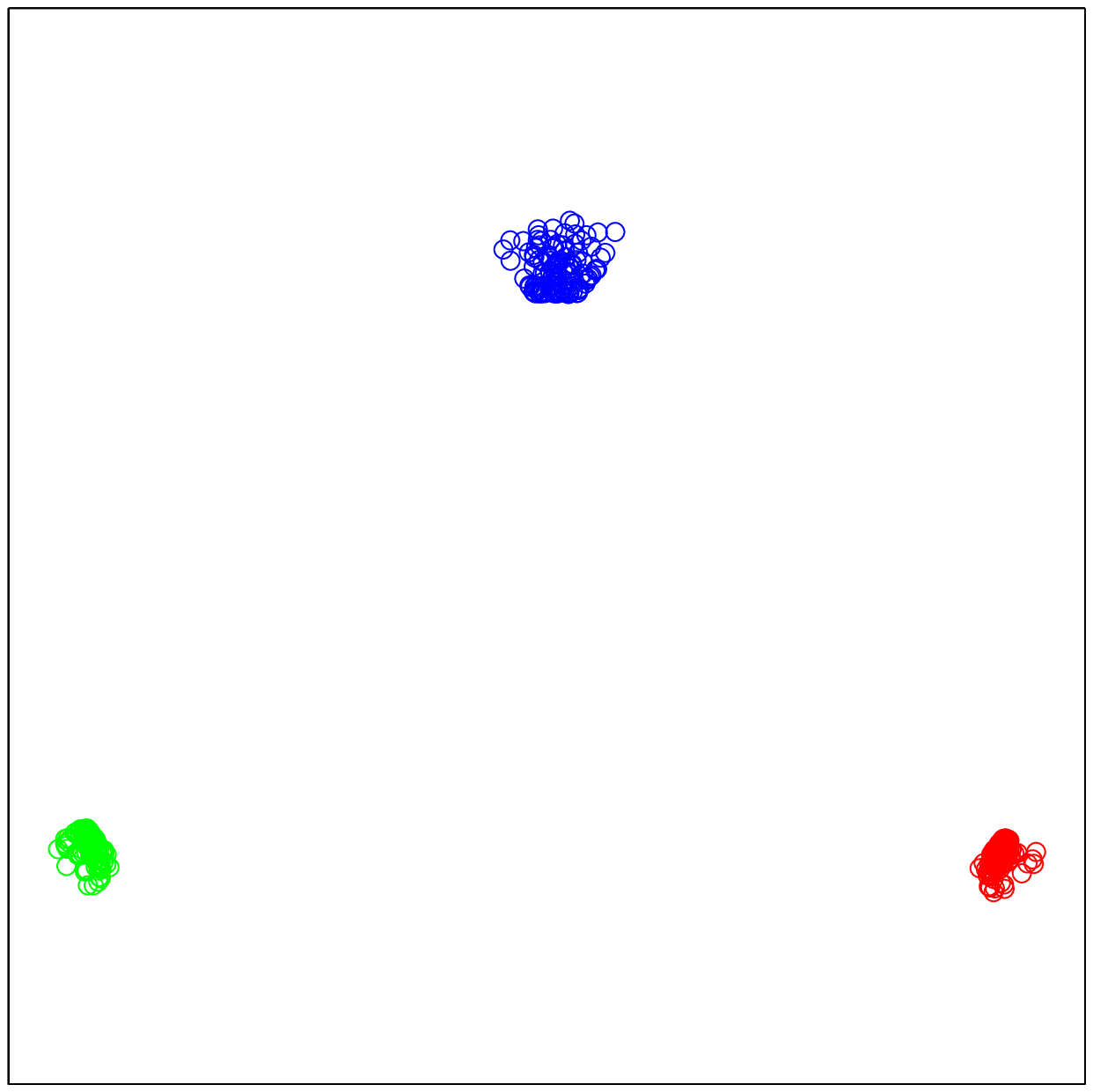} &
      \includegraphics[width=0.089\linewidth]{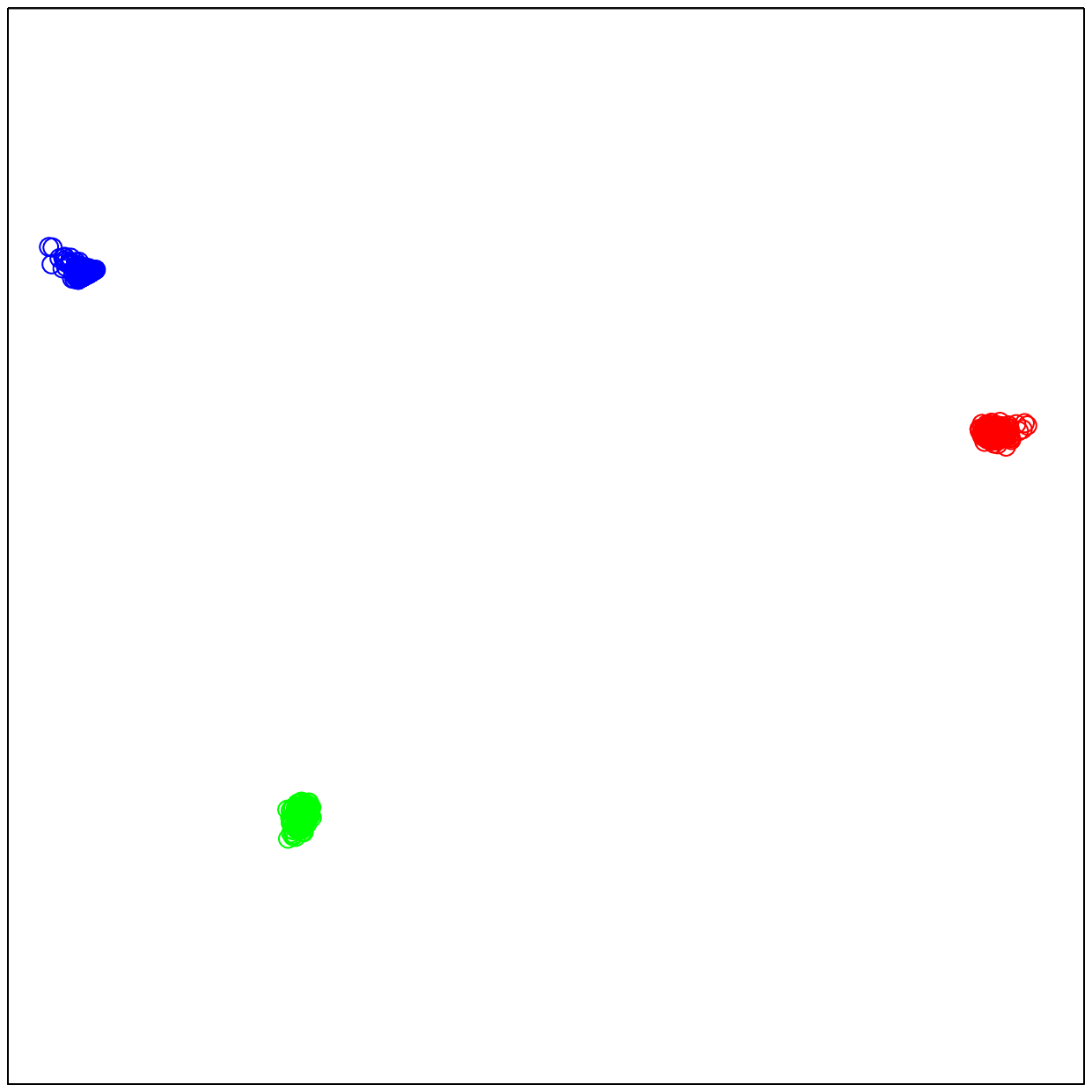} &
      \includegraphics[width=0.089\linewidth]{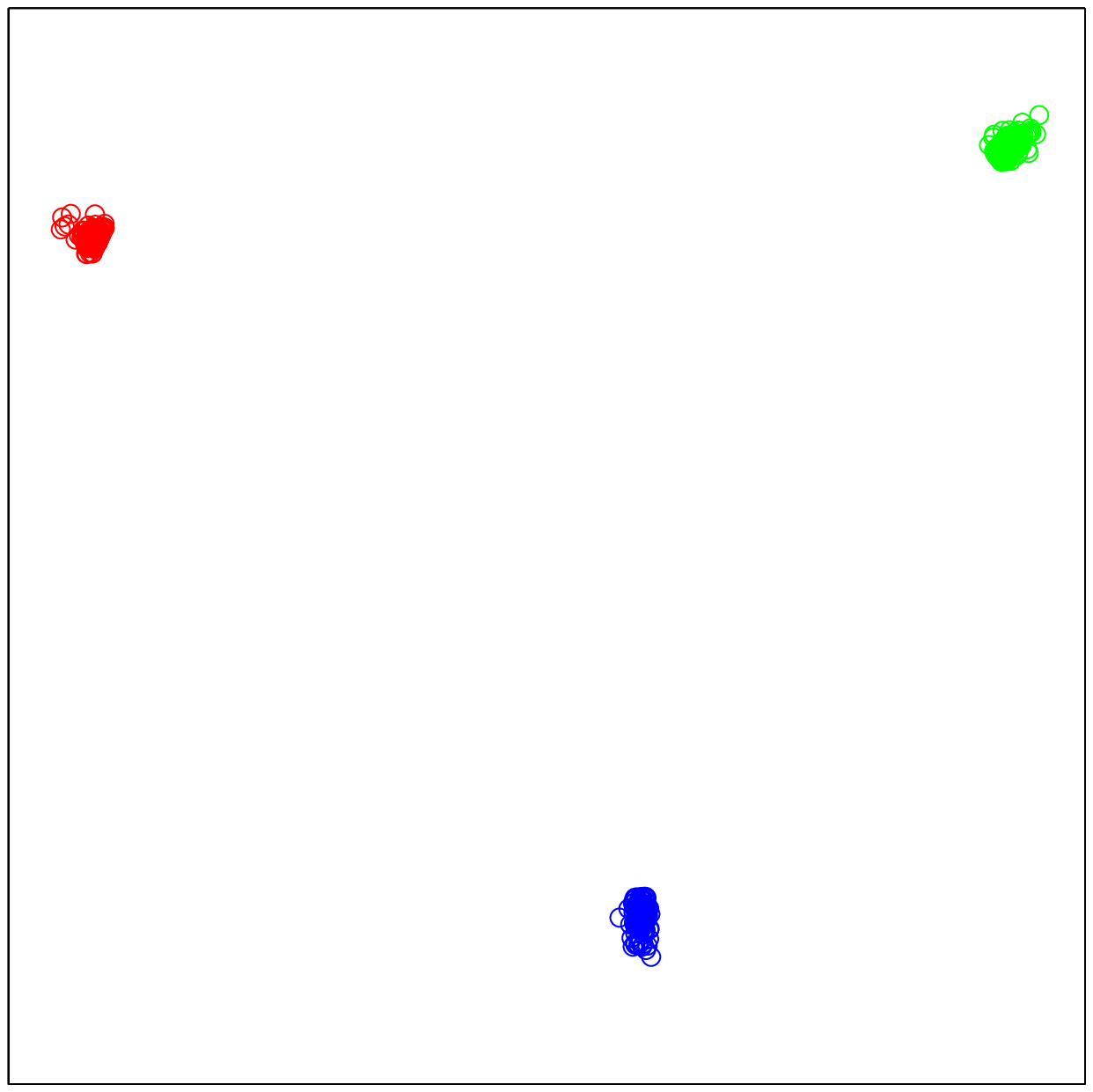} &
      \includegraphics[width=0.089\linewidth]{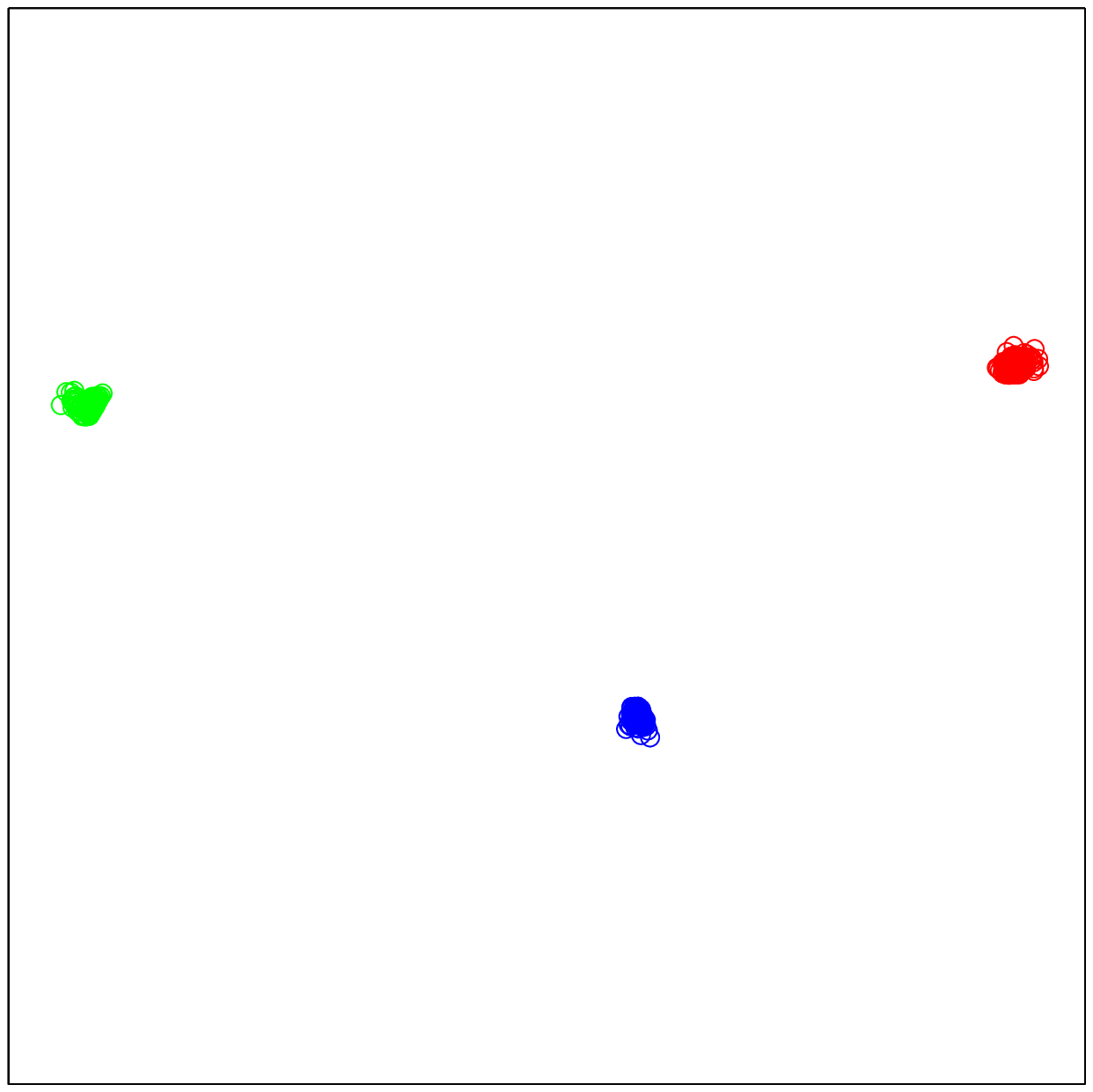} &
      \includegraphics[width=0.089\linewidth]{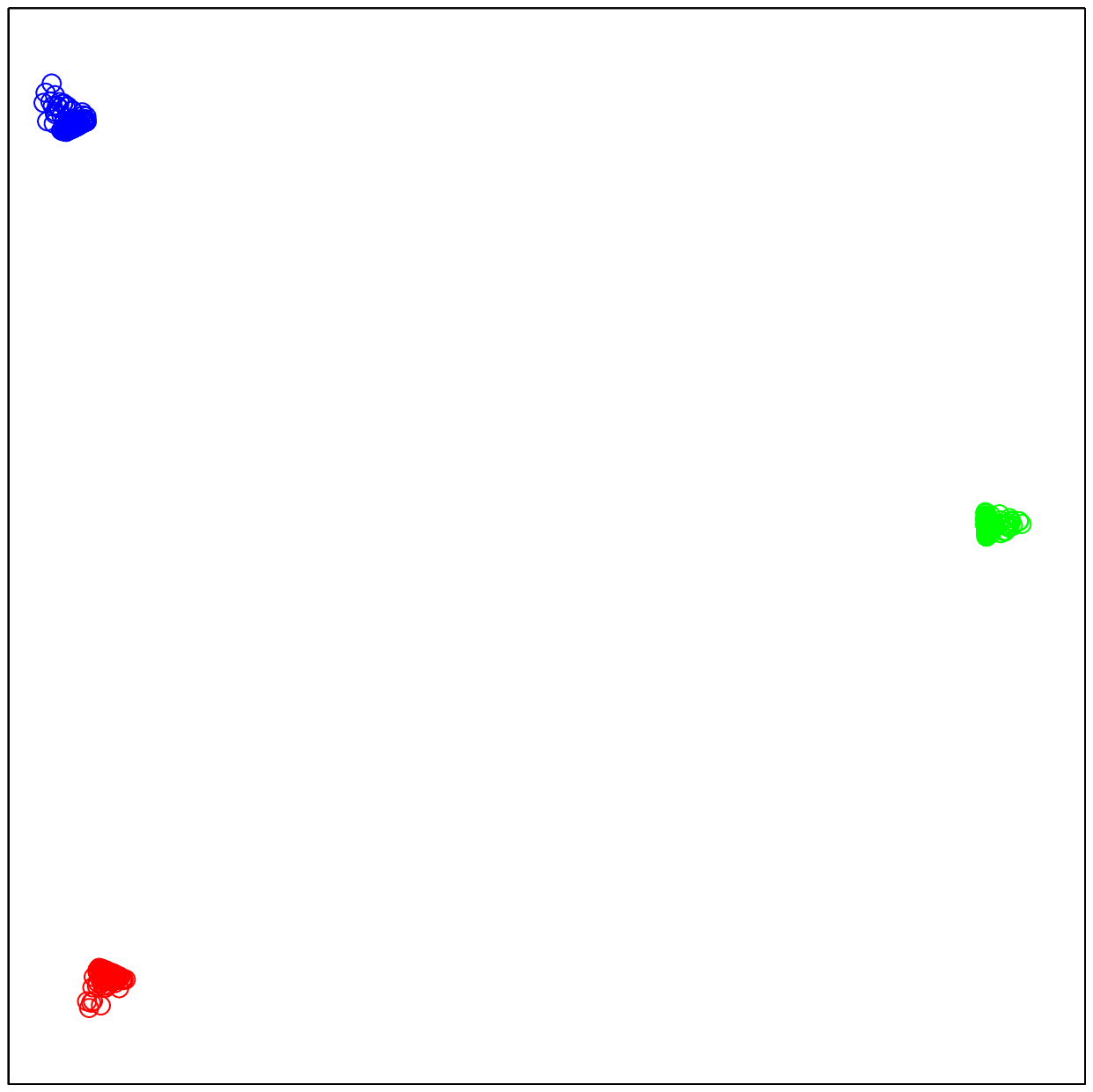} \\[-.6ex]
      \rotatebox{90}{\hspace{1.5ex}$K=4$} &
      \includegraphics[width=0.089\linewidth]{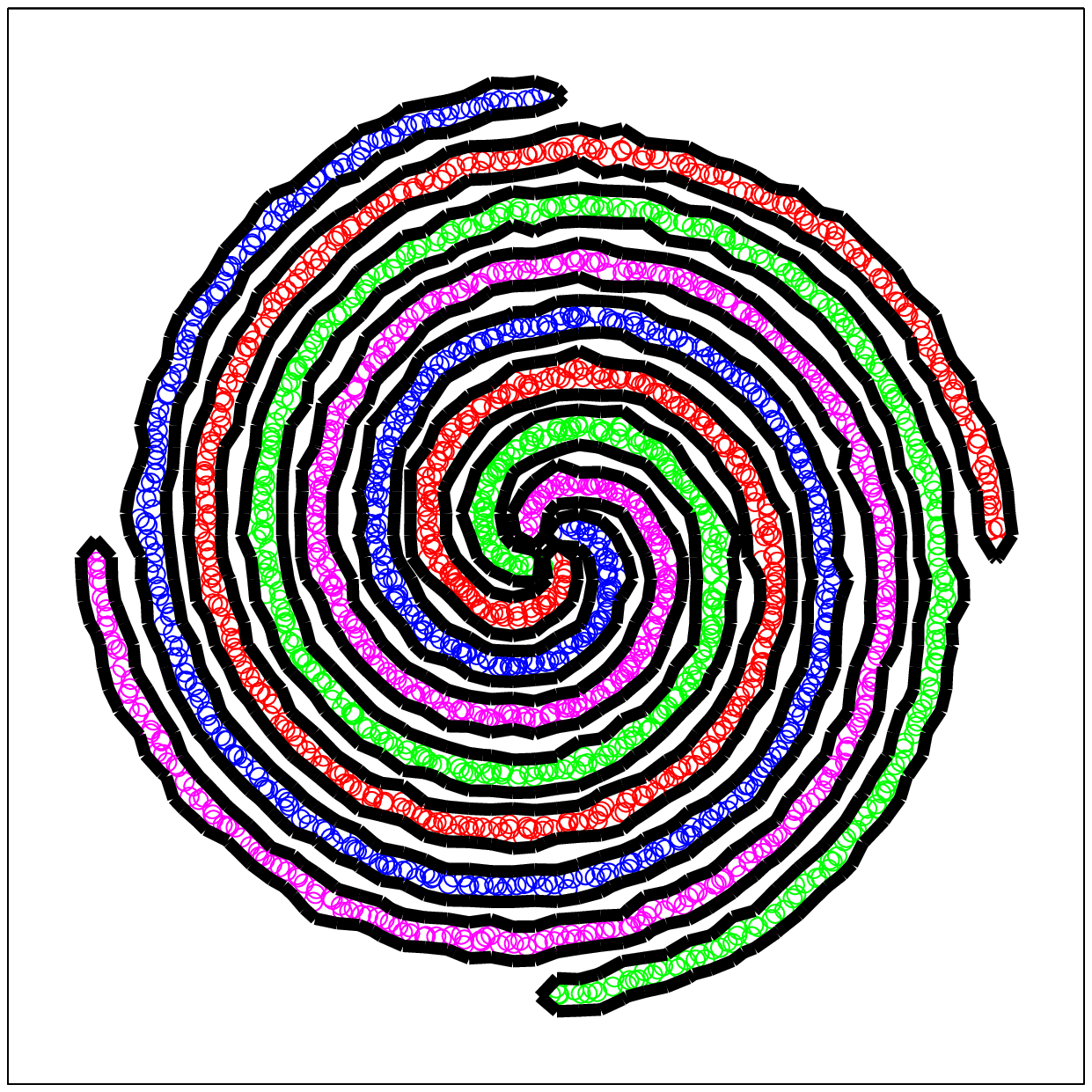} &
      \includegraphics[width=0.089\linewidth]{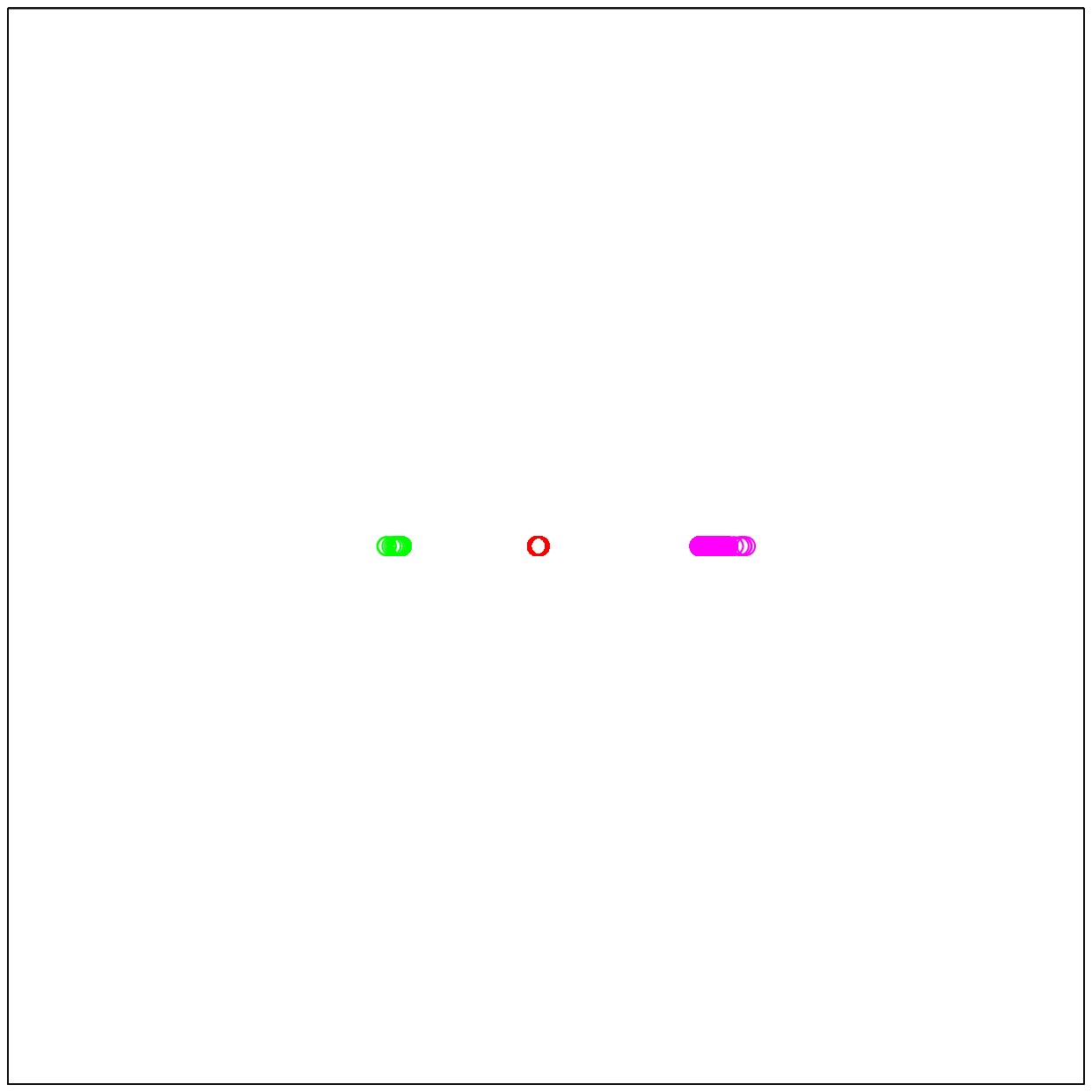} &
      \includegraphics[width=0.089\linewidth]{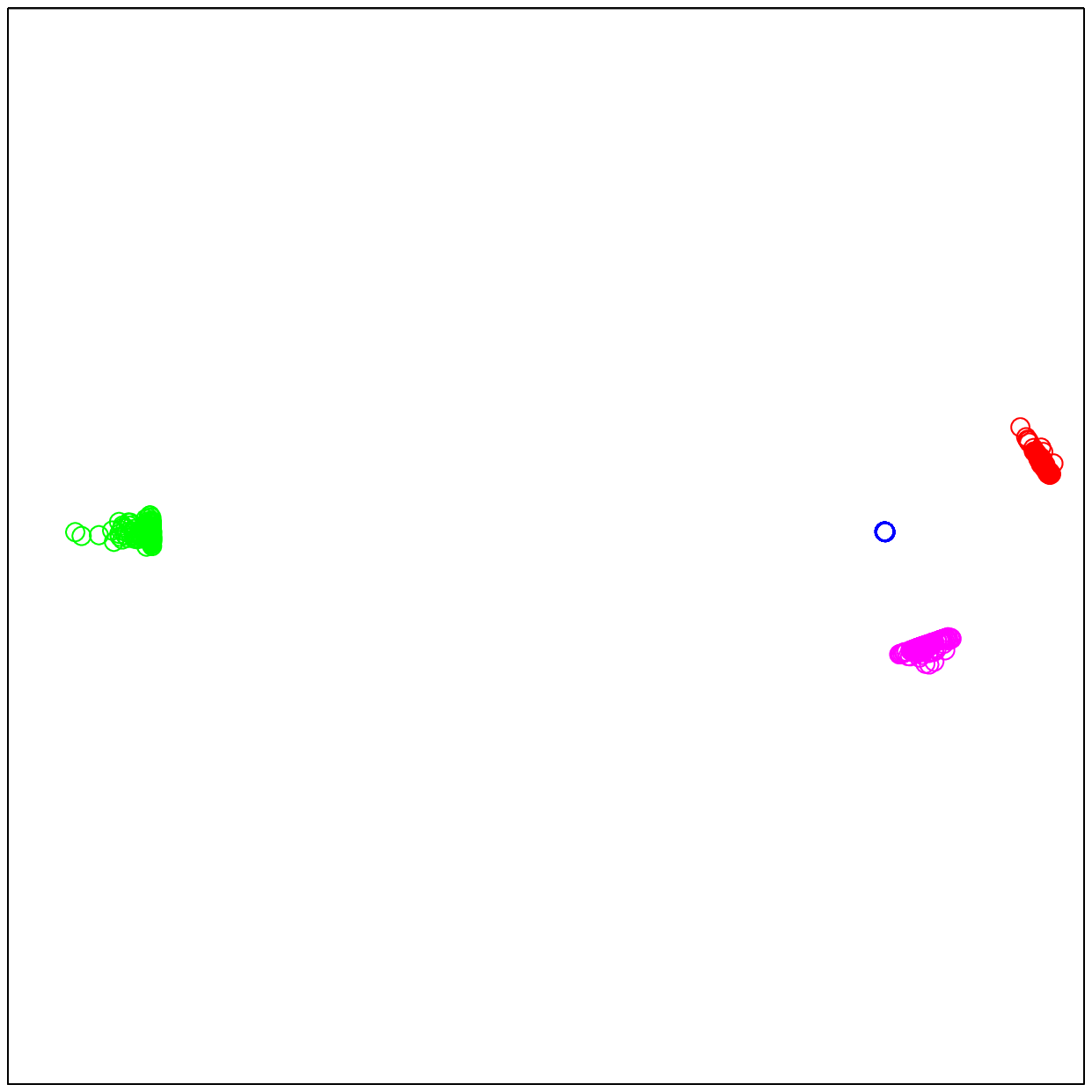} &
      \includegraphics[width=0.089\linewidth]{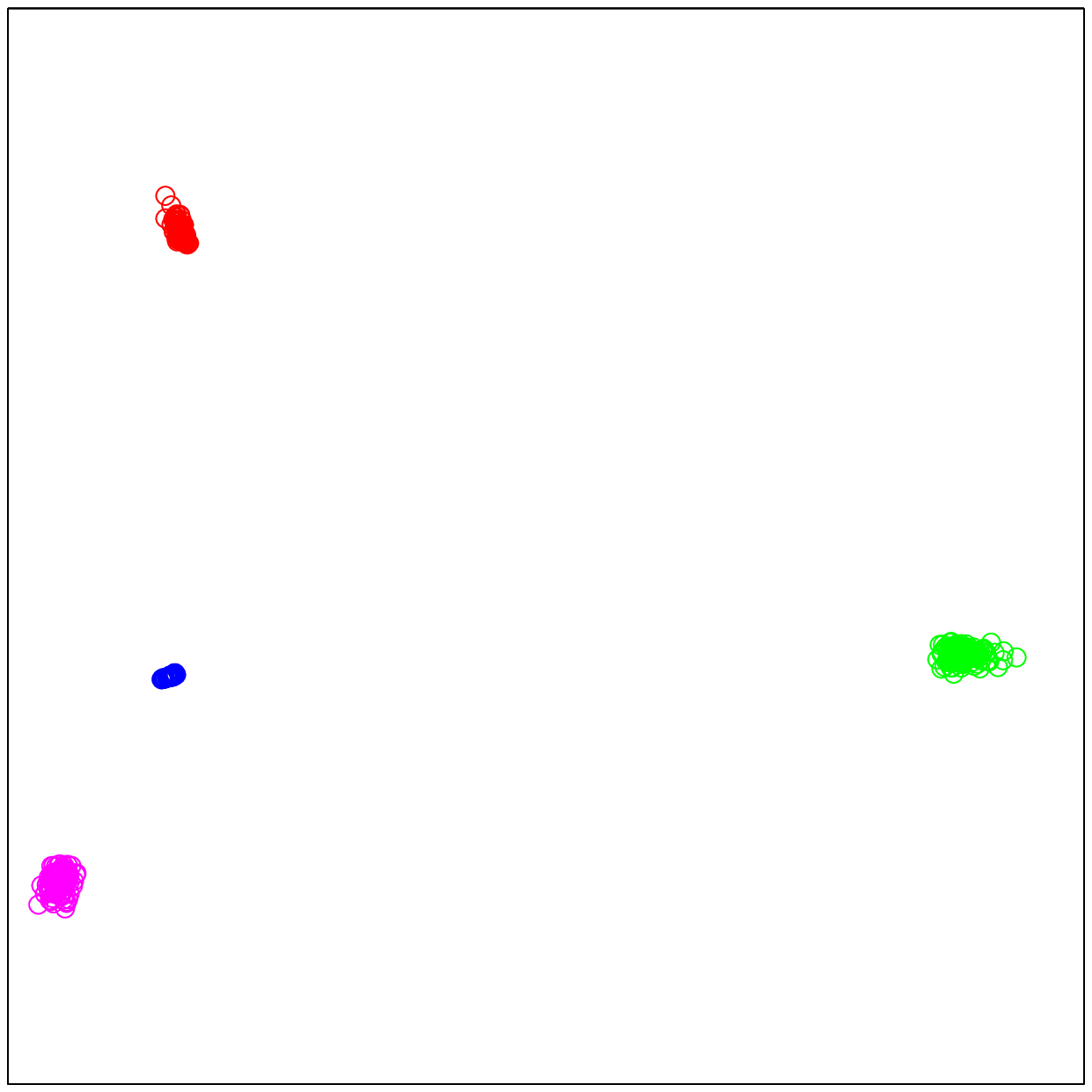} &
      \includegraphics[width=0.089\linewidth]{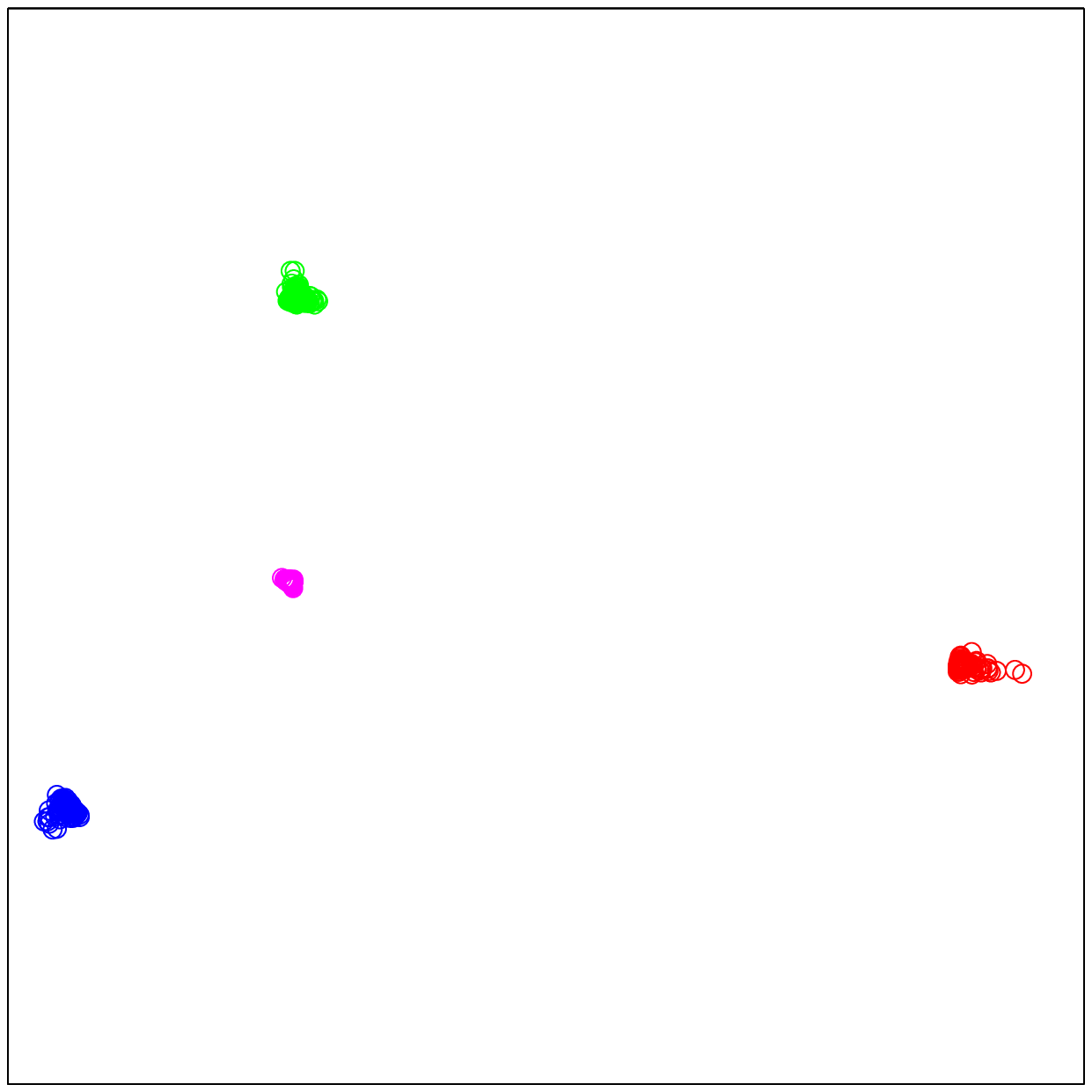} &
      \includegraphics[width=0.089\linewidth]{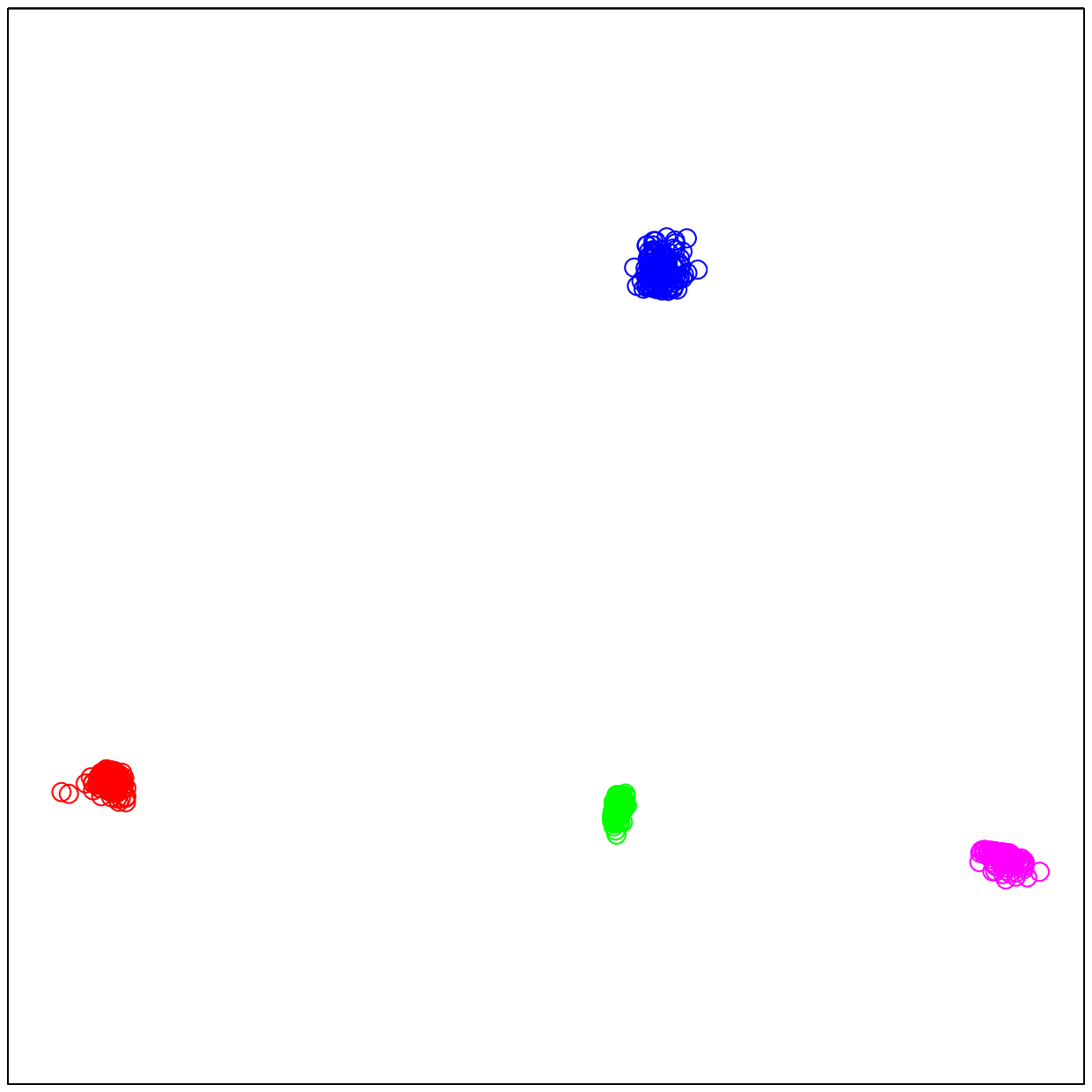} &
      \includegraphics[width=0.089\linewidth]{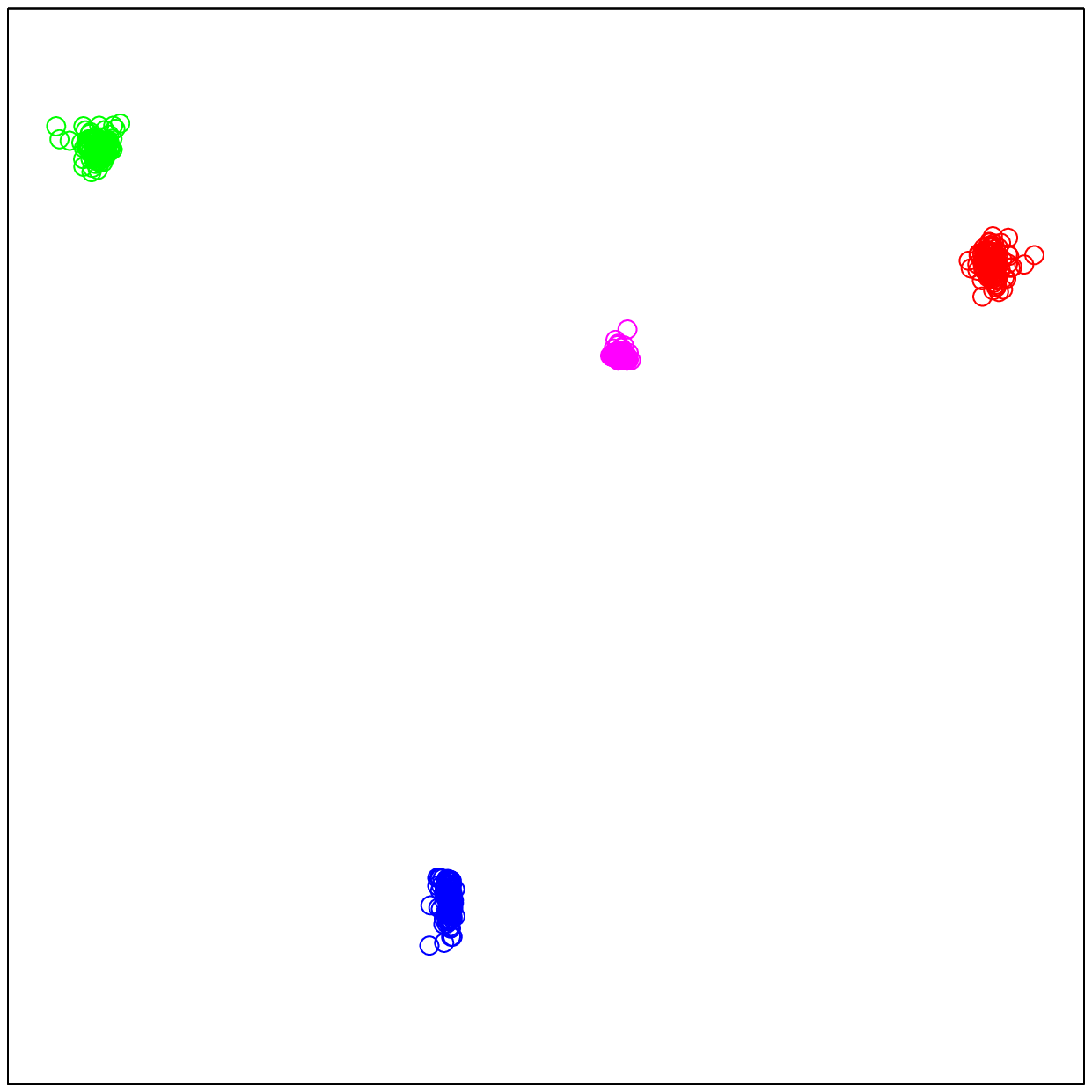} &
      \includegraphics[width=0.089\linewidth]{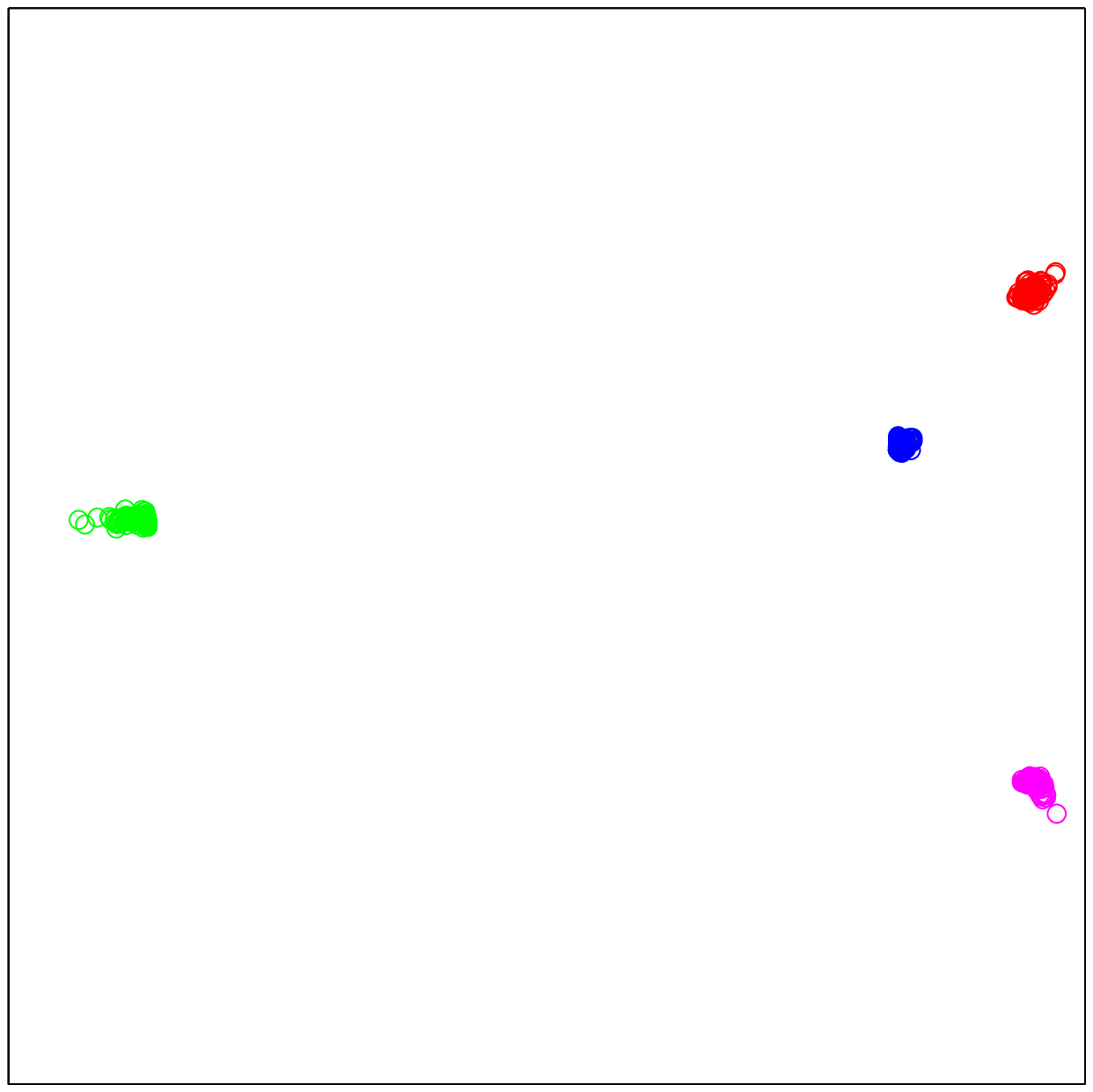} &
      \includegraphics[width=0.089\linewidth]{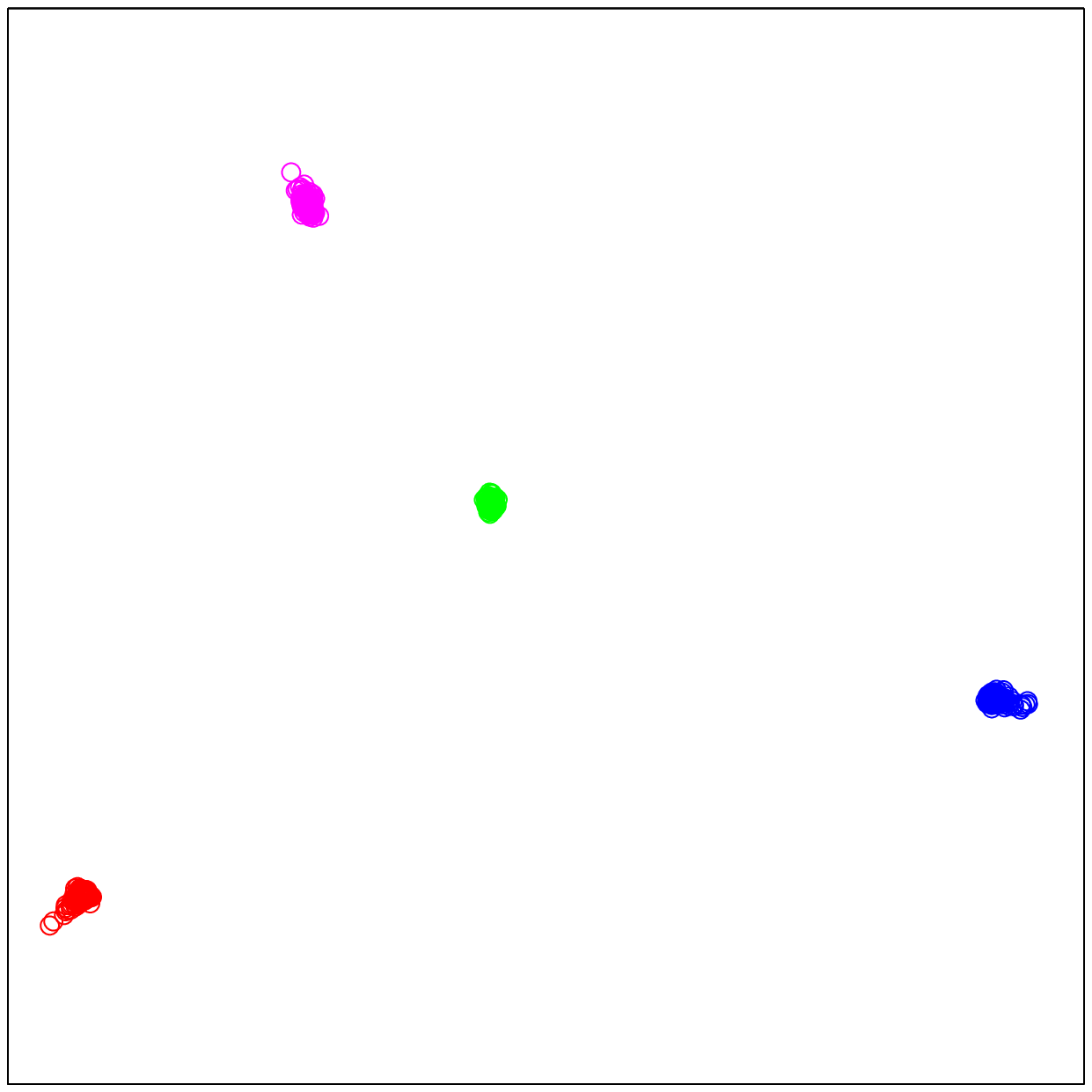} &
      \includegraphics[width=0.089\linewidth]{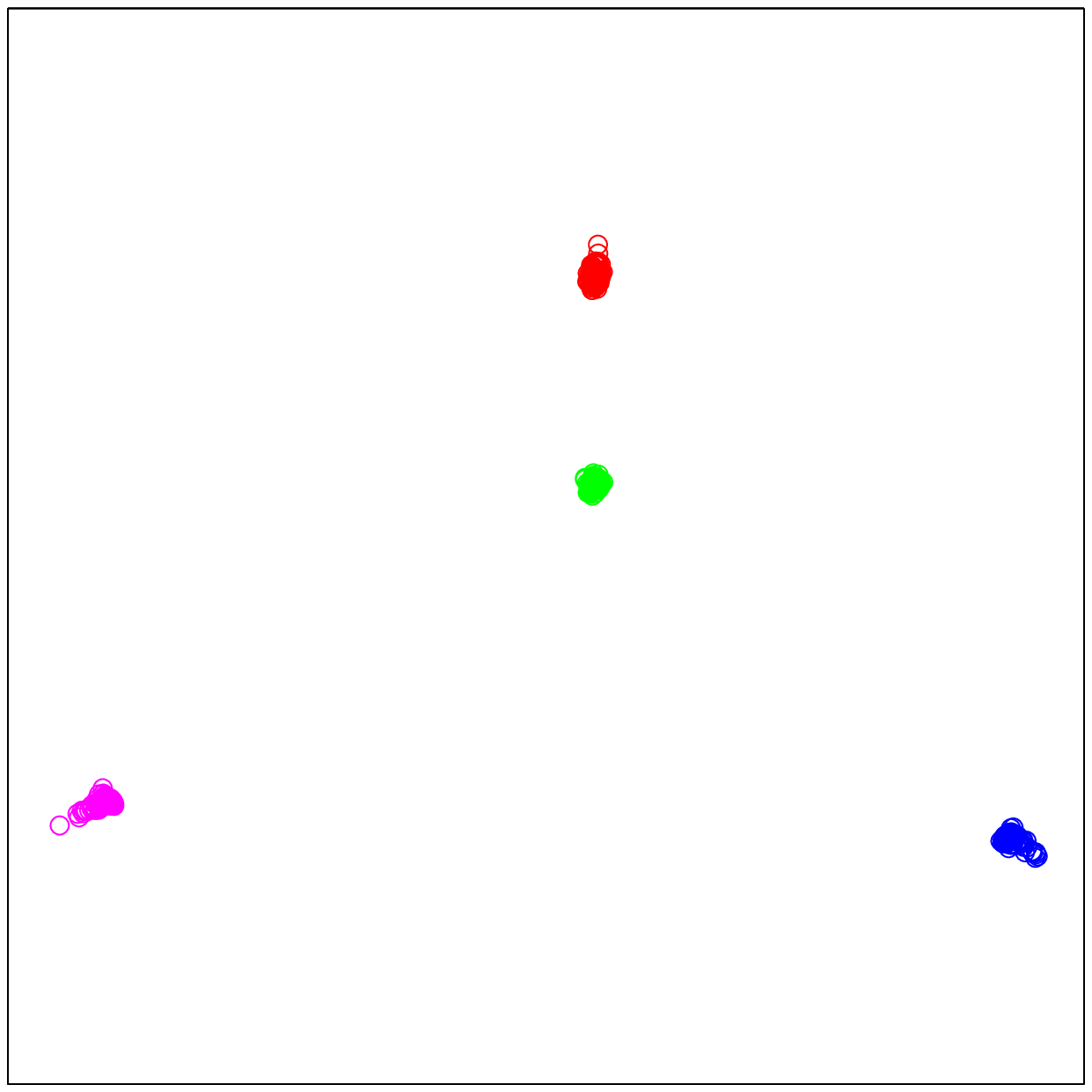} &
      \includegraphics[width=0.089\linewidth]{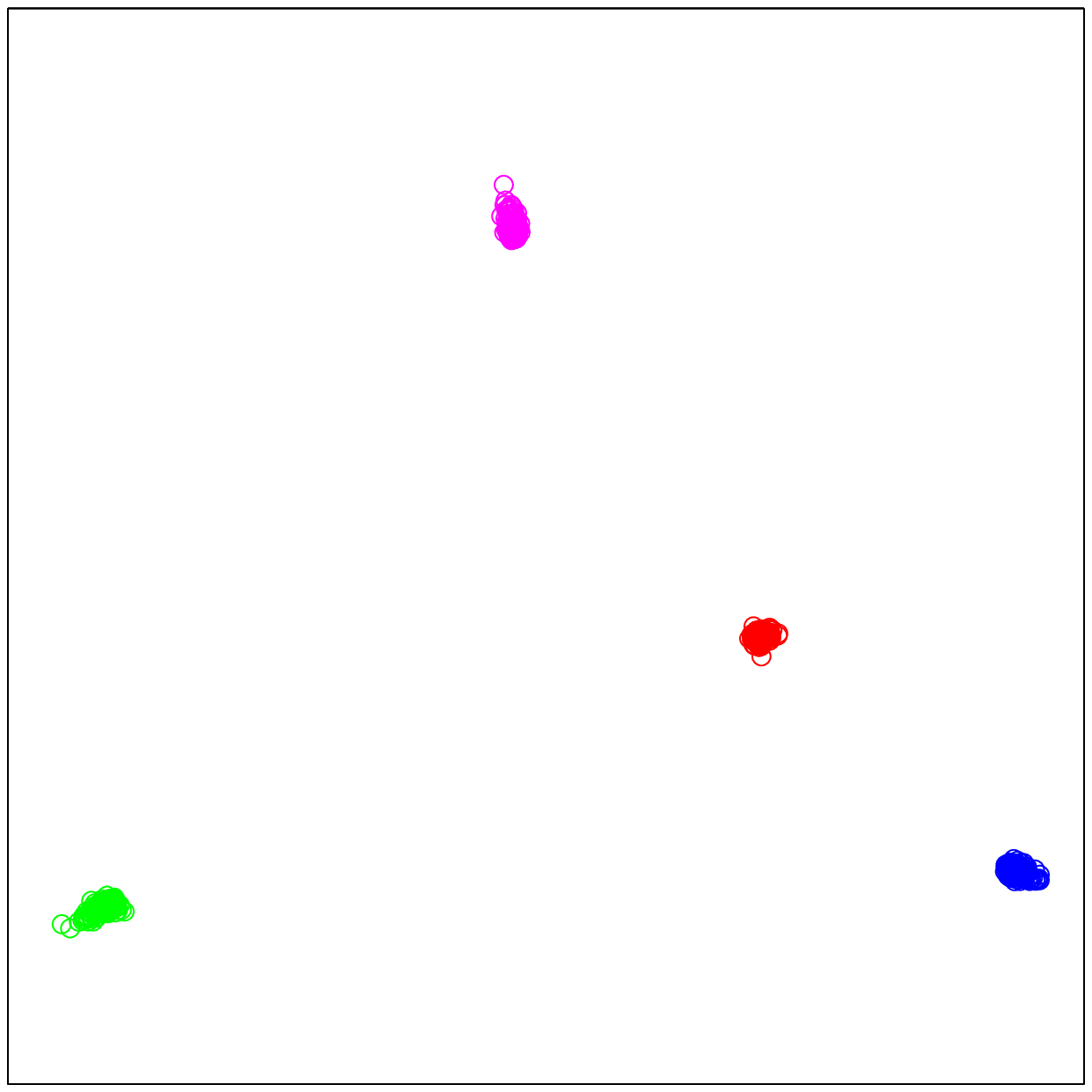} \\[-.6ex]
      \rotatebox{90}{\hspace{1.5ex}$K=5$} &
      \includegraphics[width=0.089\linewidth]{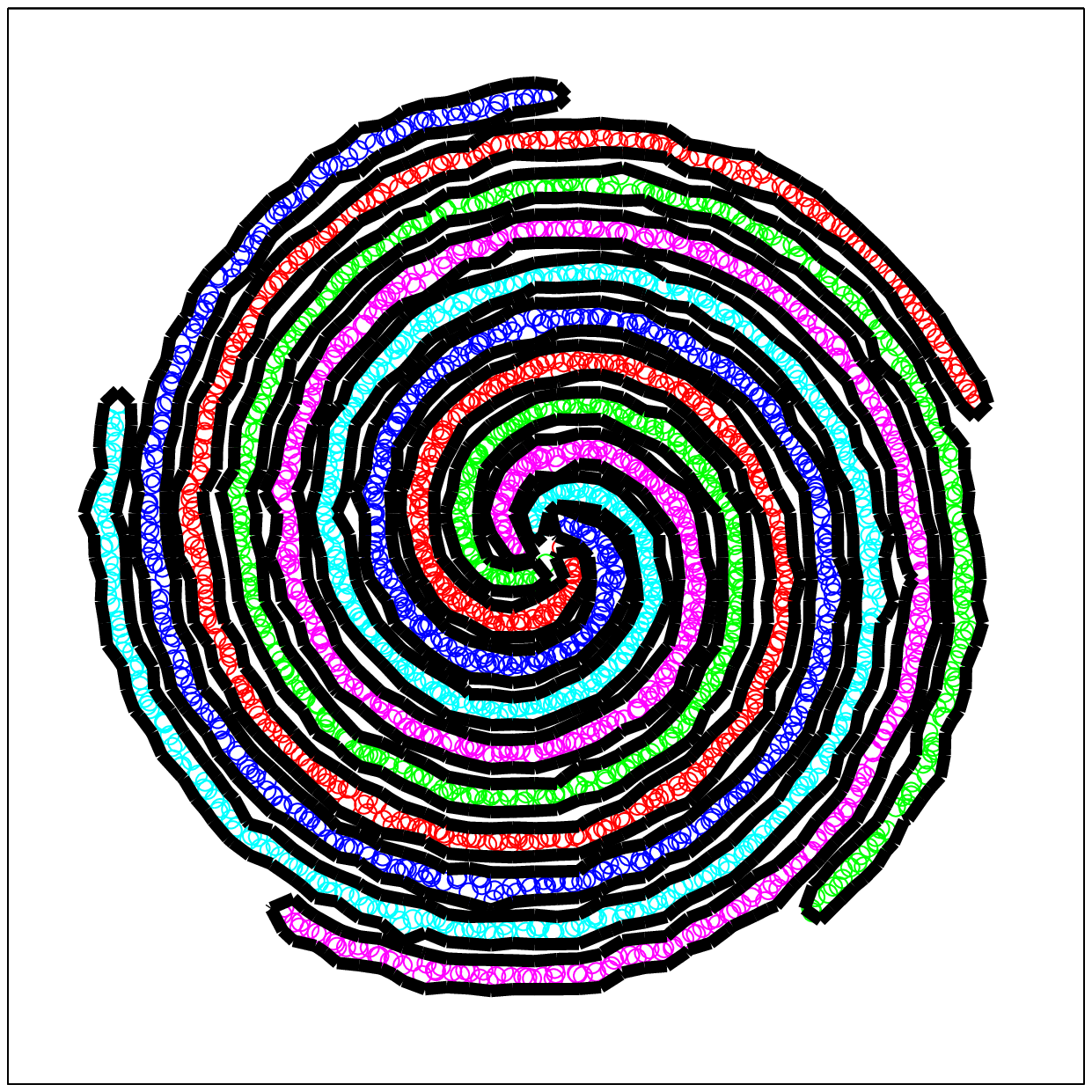} &
      \includegraphics[width=0.089\linewidth]{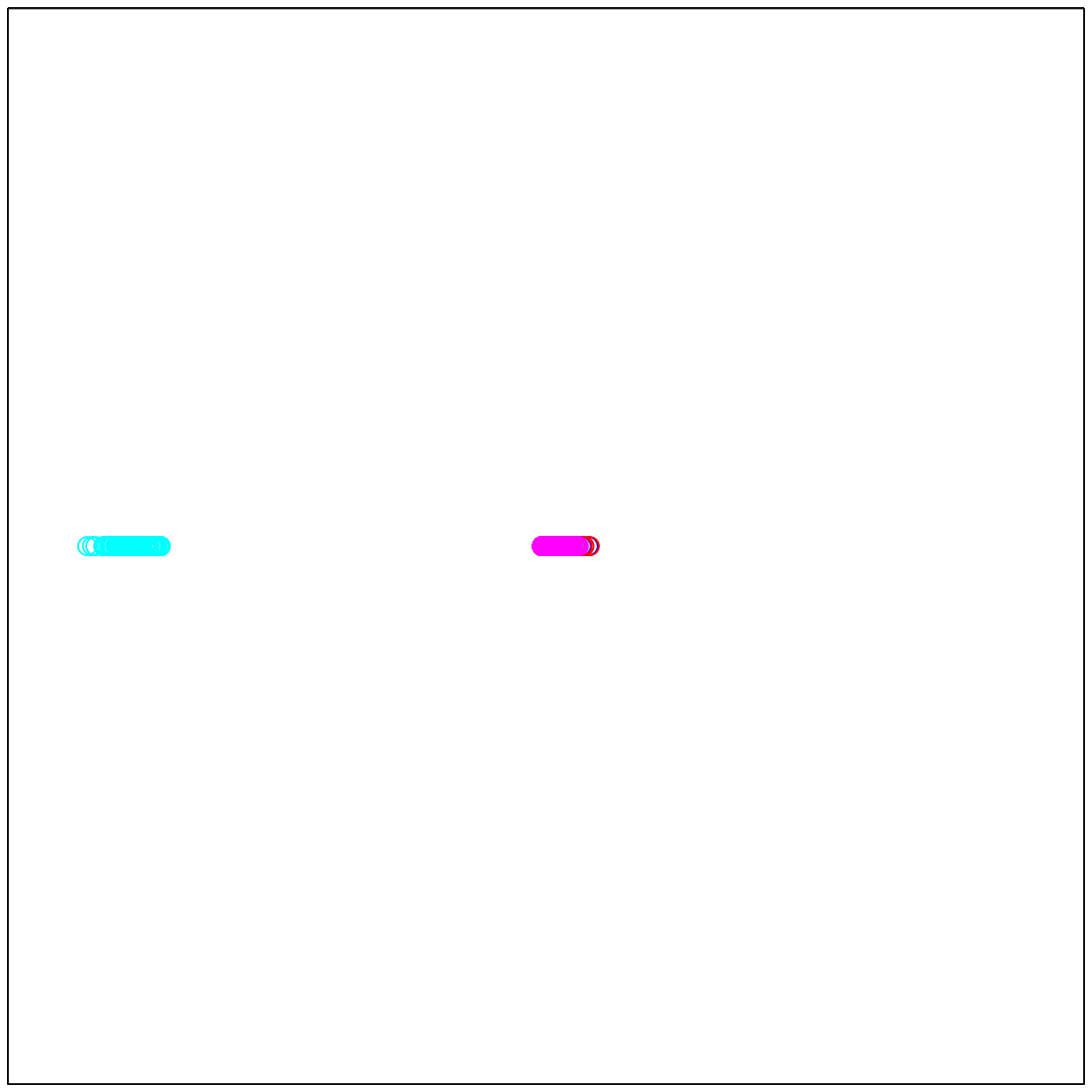} &
      \includegraphics[width=0.089\linewidth]{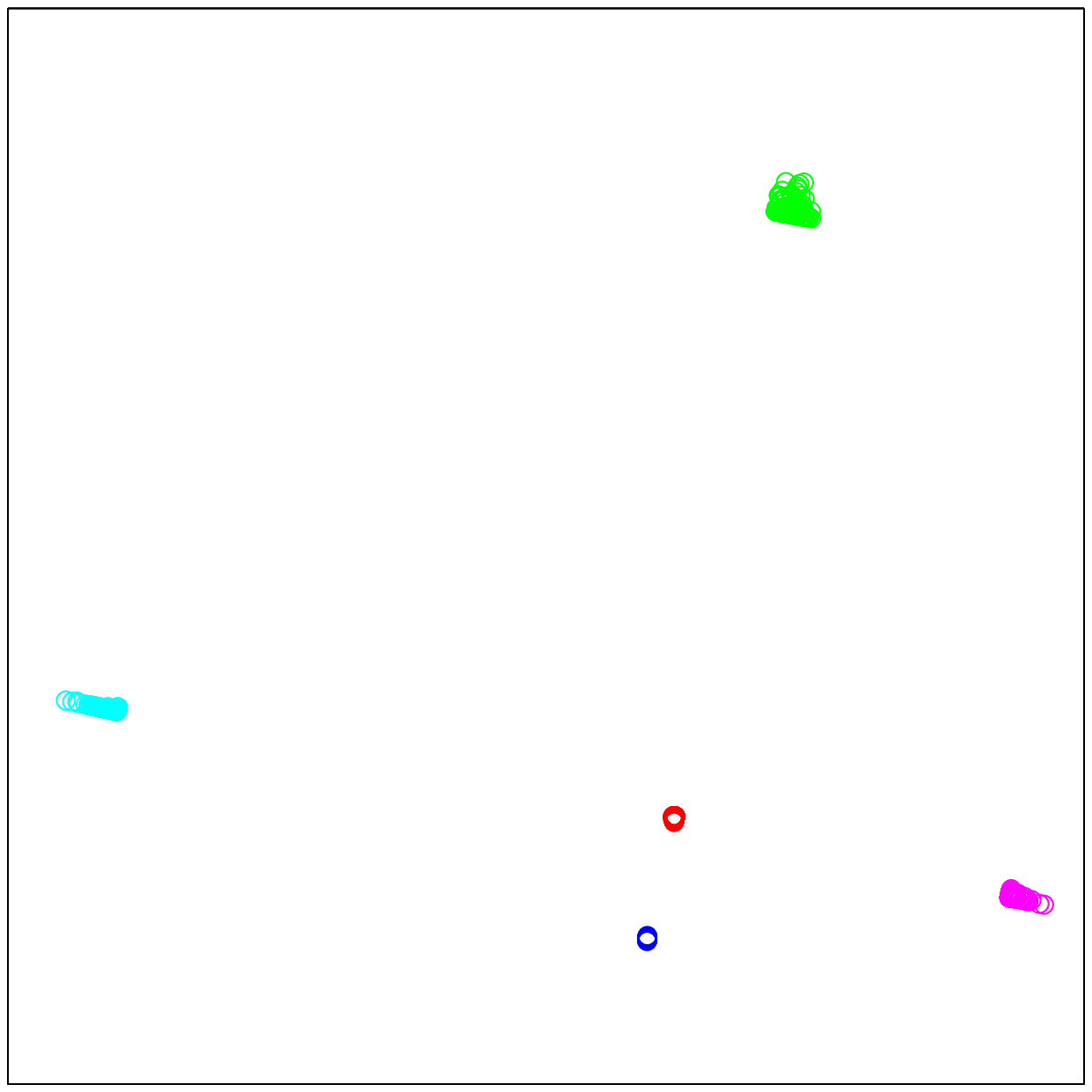} &
      \includegraphics[width=0.089\linewidth]{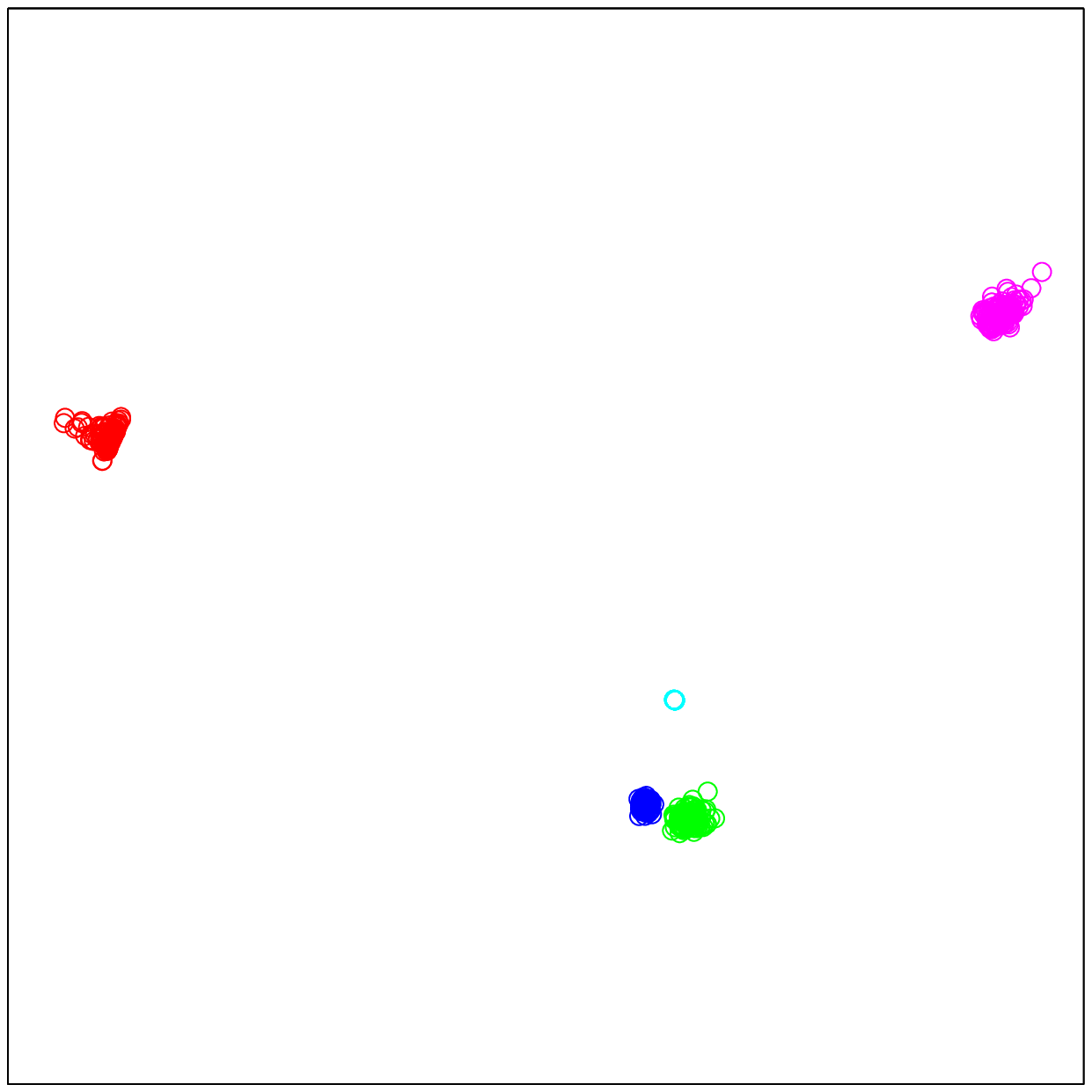} &
      \includegraphics[width=0.089\linewidth]{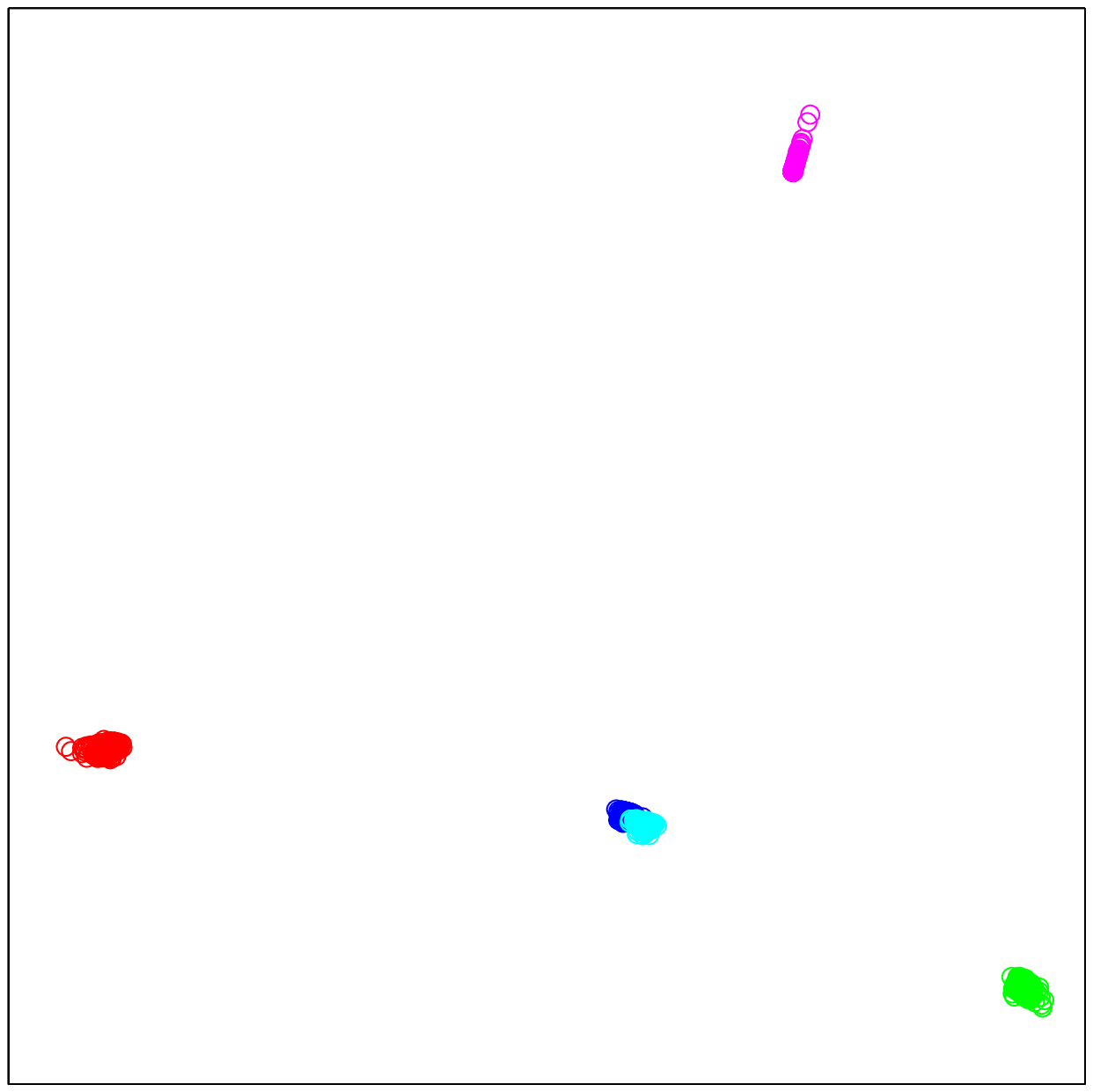} &
      \includegraphics[width=0.089\linewidth]{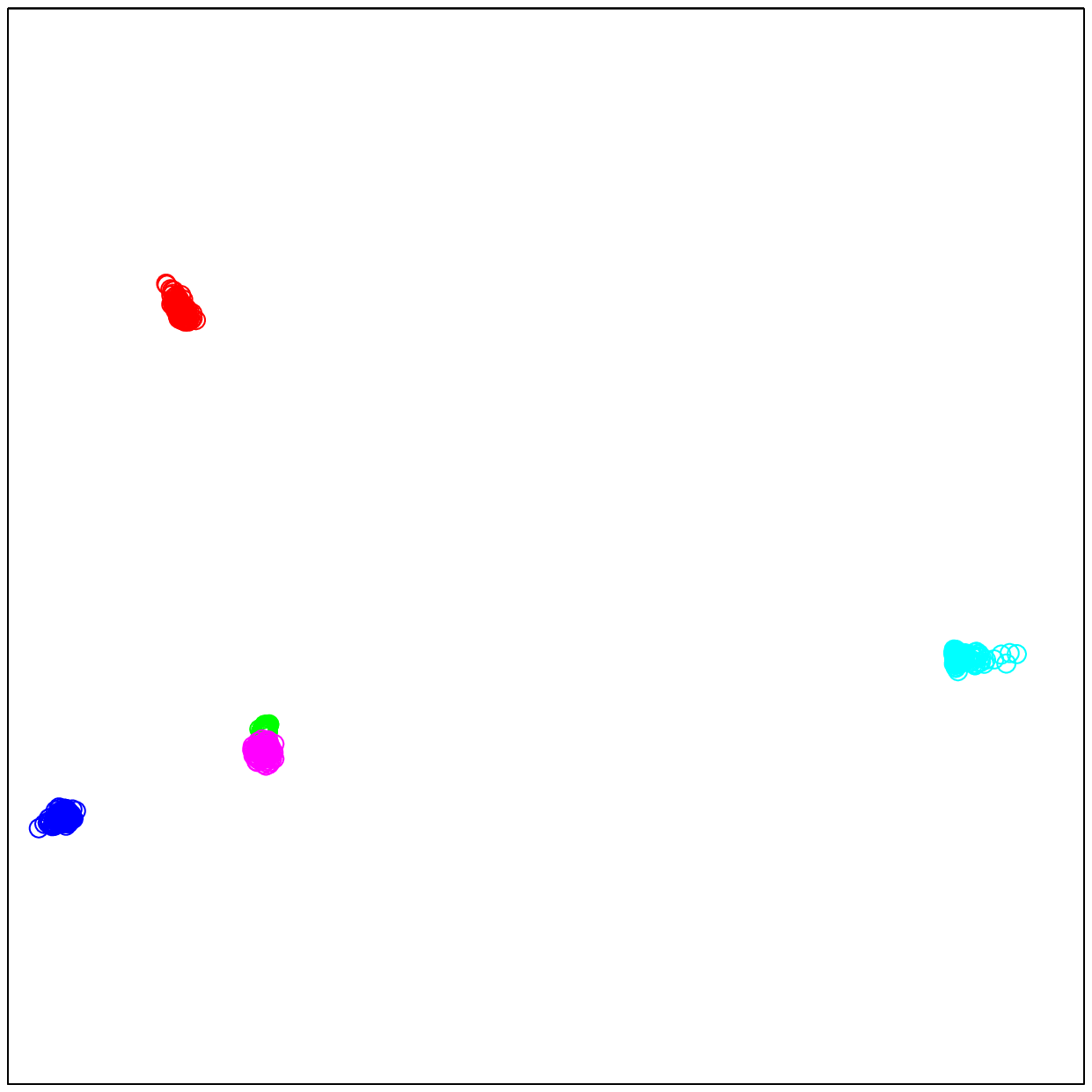} &
      \includegraphics[width=0.089\linewidth]{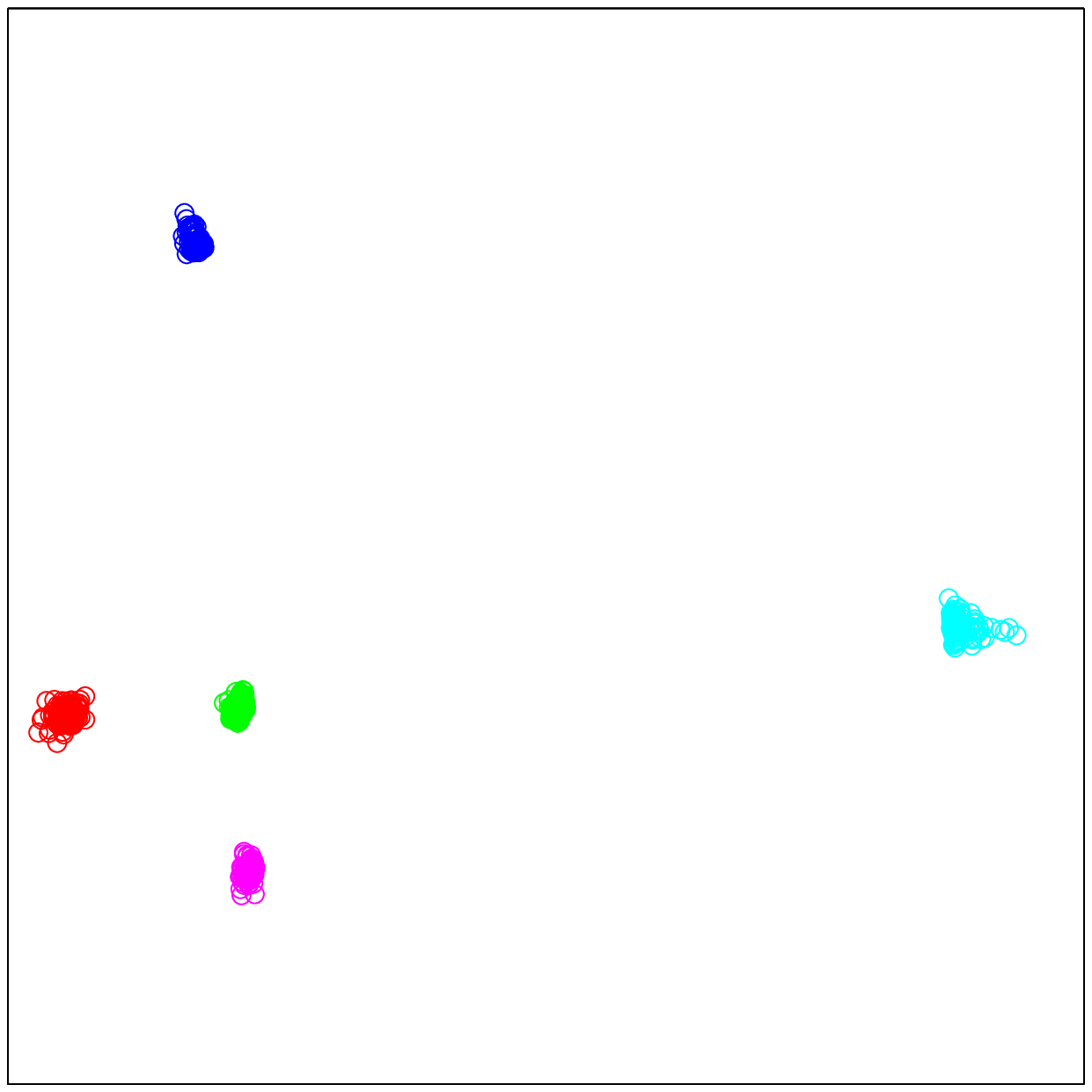} &
      \includegraphics[width=0.089\linewidth]{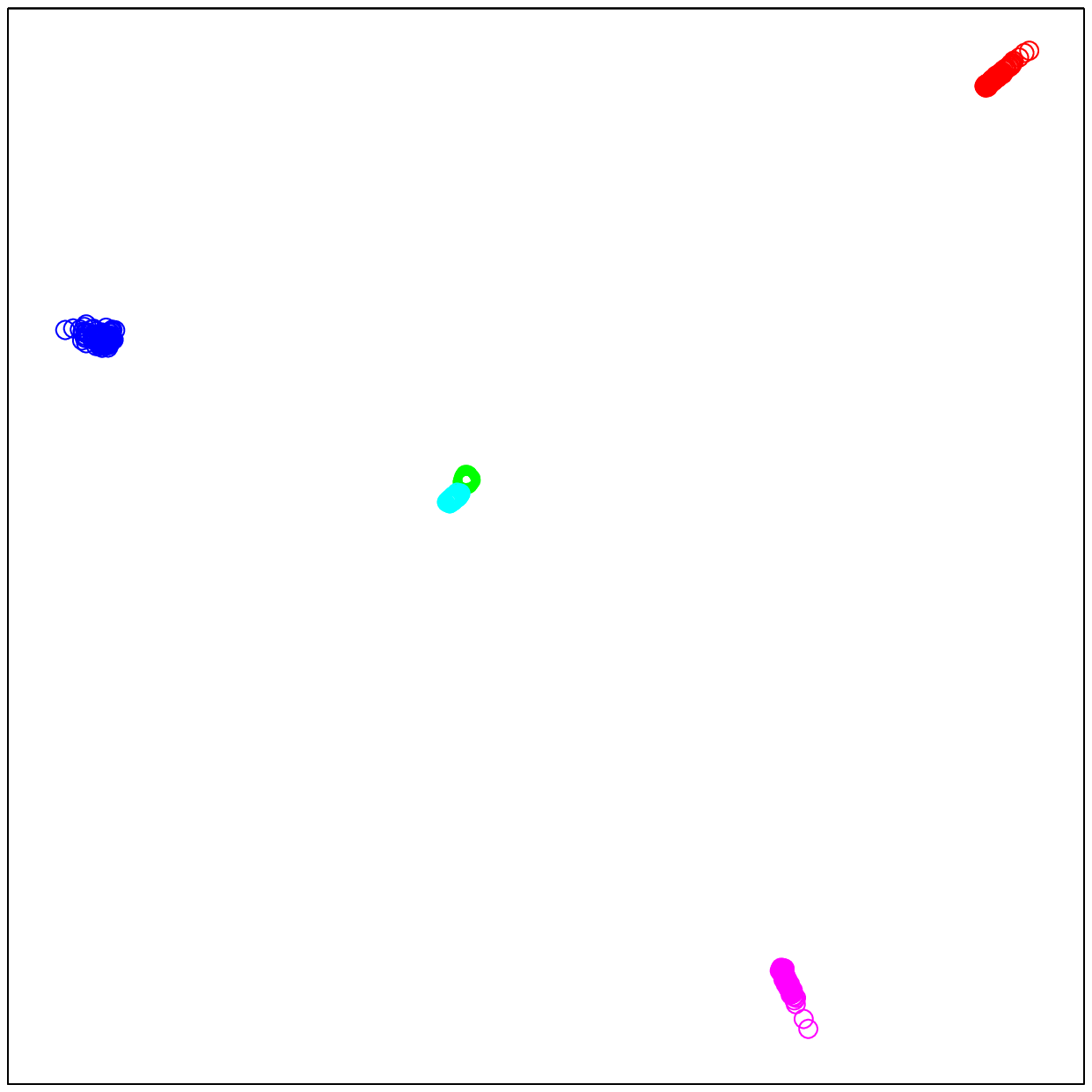} &
      \includegraphics[width=0.089\linewidth]{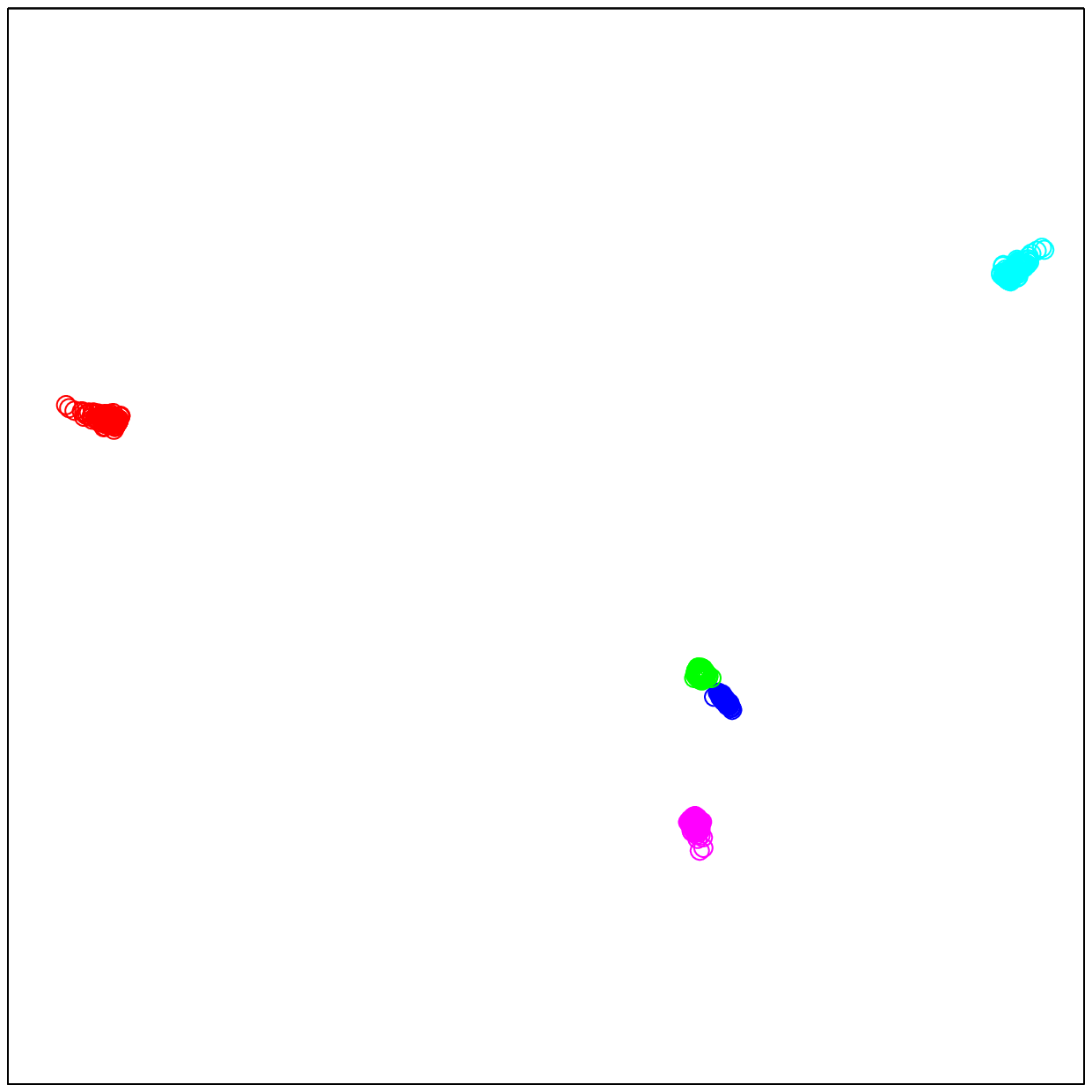} &
      \includegraphics[width=0.089\linewidth]{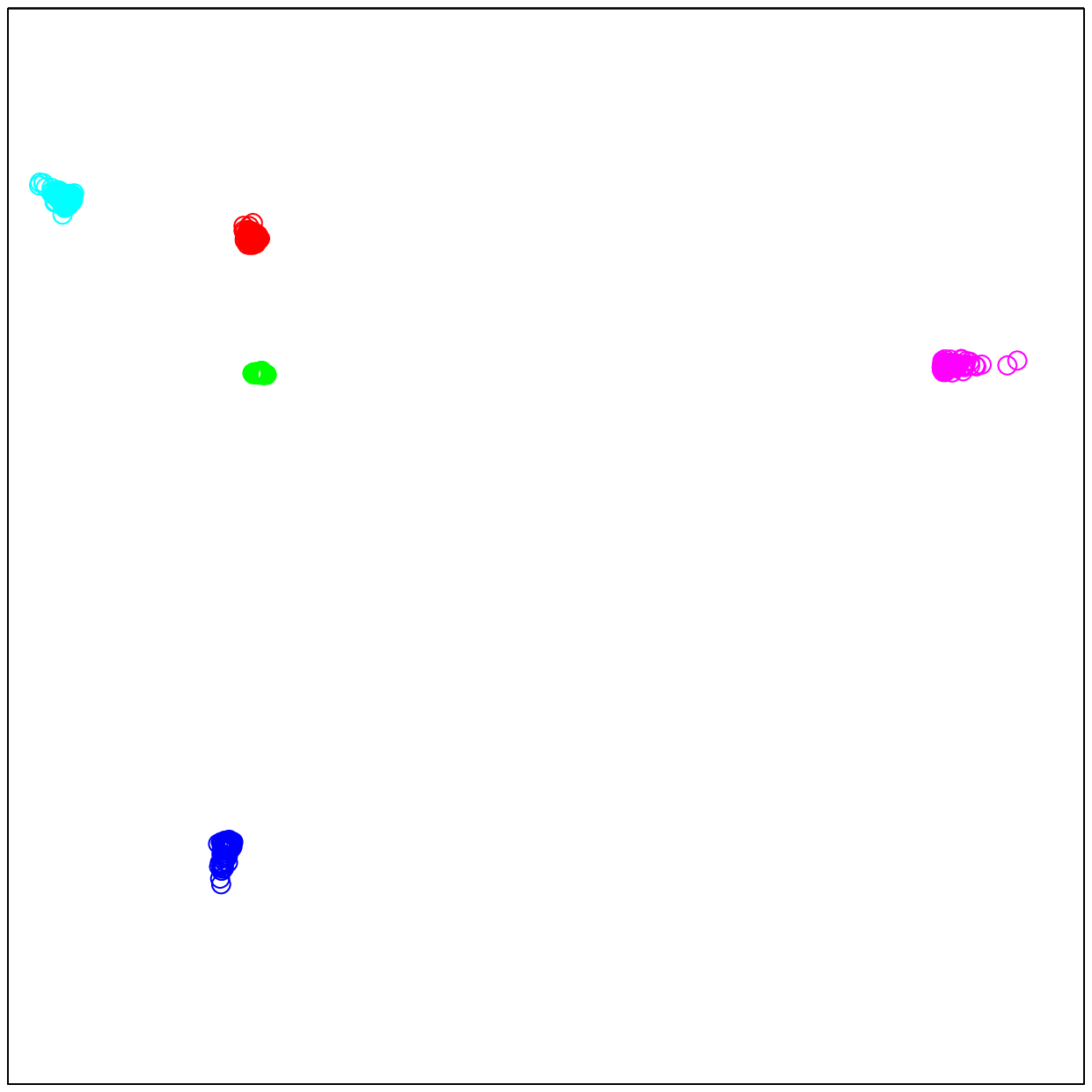} &
      \includegraphics[width=0.089\linewidth]{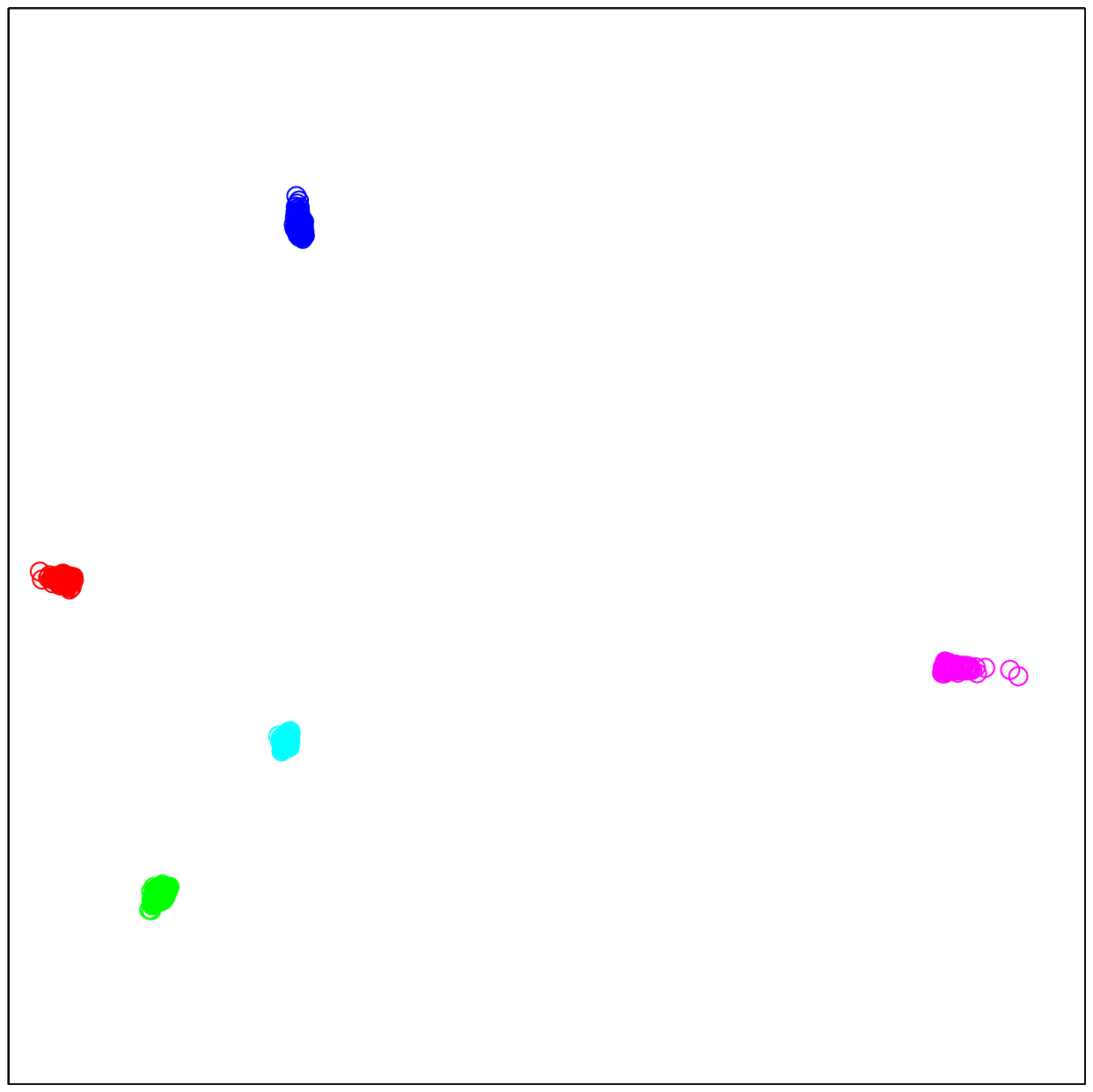} \\[-.6ex]
      \rotatebox{90}{\hspace{1.5ex}$K=6$} &
      \includegraphics[width=0.089\linewidth]{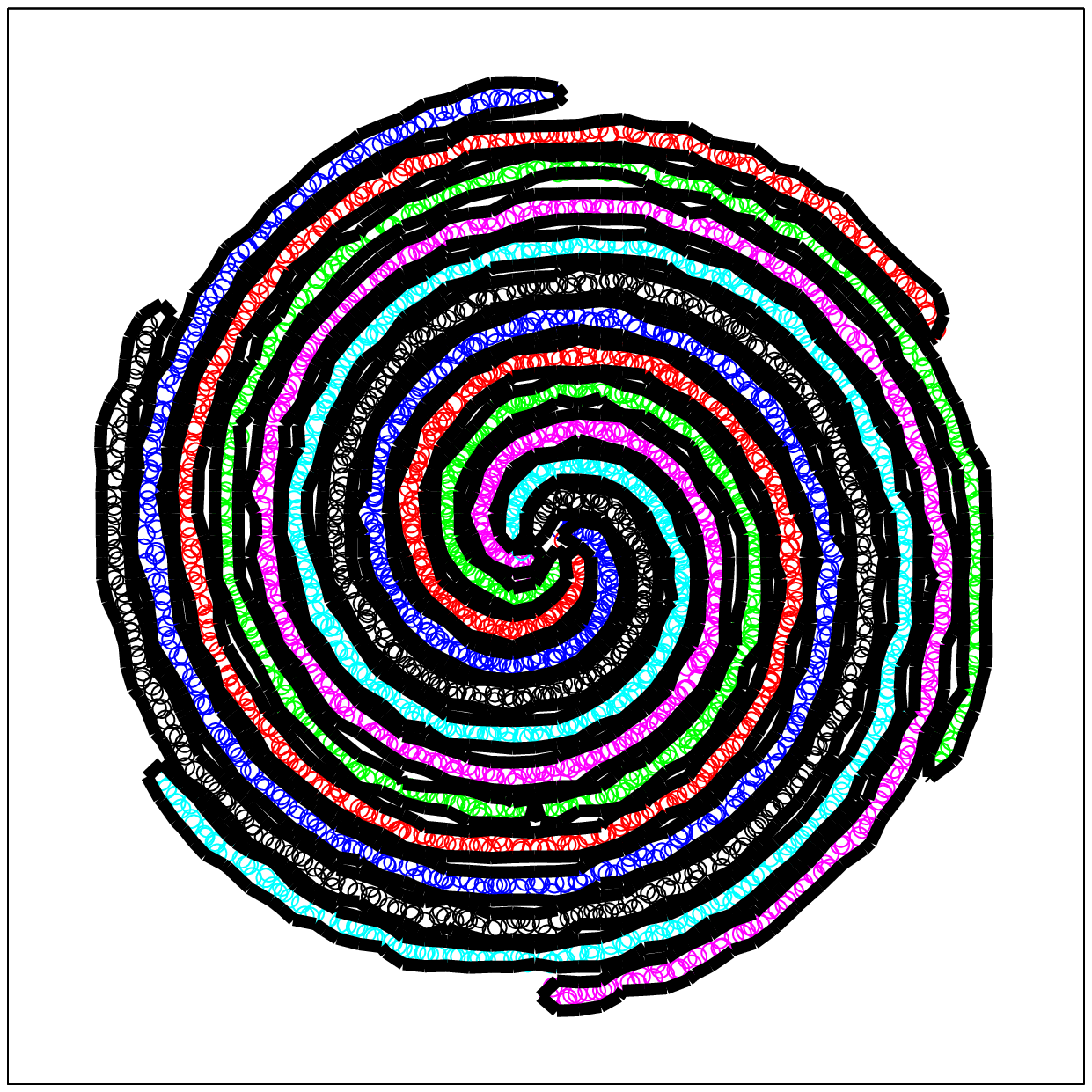} &
      \includegraphics[width=0.089\linewidth]{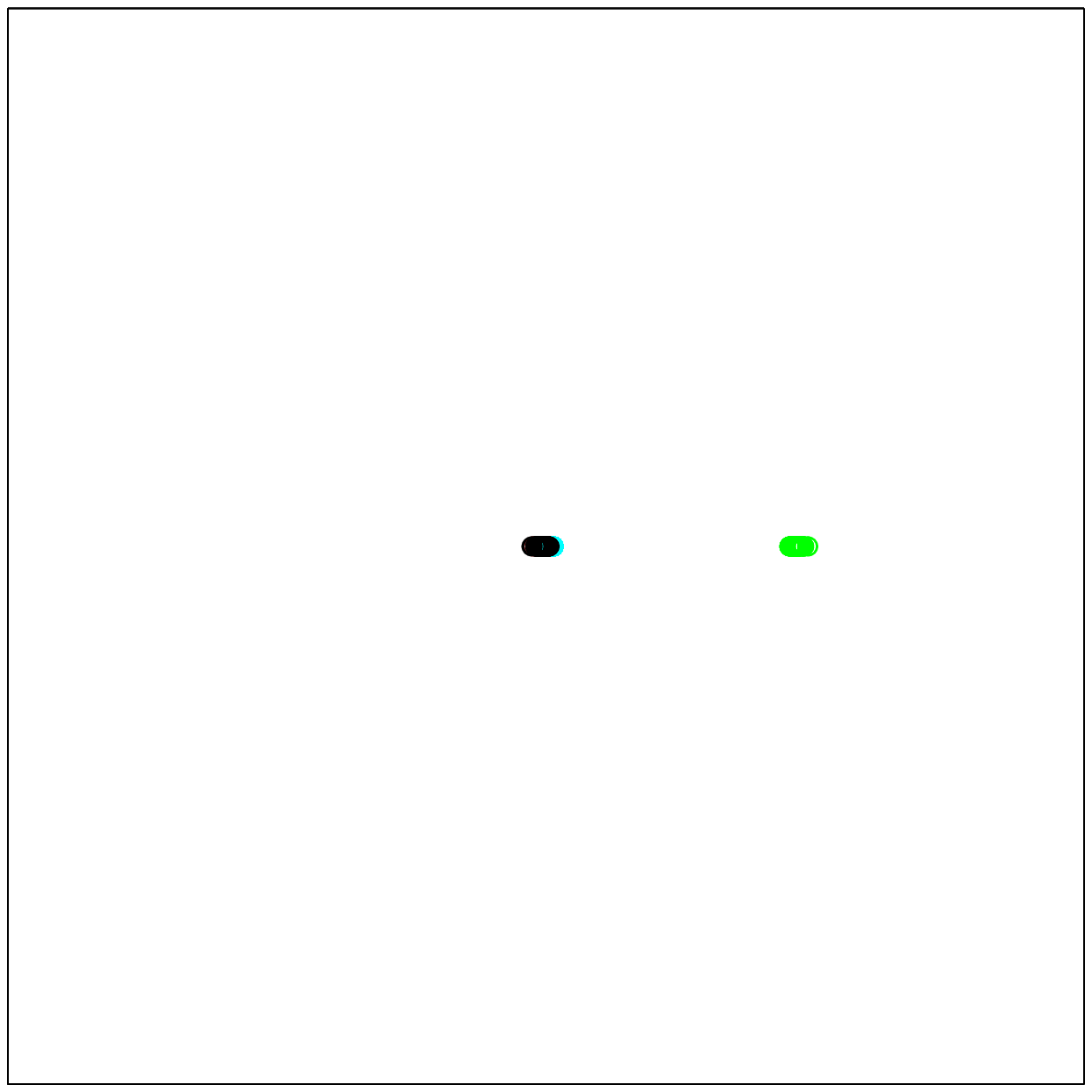} &
      \includegraphics[width=0.089\linewidth]{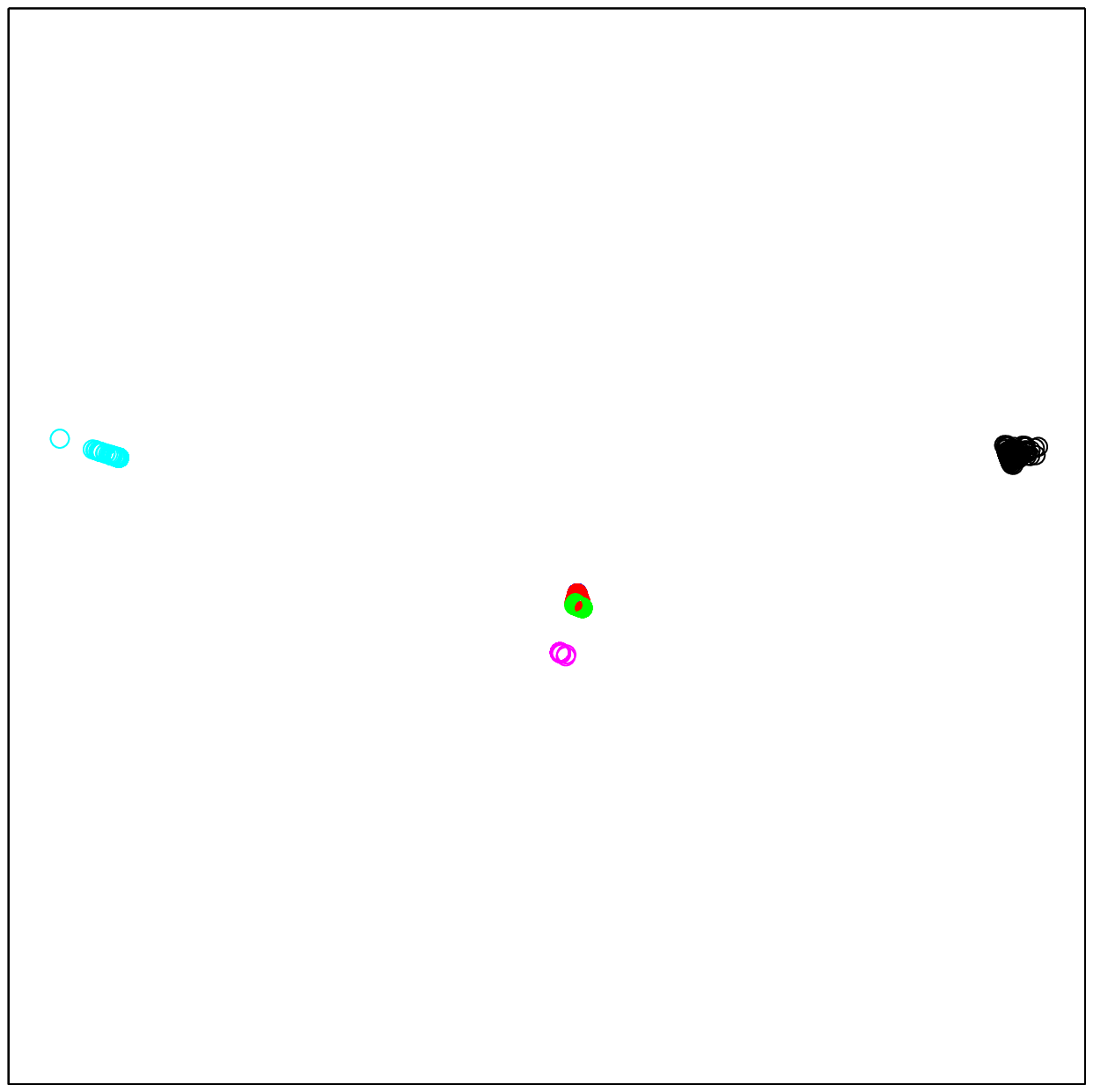} &
      \includegraphics[width=0.089\linewidth]{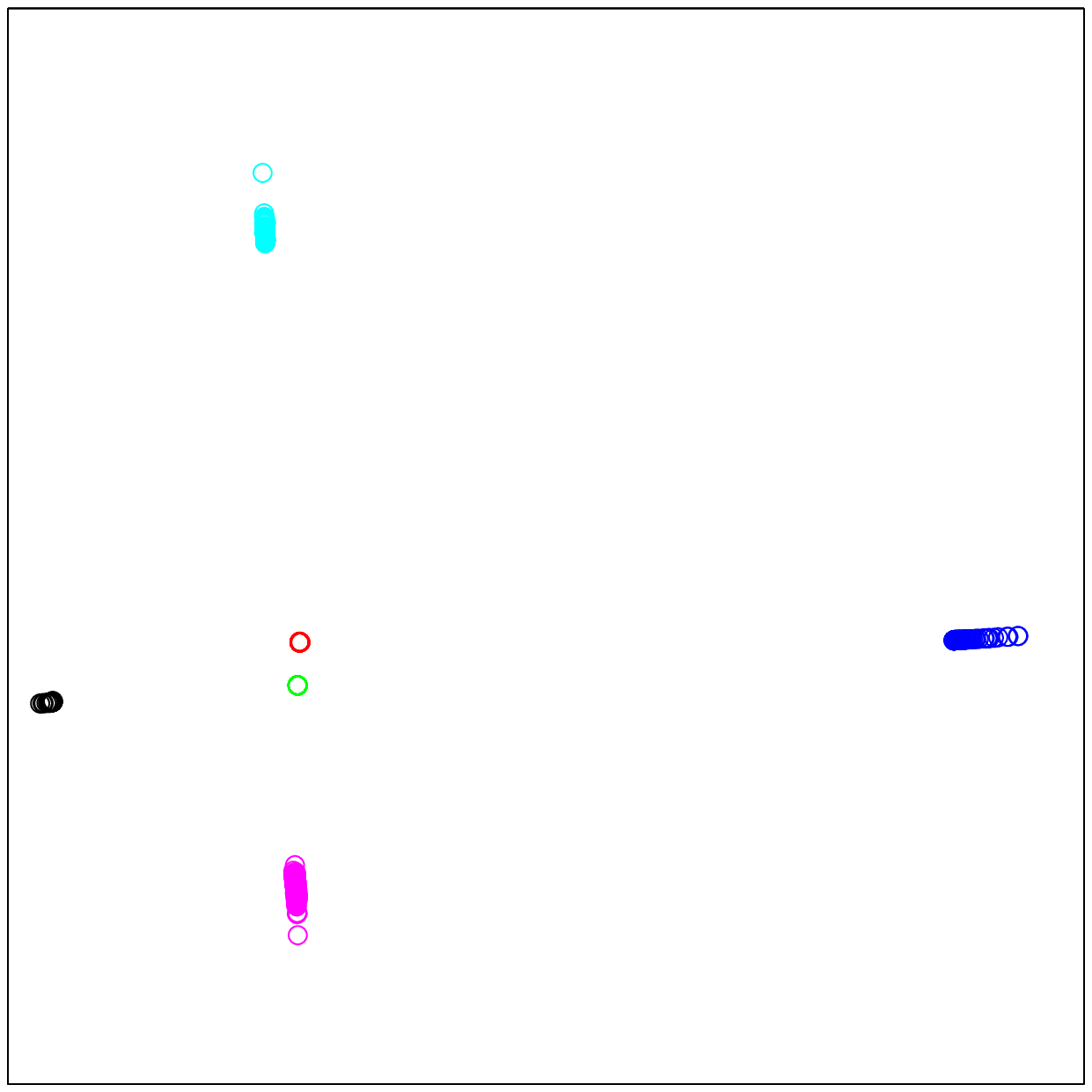} &
      \includegraphics[width=0.089\linewidth]{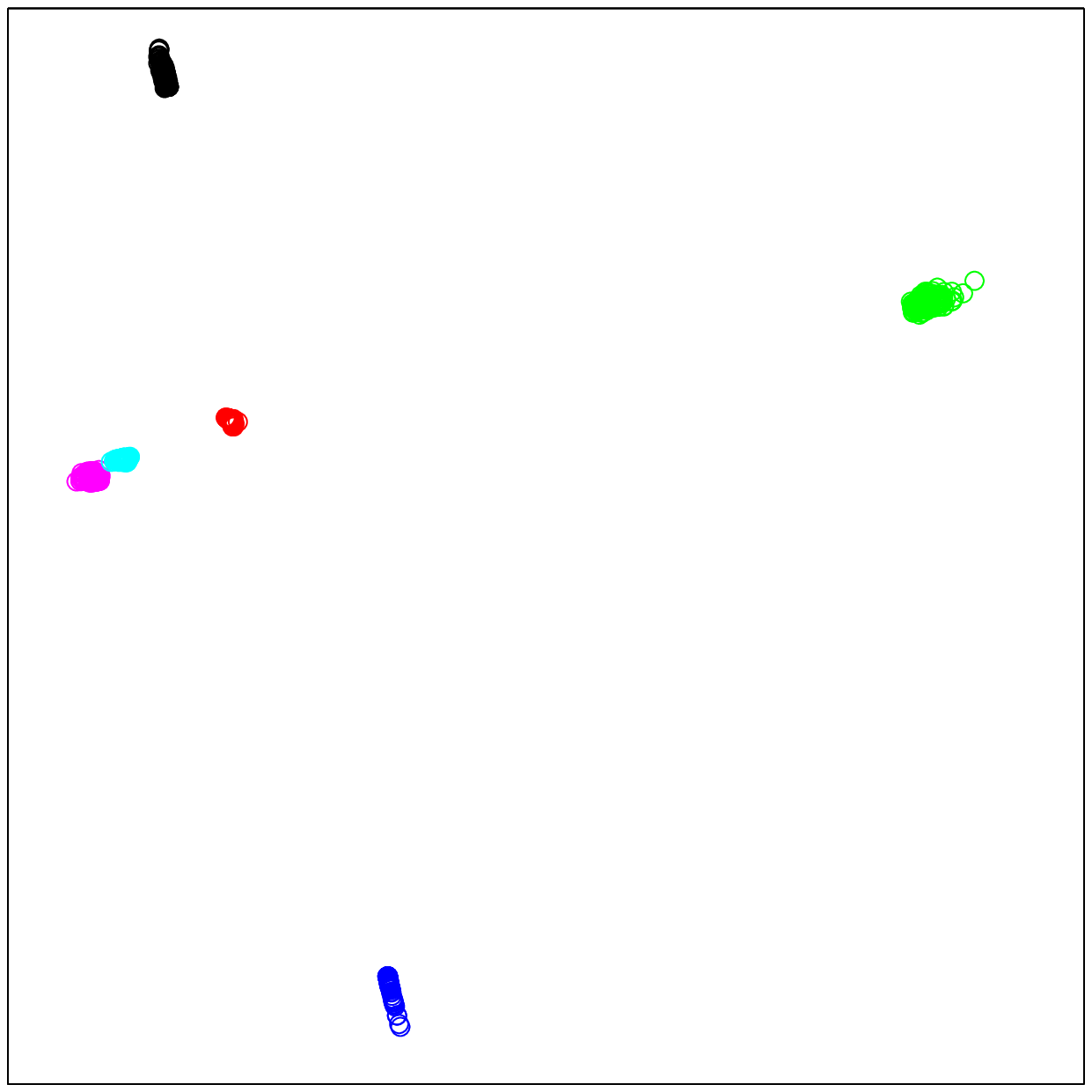} &
      \includegraphics[width=0.089\linewidth]{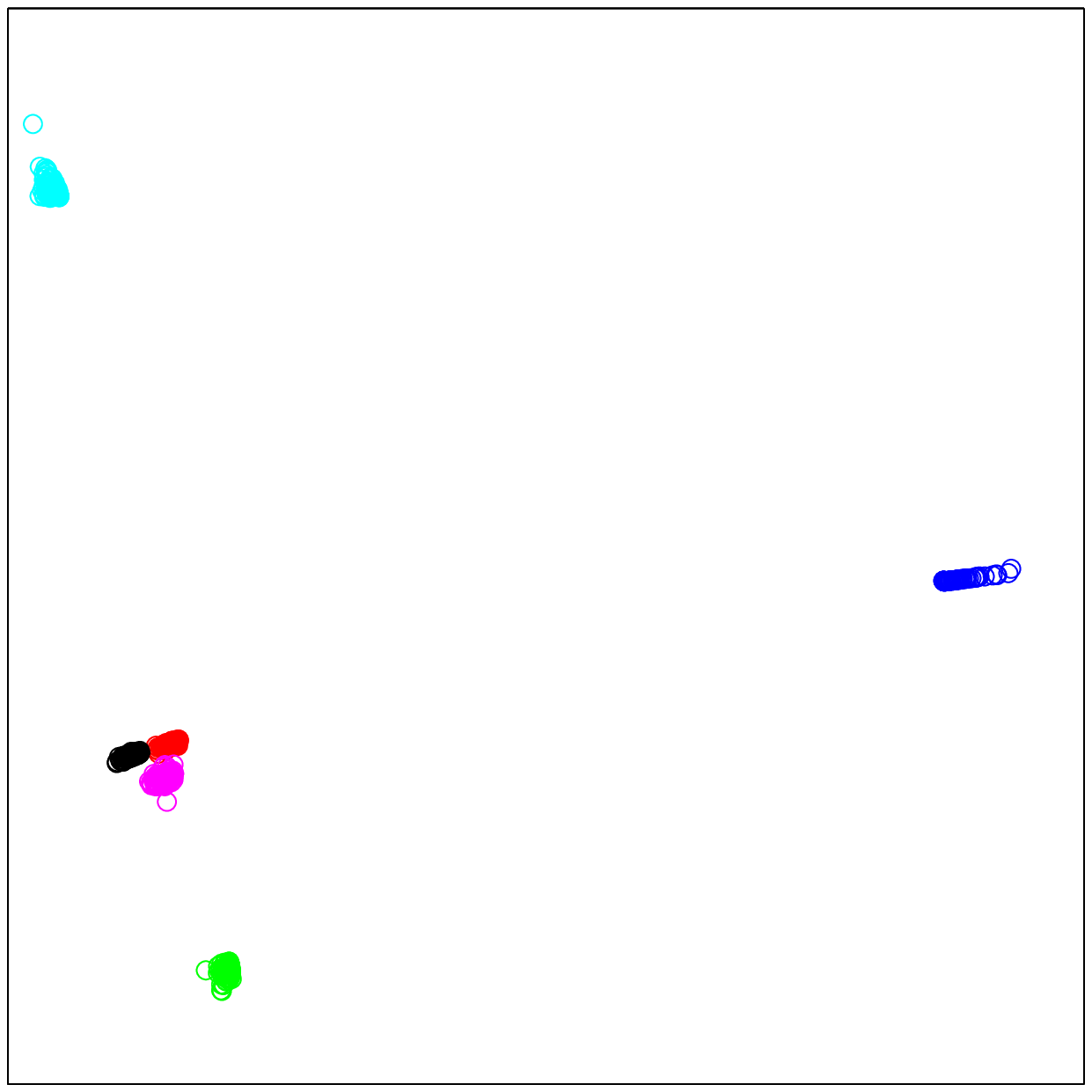} &
      \includegraphics[width=0.089\linewidth]{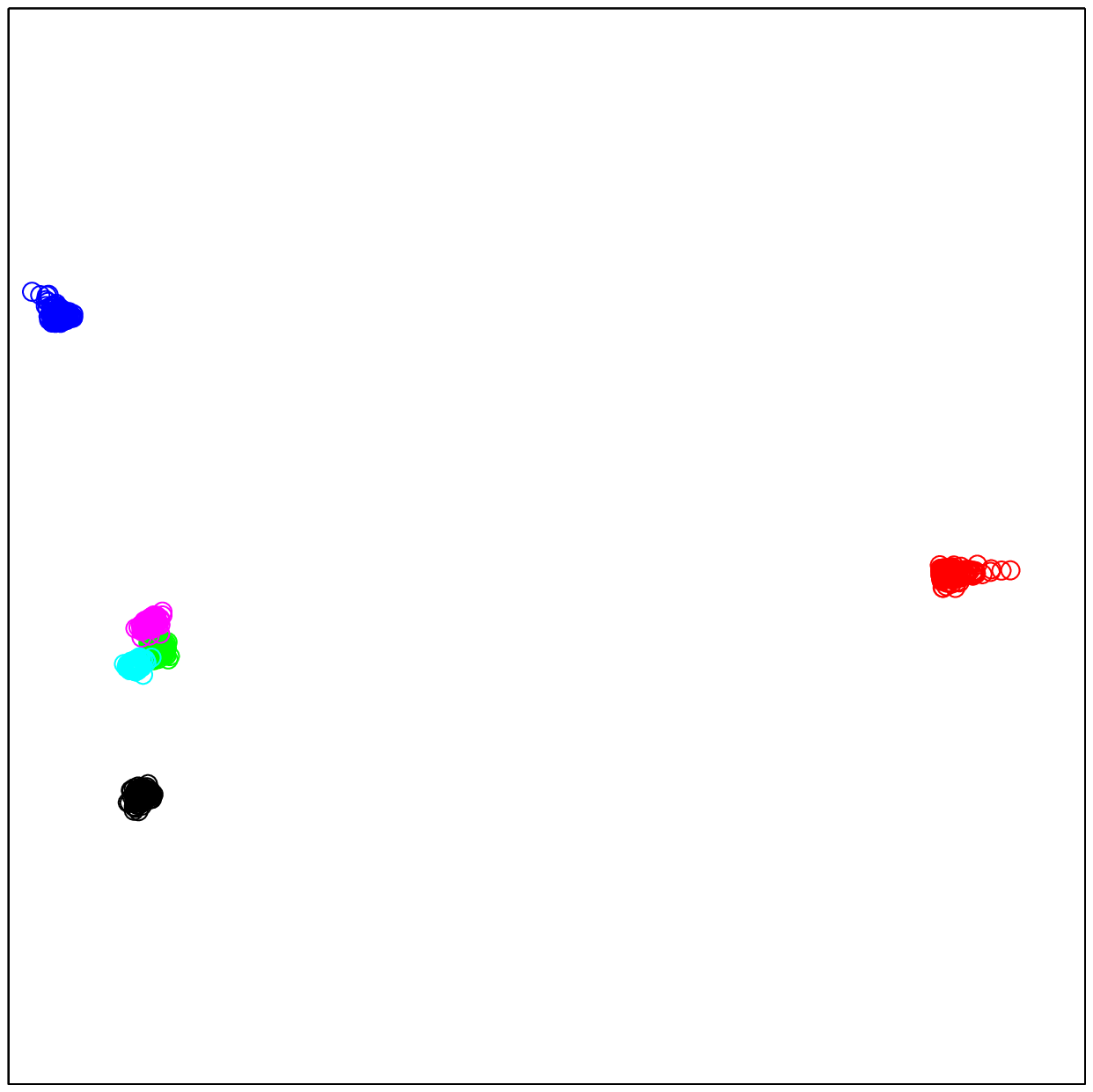} &
      \includegraphics[width=0.089\linewidth]{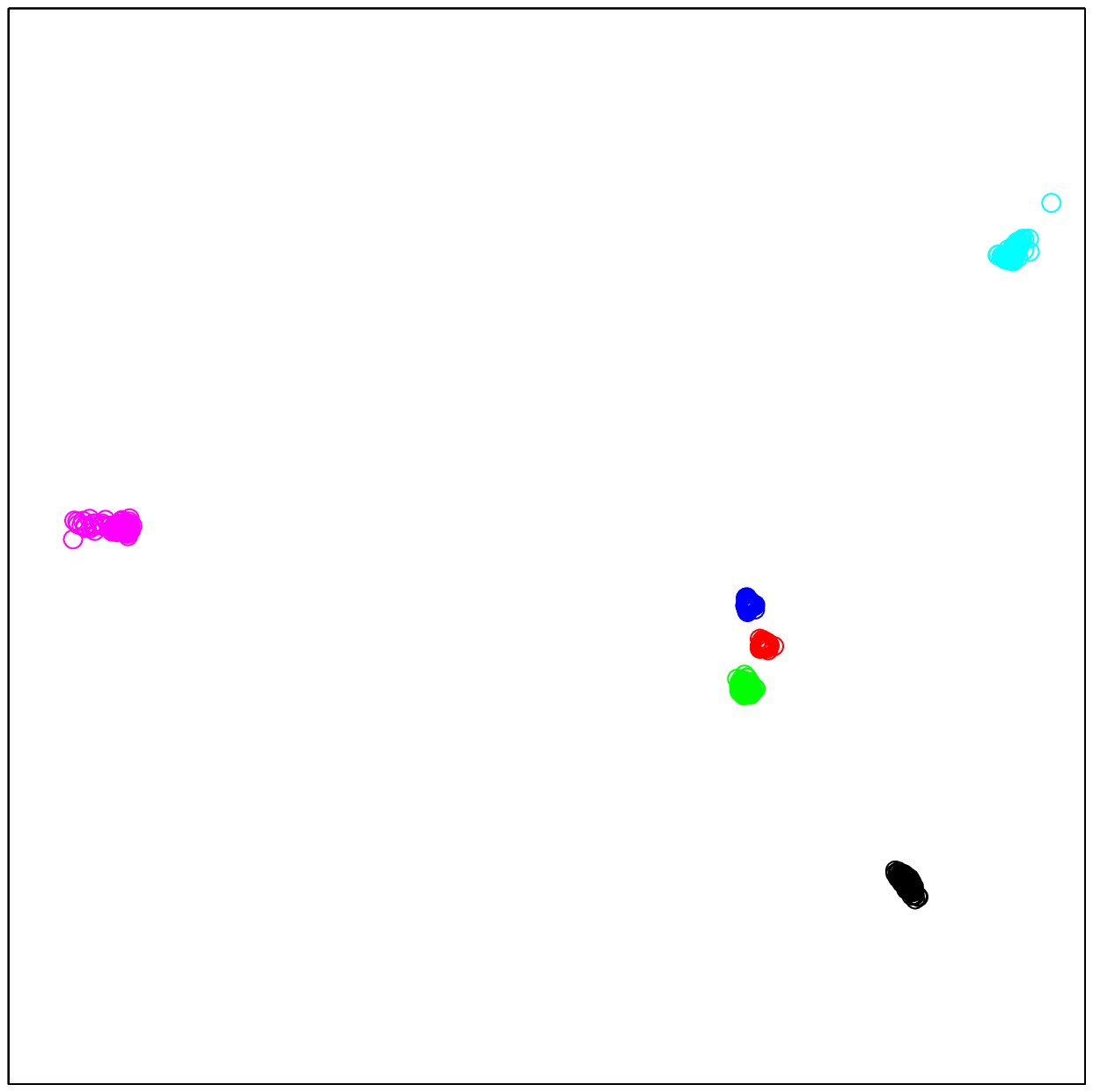} &
      \includegraphics[width=0.089\linewidth]{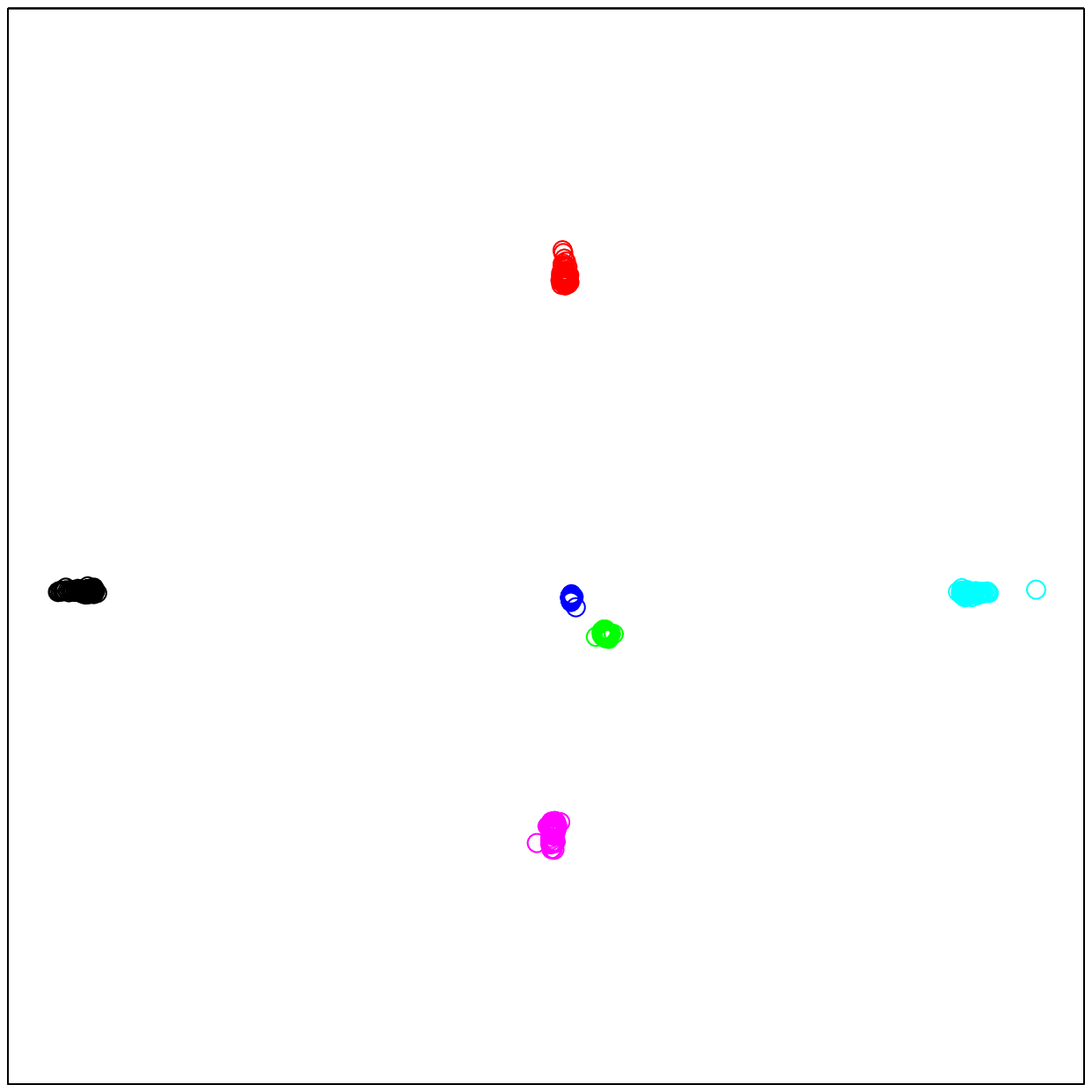} &
      \includegraphics[width=0.089\linewidth]{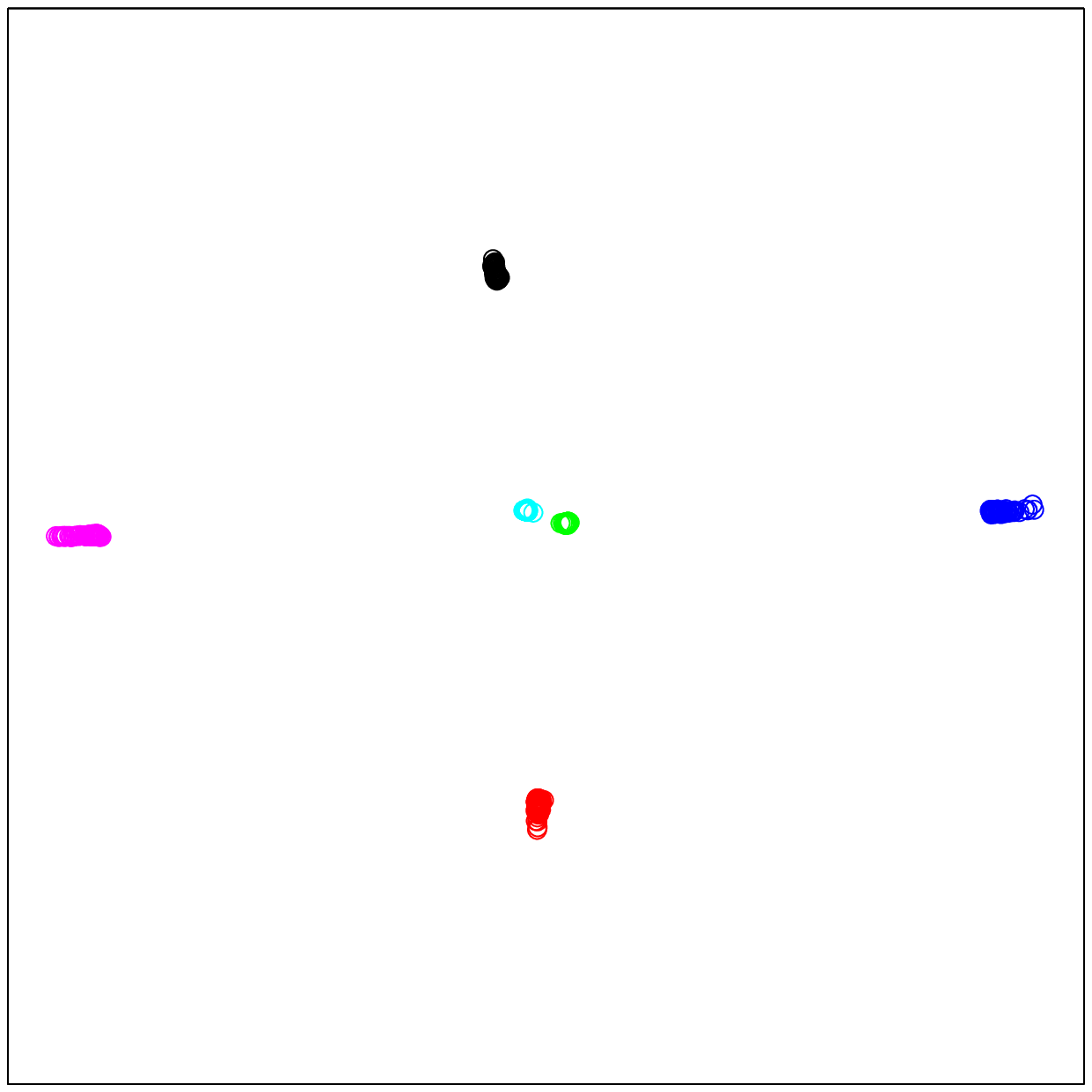} &
      \includegraphics[width=0.089\linewidth]{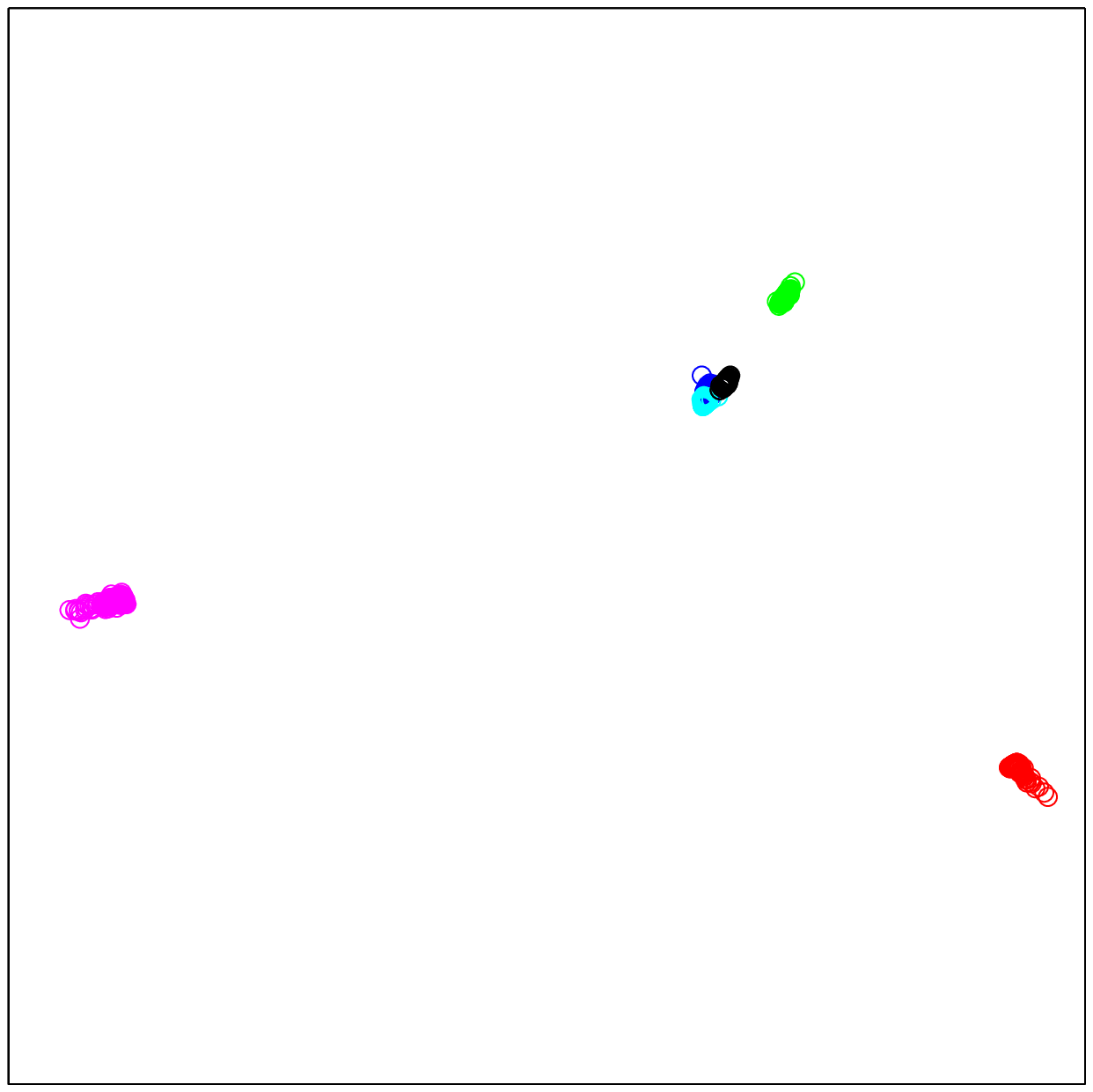}
    \end{tabular}
    \caption{Results on the $K$-spirals dataset. Each row shows results for a different $K$. First column: datasets and decision boundaries (black curve, you may need to zoom in) in input space obtained at dimension $L=K-1$. Columns $L=1$ to $10$: the latent projections $\F(\X)$ obtained at dimension $L$ (visualized in 2D using PCA).}
    \label{f:kspirals}
  \end{center}
\end{figure*}

We now study the role of the latent dimension $L$, the number of classes $K$ and the geometric configuration of the latent projections. We vary $K$ in the spirals dataset from $2$ to $6$, with $500$ samples in each spiral, and run our algorithm with \F\ being nonparametric RBFs and $L=1,\dots,10$, fixing other hyperparameters at reasonable values. Fig.~\ref{f:kspirals} shows the classification results and projections. We find we do not always obtain a perfect classification using only $L=2$ dimensions for the one-vs-all SVMs. Instead, we find a common behavior for different $K$: the classification performance improves drastically as $L$ increases in the beginning, and then stabilizes after some critical $L$, by which time the training samples are perfectly separated. This is because once the classes are separable in $\bbR^L$, they can also be (more easily) separable in higher dimensions with more degrees of freedom to arrange them. We observe in this experiment that, typically with $L=K-1$ dimensions, the classes all form a point-like cluster and approximately lie on vertices of a regular simplex, with zero classification error (decision boundaries shown in the first column). This gives a recipe to choose $L$: starting from $L=K-1$, increase $L$ until the classification error does not improve. It also gives a recipe to initialize \Z, namely to the ideal configuration of $K$ centroids located in the corners of a simplex, so all the $\z_n$ from the same class are set to the centroid of that class. Comparing this experimentally with random initial \Z, we observe that the simplex-based \Z\ leads to much faster convergence (usually 1 iteration), although interestingly the quality of the minimum found is similar to that of the random initial \Z.

In summary, these results are in approximate agreement with our ideal prediction, although in practice, the extent to which \F\ collapses classes depends on the number of basis functions. The more BFs, the more flexible \F\ is and the closer the latent projections are to $K$ centroids in a maximally linearly separable arrangement as described above, and this behavior arises from the joint optimization of DR and classifier. Note that finding an \F\ that maps a set of points to the same point is trivial: it is constant. But what we seek is an \F\ that maps each class' points to the class centroid. This is a piecewise constant mapping which is much harder to learn.

\begin{figure}[t]
  \centering
  \begin{tabular}{@{}c@{\hspace{0.01\linewidth}}c@{\hspace{0.01\linewidth}}c@{\hspace{0.01\linewidth}}c@{}}
    \begin{tabular}{c|c}
      \hline
      Method & Error (stdev), \% \\
      \hline
      NN & 19.16 (0.74)\\
      Linear SVM & 13.5 (0.72) \\
      PCA ($L=2$) & 42.10 (1.22) \\
      LDA ($L=1$) & 14.21 (1.63) \\
      LMNN ($L=2$) & 15.91 (1.65) \\
      \textbf{Ours ($L=1$)}  & \textbf{13.12 (0.67)} \\
      \textbf{Ours ($L=2$)}  & \textbf{12.94 (0.82)} \\
      \textbf{Ours ($L=20$)} & \textbf{12.76 (0.81)} \\
      \hline
    \end{tabular} &
    \begin{tabular}[c]{@{}c@{}}
      PCA \\
      \includegraphics[width=0.22\linewidth,bb=140 225 496 581]{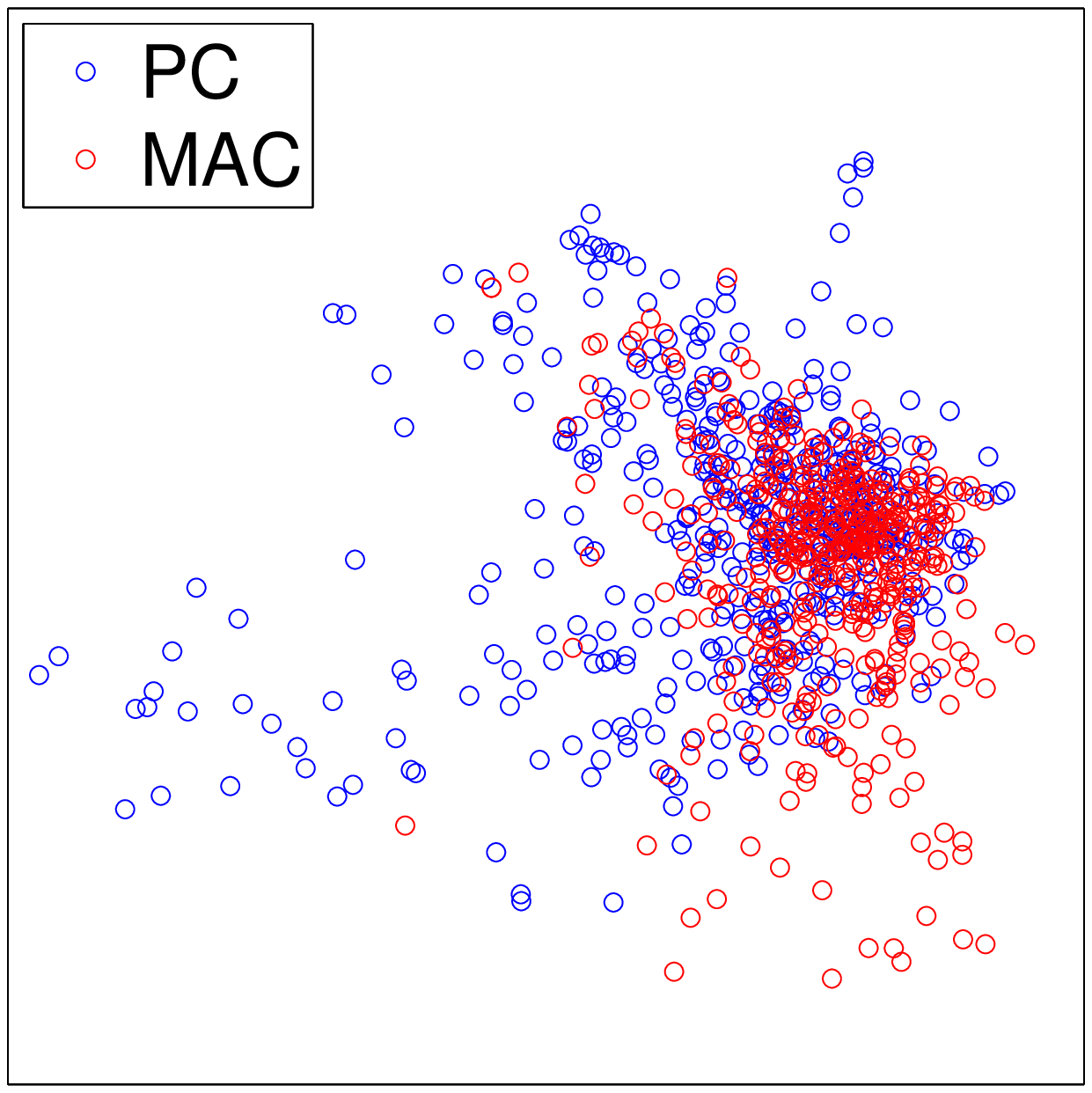}
    \end{tabular} &
    \begin{tabular}[c]{@{}c@{}}
      LDA \\
      \includegraphics[width=0.22\linewidth,bb=140 225 496 581]{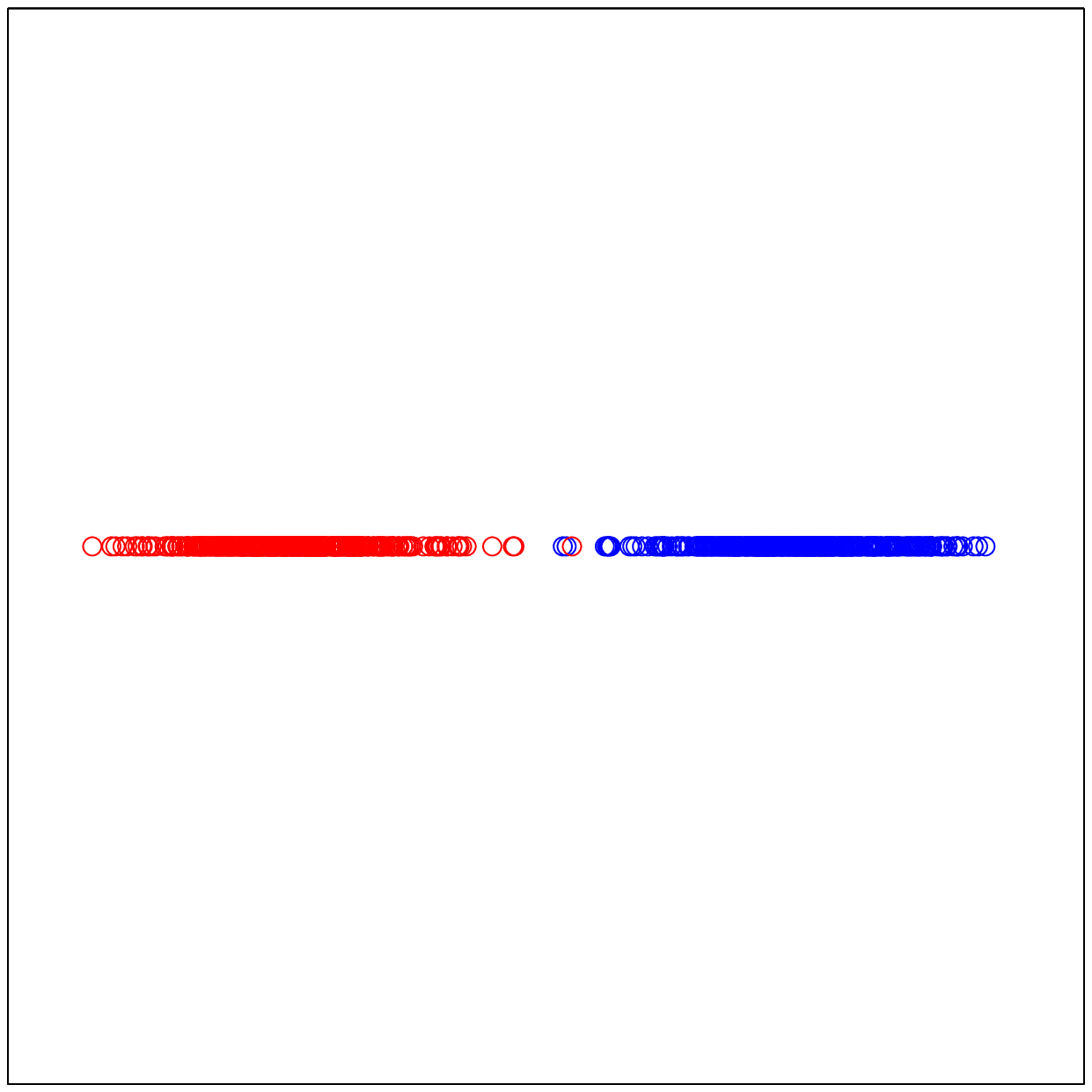}
    \end{tabular} &
    \begin{tabular}[c]{@{}c@{}}
      LMNN \\
      \includegraphics[width=0.185\linewidth]{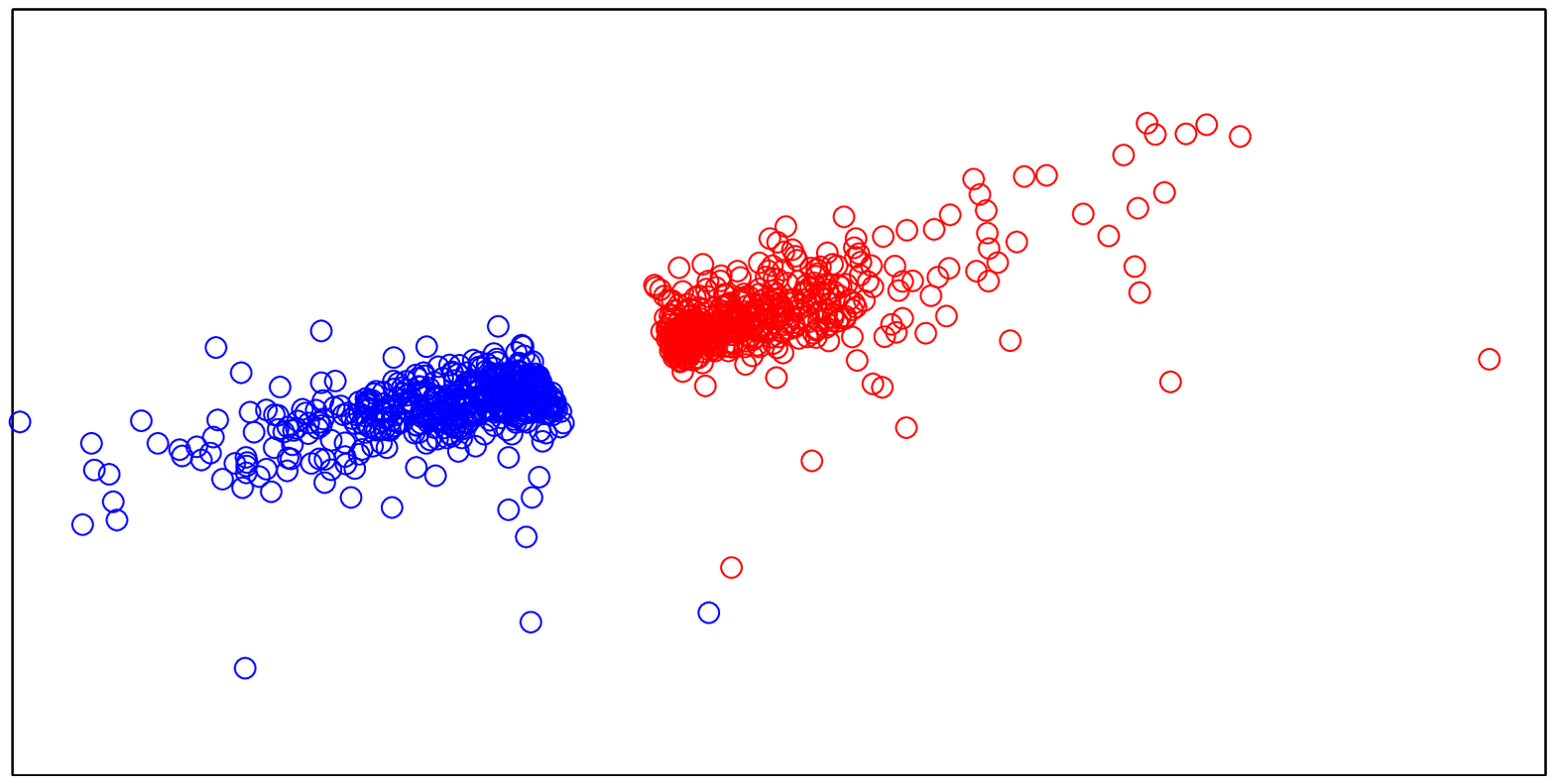} \\
      Ours \\
      \includegraphics[width=0.185\linewidth]{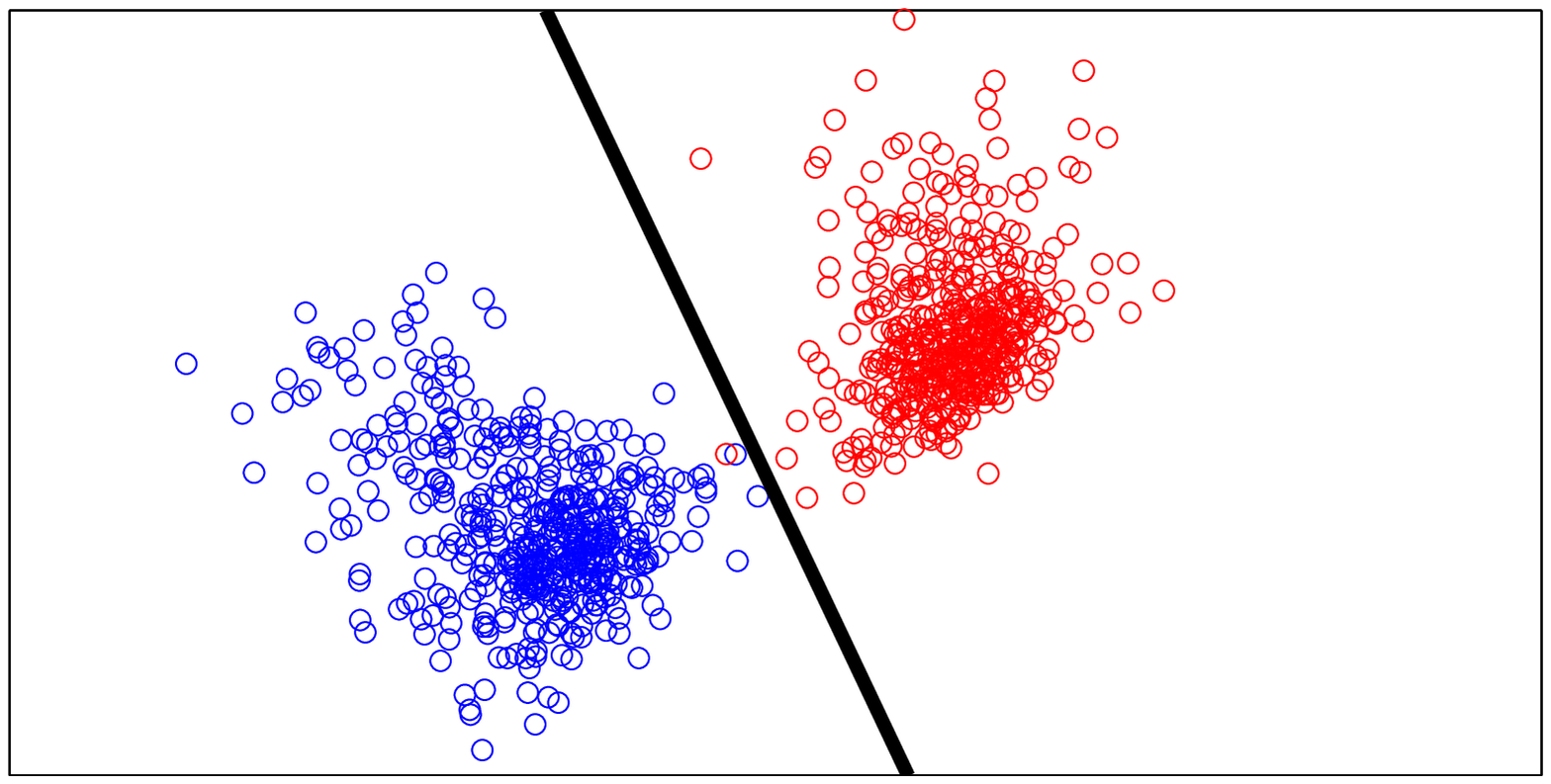}
    \end{tabular}
  \end{tabular}
  \caption{Results on the PC/MAC subset of 20 newsgroups. \emph{Left}: mean test error rate over 10 splits (standard deviation in parenthesis). \emph{Right}: projections by different algorithms.}
  \label{f:PCMAC_results}
\end{figure}

\subsection{Comparison with other classifiers}

We compare with directly applying a nearest neighbor classifier (NN) and linear (LSVM) or Gaussian kernel SVMs (GSVM) on the original inputs, with unsupervised DR methods PCA/Gaussian Kernel PCA (KPCA) followed by nearest neighbor, and with filter approaches such as LDA/KLDA (Gaussian kernel) followed by nearest neighbor. Hyperparameters for these algorithms (kernel width $\sigma$ for kernel SVM, KPCA, and KLDA, penalty parameter $C$ for SVMs) are chosen by grid search on a separate validation set. In our algorithm, we initialize \Z\ randomly.

\paragraph{Document  binary classification}

We pick two classes from the $20$ newsgroup dataset (comp.sys.ibm.pc.hardware and comp.sys.mac.hardware), remove words appearing in $5$ or fewer documents, reduce the dimension to $2\,989$, and then extract the TFIDF features. We further reduce the input dimension to $1\,000$ with PCA (keeping $>98\%$ of the variance). For evaluation purposes, we create $10$ different $80$/$20$ splits of the $1\,162$ training items into training and validation set. For each split, we let all algorithms pick the optimal hyperparameters based on validation error, and evaluate the optimal models on the test set. Due to the high dimensionality and scarcity of samples, we use linear \F\ for this problem (we did try RBFs for \F\ and obtained similar results). We also run Large Margin Nearest Neighbor (LMNN; \citealp{WeinberSaul09a}), a metric learning algorithm, with the recommended number of target neighbors $3$. We report the mean and standard deviation of test error rates over 10 splits for all methods in fig.~\ref{f:PCMAC_results}(left).

Our classification performance is quite robust to the choice of $L$, although in general more dimensions may bring a little improvement in the error rate at higher computational cost. We thus fix the latent dimensions to be $2$ for other methods. The results show that using class information in dimensionality reduction is superior to not using it (e.g.\ PCA), and we outperform others consistently with different $L$. Fig.~\ref{f:PCMAC_results}(right) shows the 2D projections of several algorithms, where supervised DR algorithms manage to separate the classes and PCA does not.

\paragraph{MNIST odd/even classification} 

We perform a binary classification task of discriminating odd digits ($1,3,5,7,9$) from the even digits ($0,2,4,6,8$) on the MNIST benchmark. We vary the training set size to be $100$, $200$, $500$, $1\,000$, $2\,000$, $3\,000$, $4\,000$, and $5\,000$, including equal number of images randomly sampled for each digit. We use in all cases a balanced validation set of $10\,000$ images from which we pick optimal hyperparameters for all methods. The MNIST test set ($10\,000$ samples) is used for evaluating the classification performance. Since this dataset is likely not linearly separable (as can be inferred from the error rate of the linear SVM), we fix the dimension of our latent space to be $L=2$, and use nonparametric RBFs for \F. We compare our algorithm with nonlinear DR algorithms KPCA ($L=2$) and KLDA ($L=1$). We also explore a two-step approach of using linear DR (PCA/LDA) followed by a Gaussian SVM. We cross-validate $L$ for PCA, while LDA uses $L=1$.

\begin{figure}[t]
  \centering
  \begin{tabular}{@{}c@{}c@{}c@{}}
    \psfrag{error}[b][]{error rate (\%)}
    \psfrag{size}[][B]{training set size ($\times 10^2$)}
    % \psfrag{LSVM}[l][l]{\tiny linear SVM}
    % \psfrag{GSVM}[l][l]{\tiny Gaussian SVM}
    % \psfrag{NN}[l][l]{\tiny NN}
    % \psfrag{KPCA}[l][l]{\tiny KPCA}
    % \psfrag{KLDA}[l][l]{\tiny KLDA}
    % \psfrag{Ours}[l][l]{\tiny Ours}
    % \psfrag{LDA+GSVM}[l][l]{\tiny LDA+Gaussian SVM}
    % \psfrag{PCA+GSVM}[l][l]{\tiny PCA+Gaussian SVM}
    % \psfrag{PCA+Ours}[l][l]{\tiny PCA+ours}
    \begin{tabular}[c]{@{}c@{}}
      \includegraphics[width=0.50\linewidth]{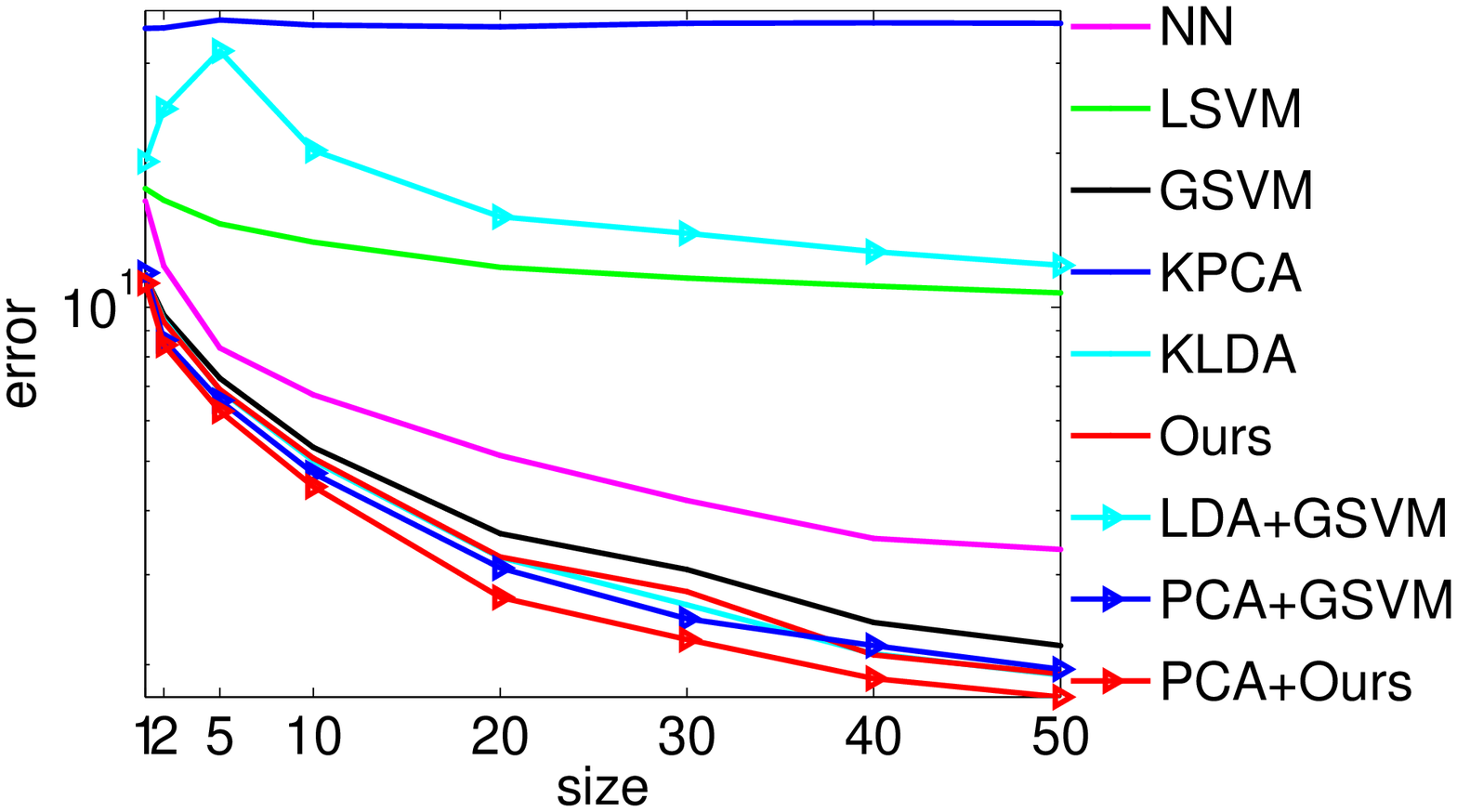}
    \end{tabular} &
    \begin{tabular}[c]{@{}c@{}}
      KPCA \\
      \includegraphics[width=0.25\linewidth]{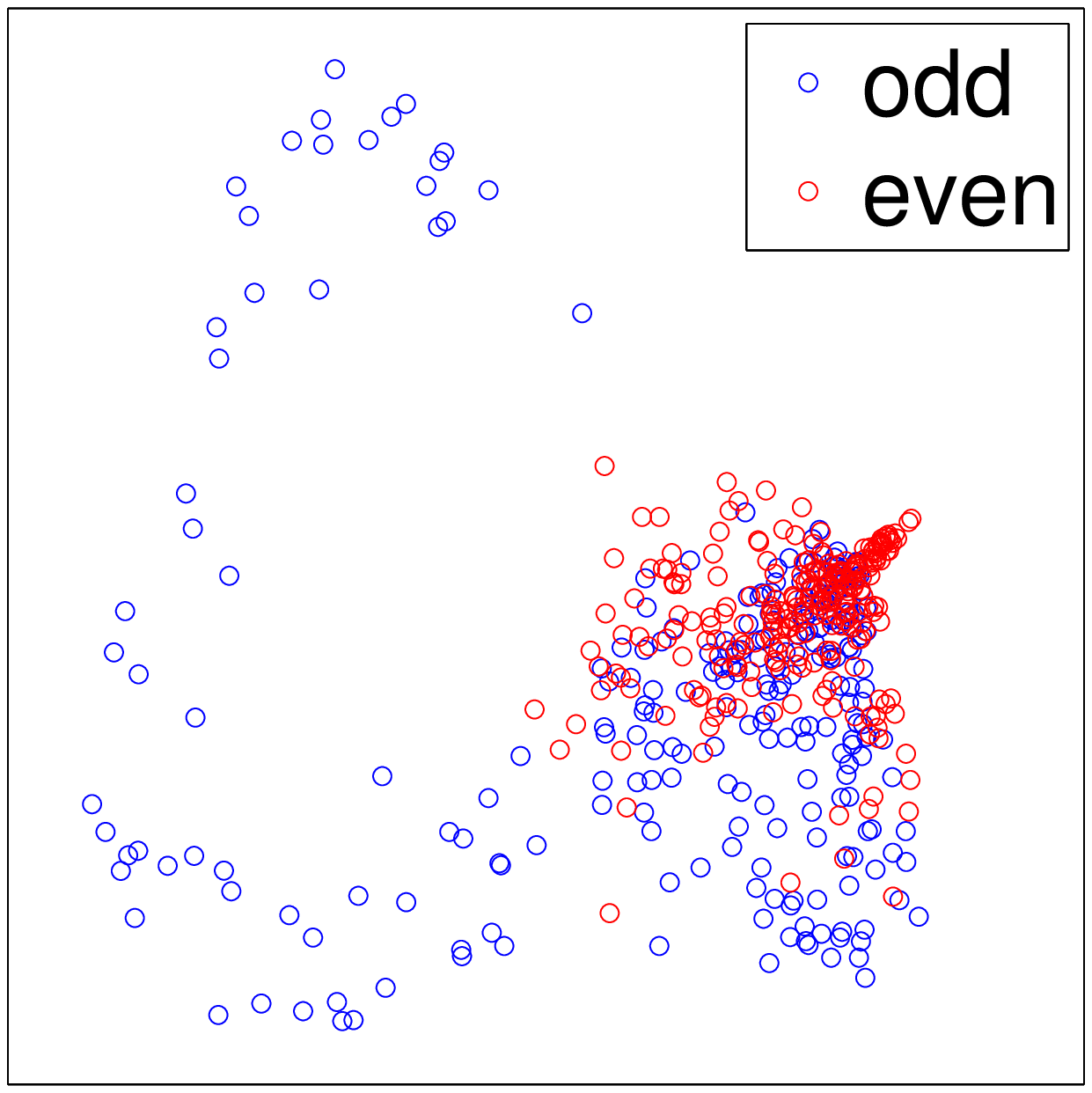}
    \end{tabular} &
    \begin{tabular}[c]{@{}c@{}}
      Ours \\
      \includegraphics[width=0.25\linewidth]{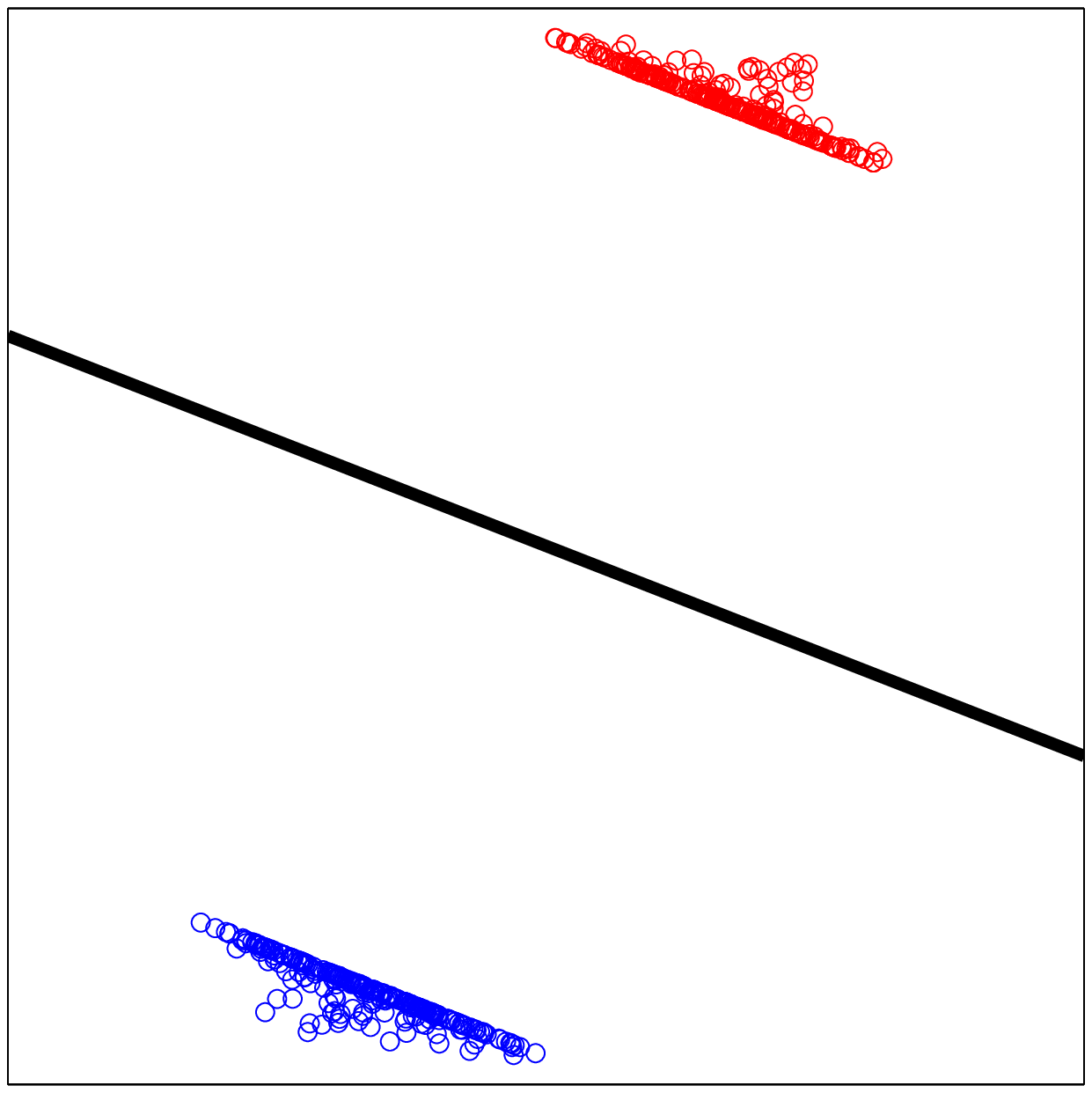} 
    \end{tabular}
  \end{tabular}
  \caption{Results on MNIST odd/even classification. \emph{Left}: errors of different algorithms over dataset sizes. \emph{Right}: 2D projections by KPCA and our algorithm.}
  \label{f:ODDEVEN_errors}
\end{figure}

Figure~\ref{f:ODDEVEN_errors} (top) shows the test errors of all methods over different training set sizes. On the original inputs, our algorithm and  KLDA (given very fine grid search for optimal kernel width) perform the best. Their clear advantage over KPCA demonstrates again that class information should be incorporated in DR. We expect our model to have a regularization effect and improve generalization more in smaller training sets. Indeed, we find that we improve over the nearest neighbor classifier (known to achieve near optimal error for large training sets; \citealp{Duda_01a}) most for $100$-$200$ training samples. Fig.~\ref{f:ODDEVEN_errors}(bottom) shows the 2D projection of the training set of $500$ samples obtained by KPCA and our algorithm. The classes are perfectly separated by our algorithm but not by KPCA, which uses no label information.

Due to the limited power of its DR mapping, LDA+Gaus\-sian SVM performs poorly. PCA+Gaussian SVM performs well (slightly better than our algorithm on original inputs), though only at a much larger $L$ (around 40). We hypothesize this is because PCA is able to remove some noise that is not learned by us from the inputs. We then trained our algorithm on the PCA projection, further reducing the dimension to $L=2$, and obtained consistently better accuracy (shown in fig.~\ref{f:ODDEVEN_errors} as PCA+Ours). 

\paragraph{MNIST 10-classes}

We now consider the problem of classifying the $10$ digit classes of MNIST. We randomly sample $10\,000$ images for training and $10\,000$ for validation. The original test set is used for evaluating different algorithms. We were not able to run KPCA and KLDA because the naive implementation of these algorithms requires a huge memory space to store the kernel matrix of $N\times N$ and solve a dense eigenvalue problem. Our algorithm uses $2\,500$ BFs with centers chosen by $K$-means.

We searched hyperparameters as follows in order to avoid unnecessary regions of the hyperparameter space. We first do a careful grid search for the kernel width and penalty parameter for the Gaussian SVM. The optimal kernel width ($\sigma=4.0$) is also used as the RBF kernel width of \F. Then $2\,500$ centers are chosen by K-means as RBF basis centers. The regularization parameter $\lambda$ of \F\ is fixed to $10^{-3}$ based on experiments on a much smaller training set (there exists a wide range of $\lambda$ for which our method works well). Thus we only search for the penalty parameter $C$ of \g, which is common to all $10$ SVMs, chosen from $\{10^{-3},10^{-2},10^{-1},1,10^1,10^2,10^3\}$.

We again explored the approach of PCA/LDA+Gaussian SVM as in the other MNIST experiment and obtained a similar result: LDA ($L=9$)+Gaussian SVM performs poorly; PCA+Gaussian SVM performs well at a relatively larger $L$. We trained our model on the PCA projection, further reduced the dimension to $L=10$, and obtained similar performance with much fewer basis functions.

As happened in the case of binary classification, starting from a random initialization, the \Z\ projections reorganize quickly in the latent space and are well classified after a few iterations. 

\begin{figure*}[t]
  \centering
  \begin{tabular}{@{}c@{\hspace{0.03\linewidth}}c@{}}
    \begin{tabular}{@{}c|c|c@{}}
      \hline
      Method & Error (\%) & \# BFs \\
      \hline
      Nearest Neighbor & 5.34 & 10\,000 \\
      Linear SVM  & 9.20 & --- \\
      Gaussian SVM  & 2.93 & 13\,827 \\
      LDA ($9$) + Gaussian SVM & 10.67 & 8\,740 \\
      PCA ($5$) + Gaussian SVM & 24.31 & 13\,638 \\
      PCA ($10$) + Gaussian SVM & 7.44 & 5\,894 \\
      PCA ($40$) + Gaussian SVM & 2.58 & 12\,549 \\
      \textbf{Ours ($5$ , $43$)} & \textbf{4.69} & \textbf{2\,500} \\
      \textbf{Ours ($10$, $18$)} & \textbf{2.99} & \textbf{2\,500} \\
      \textbf{PCA ($40$) + Ours ($10$, $17$)} & \textbf{2.60} & \textbf{2\,500} \\
      \hline
    \end{tabular} &
    \begin{tabular}[c]{@{}c@{}}
      \includegraphics[width=0.48\linewidth]{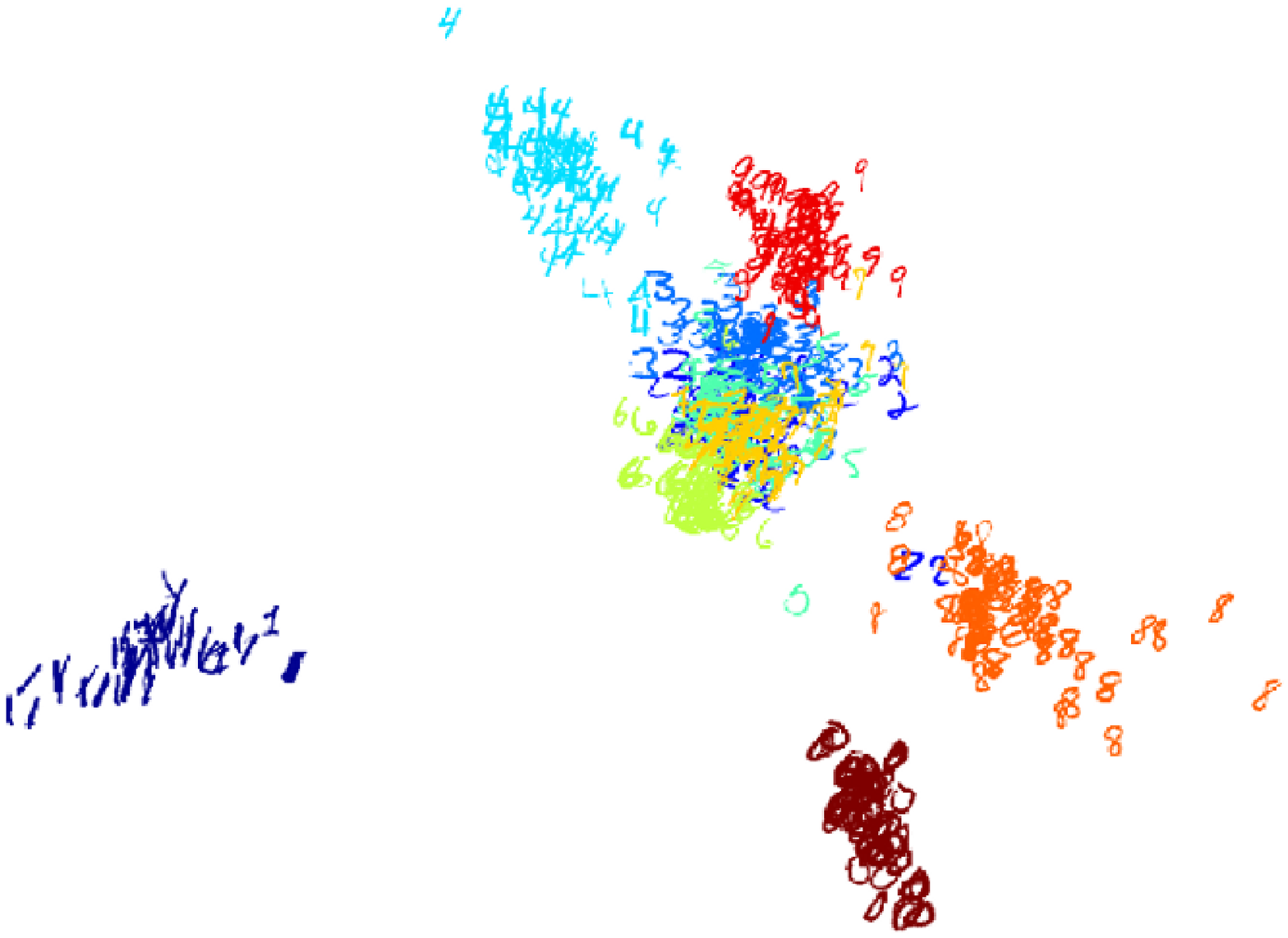}
    \end{tabular}
  \end{tabular} \\
  \begin{tabular}{@{}c@{\hspace{0.03\linewidth}}c@{\hspace{0.03\linewidth}}c@{}}
    \begin{tabular}[b]{@{}c@{}}
      \psfrag{dimension}[t][]{$L$}
      \psfrag{error}[][t]{error rate (\%)}
      \includegraphics[width=0.31\linewidth]{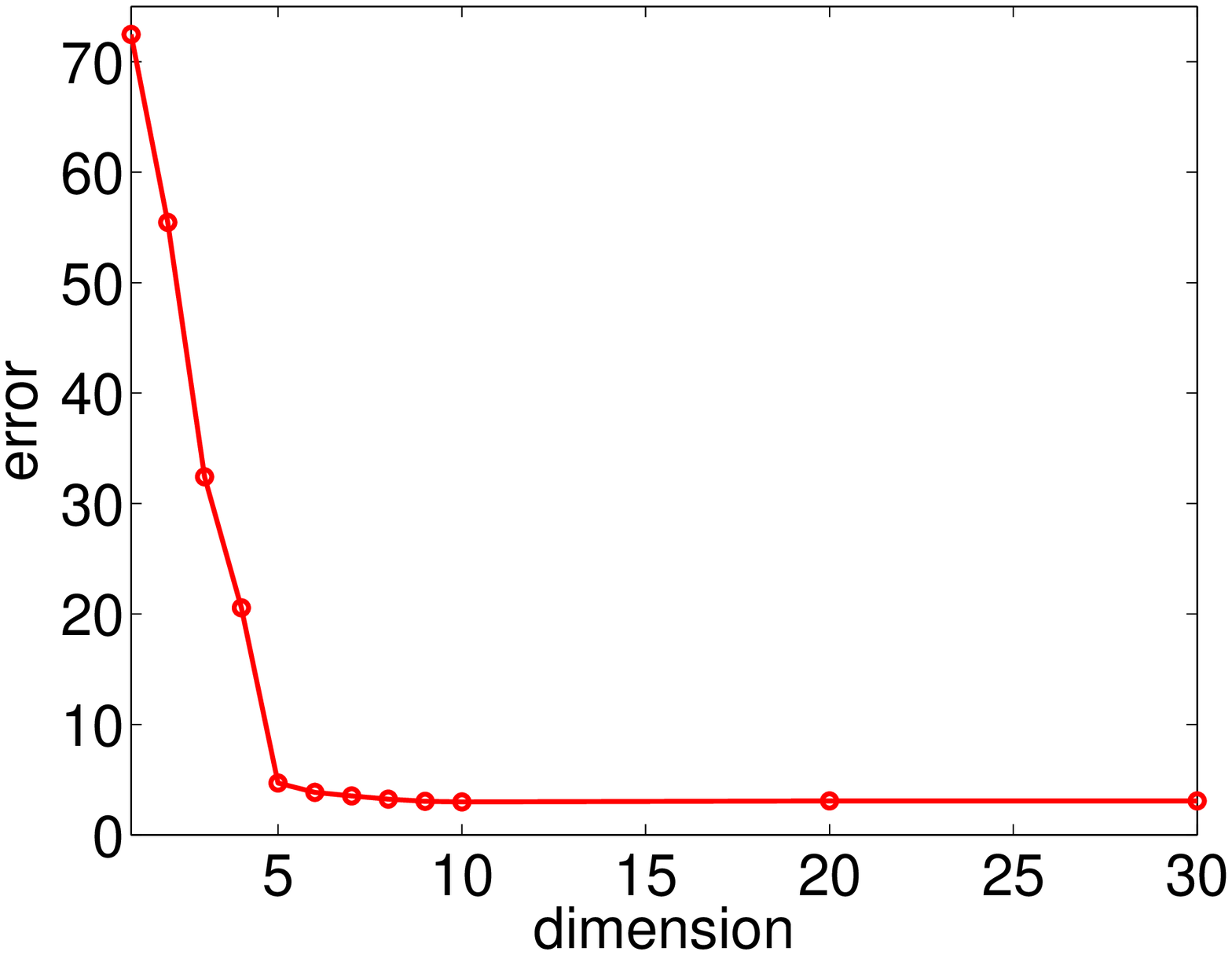}
    \end{tabular} &
    \begin{tabular}[b]{@{}c@{}}
      \psfrag{its}[t][]{iteration}
      \psfrag{obj}[][t]{objective function value}
      \includegraphics[width=0.31\linewidth]{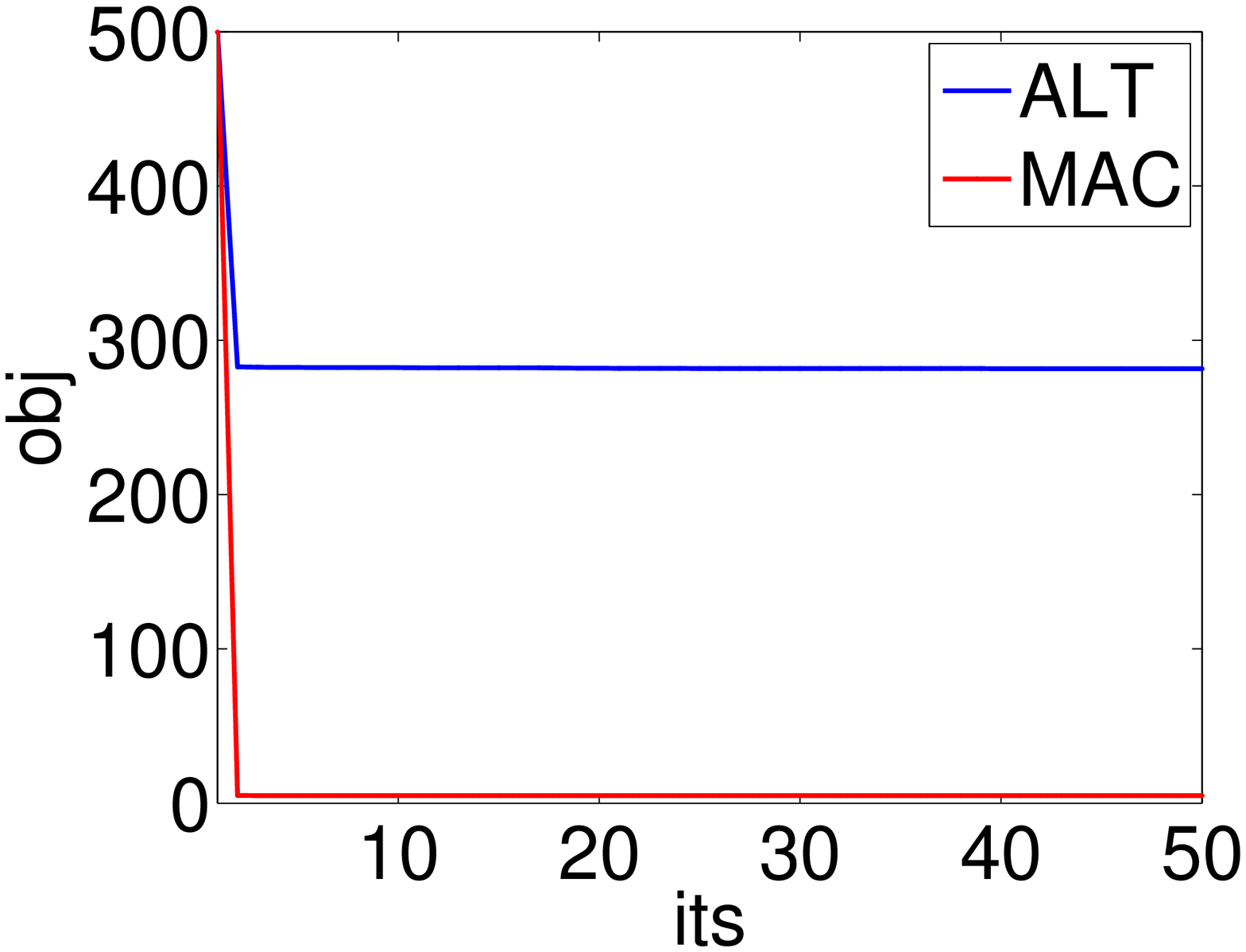}
    \end{tabular} &
    \begin{tabular}[b]{@{}c@{}}
      \psfrag{machines}[t][]{\# processors}
      \psfrag{speedup}[][t]{speedup}
      \includegraphics[width=0.31\linewidth]{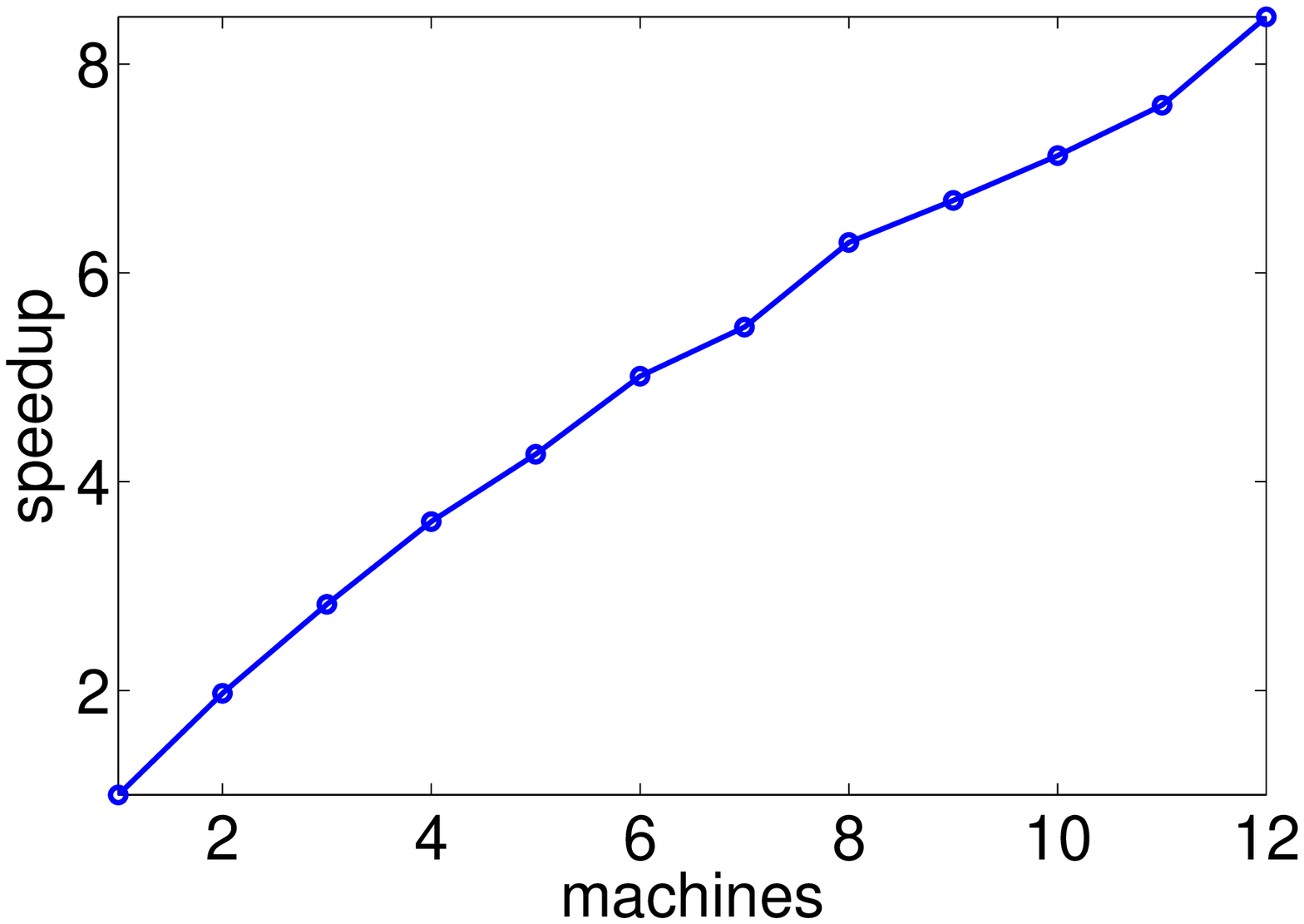}
    \end{tabular}
  \end{tabular}
  \caption{Results of MNIST 10-classes experiment. \emph{Top left}: test error rates (\%) and number of basis functions used in each method. In parenthesis, we specify $L$ for LDA/PCA, and $L$ and number of iterations used to reach early stopping for our algorithm. \emph{Top right}: latent representations obtained at dimension $L=10$ by our algorithm, visualized in 2D with PCA (to avoid cluttering, we plot a subset of $500$ images). Although several class digits appear to overlap, this is a visual artifact of the projection to 2D. \emph{Bottom left}: error rate of our algorithm over a range of latent dimensions. \emph{Bottom middle}: comparison of our algorithm with alternating minimization of the nested error. We plot the objective function value vs the iteration number. \emph{Bottom right}: speedups obtained by running our algorithm using the Matlab Parallel Processing Toolbox.}
  \label{f:MNIST_results}
\end{figure*}

Fig.~\ref{f:MNIST_results}(top left) shows the test error rates and the total number of support vectors (from the $10$ Gaussian SVMs)/basis functions used in each algorithm. Our accuracy is similar to that of the kernel SVM but with $5.5$ times fewer basis functions, obtaining a large speedup in testing time. We have again explored the approach of PCA/LDA+Gaussian SVM as in the previous experiment and obtained a similar result: LDA ($L=9$)+Gaussian SVM performs poorly; PCA+Gaussian SVM performs well at a relatively larger $L$. We train our model on the PCA projection, further reduce the dimension to $L=10$, and obtain similar performance with much fewer basis.

Fig.~\ref{f:MNIST_results}(bottom left) shows the performance of our algorithm using different values of the latent dimension $L$. Our error rates decrease quickly as $L$ increases at first. After $L=5$, it already does pretty well, and the performance does not change much for $L\ge 10$. We conjectured previously that the optimal latent configuration would be to arrange different classes on the vertices of a regular simplex. The projections achieved by our algorithm in fig.~\ref{f:MNIST_results}(bottom right) agree with that. Since we use PCA to visualize in 2D the latent representations lying in $L=10$ dimensions, some classes appear to overlap, but they are all completely separated, as can be seen by using other 2D views.

\subsection{Training runtime: comparison with alternating optimization without \Z}

We compared our three-step alternating optimization scheme with a two-step alternating optimization scheme over only \F\ and \g, which optimizes the original problem~\eqref{e:svm-nested} directly, where \F\ is a RBF network and \g\ a linear SVM. We are much faster in terms of both progress in objective function and actual runtime. The following experiment shows this in the MNIST odd/even classification problem for the case of $500$ training samples. To minimize the nested objective function in eq.~\eqref{e:svm-nested}, the two-step algorithm alternates a \g-step which solves a linear SVM on $\F(\X)$ and an \F-step that optimizes over the weights of \F\ by solving a quadratic program. Since the latent dimensionality is set to $L=2$, there are $1\,000$ weight parameters in the quadratic program, which we solve by calling the interior point solver of the CVX package \citep{GrantBoyd12a}. Fig.~\ref{f:MNIST_results}(middle) shows the nested objective function value versus iteration number during training. The time spent per iteration for our algorithm is about $1/150$ of that of alternating minimization. Therefore, the combined runtime of our \Z\ and \F\ steps is significantly lower than that of the \F\ step of alternating minimization, and besides our \Z\ step has a closed-form solution. Further, a few iterations of our algorithm suffice to find a near-optimal solution (with a small bias that continues to be eliminated as the penalty parameter $\mu$ is increased), while the progress of alternating minimization is hopelessly slow.

If instead of a RBF network, which is linear in the parameters of \F, we used a model which is nonlinear in the parameters (such as neural net), the gains of our algorithm would be even larger. The reason is that, in our algorithm, the nonlinearity over \F\ is confined to the \F-step in the form of a standard least-squares regression problem (section~\ref{s:F-step}), thanks to the use of the auxiliary coordinates \Z. Regression problems are well-studied and can be solved with efficient algorithms for many classes of functions \F. In the direct optimization over \F, this nonlinearity is embedded in the constraints of~\eqref{e:svm-nested}, which is harder to deal with.

\subsection{Parallel processing}

We also ran a simple parallel version of our algorithm in the MNIST 10-classes problem. This solves in parallel for the $10$ SVMs in the \g-step, and for the coordinates of all training points in the \Z-step (within each respective step, all these problems are independent). Given that our code is in Matlab, we used the Matlab Parallel Processing Toolbox. The programming effort is insignificant: all we do is replace the ``for'' loop over SVMs (in the \g-step) or over data points (in the \Z-step) with a ``parfor'' loop. Matlab then sends each iteration of the loop to a different processor. We ran this in a shared-memory multiprocessor machine%
\footnote{An Aberdeen Stirling 148 computer having 4 physical CPUs (Intel Xeon CPU L7555@ 1.87GHz), each with 8 individual processing cores (thus a total of 32 actual processors), and a total RAM size of 64 GB.},
using up to 12 processors (a limit imposed by our Matlab license). We obtain an impressive speedup of up to $8$ times, as shown in fig.~\ref{f:MNIST_results}(bottom right). Even larger speedups may be possible if using other parallel computation models, since the Matlab Parallel Processing Toolbox is quite inefficient.

\section{Discussion}

\paragraph{Filters}

Our algorithm illuminates the behavior of filter approaches. Such approaches optimize a proxy objective function constructed from $(\x_n,y_n)$ over \F\ and then learn the classifier \g\ by optimizing the classification error constructed from $(\F(\x_n),y_n)$ over \g. This is like our \F- and \g-steps, but, firstly, it uses the ``wrong'' objective for \F, and, secondly, without the coordination through the \Z\ variables, the process cannot be iterated to converge to a minimum of the joint problem. We can then view our algorithm as a \emph{corrected, iterated filter} approach. Since in practice it converges in a few iterations, its cost is just a little larger, but one need not introduce a proxy objective and yet obtains a true (local) minimum. Thus, we learn the following two lessons: (1) if we want to use a filter \F, the ideal filter would consist of mapping the inputs to class centroids located on the corners of a simplex in latent space; (2) we do not really need a filter approach, because we can train the optimal, wrapper approach nearly as efficiently and simply.

\paragraph{The role of dimensionality reduction in linear classification}

Being able to find true optima of the classification error allowed us to study the role of nonlinear DR as a preprocessing step for linear classification. With an ideally flexible DR mapping \F, the best possible preprocessing is precisely to remove all variation that is unrelated to the class label, including variation within a manifold---an extreme form of denoising. The input domains are ``denoised'' so they collapse to nearly zero-dimensional regions. In practice, \F\ belongs to a certain function class (given by the choice of model and number of parameters), and the ideal where classes collapse is only approached, but is clearly there. Using a latent space of $L=2$ dimensions is theoretically sufficient but, with a limited \F, using up to $L=K-1$ helps to improve the separation. Note that collapsing classes requires a genuinely nonlinear DR. The problem formulation of eq.~\eqref{e:svm-nested} does not explicitly seek to collapse classes, but this behavior emerges anyway from the assumption of low-dimensional representation, if trained jointly with the classifier. Thus, rather than making the classifier work hard to approximate a possibly complex decision boundary, we help it by moving the data around in latent space so the boundary is simpler.

This clashes with the widely held view that a good supervised DR method should produce representations (often visualized in 2D) where the manifold structure of each class is displayed. In fact, with an optimal DR the entire manifold will collapse. This is different from unsupervised DR, where we do want to extract informative features that tell us something about the data variability; and from supervised regression, where only some of the input dimensions should be collapsed (those which do not alter the output).

\paragraph{SVMs and kernel learning}

Our method and kernel SVMs can be seen as constructing a classifier as an expansion in basis functions $y = \g(\F(\x)) = \vv^T \bPhi(\x) + b$, with $M$ BFs in our case and $S$ support vectors for the SVM. The SVMs do this nonparametrically, at the cost of constructing an $N \times N$ kernel matrix and solving the corresponding problem, which is expensive---although much research work has sought to approximate this \citep{SchoelSmola01a} and reduce the number of SVs \citep{Bi_03a}. The basis functions $\bPhi(\x)$ are given by the kernel, which is selected by the user, and the space they implicitly map to is typically infinite-dimensional. The number of SVs $S$ is not set by the user but comes out automatically and can be quite large in practice. Our method is a parametric approach, where the user can set the number of BFs $M$, and the mapping \F\ (in this paper, an RBF mapping) maps to a low-dimensional space. The result is a competitive nonlinear classifier, with scalable training and efficient at test time. Having $M$ as a user parameter also allows a simple, direct way for the user to trade off test runtime for accuracy, which is crucial in real-time problems, such as embedded systems, where a typical SVM classifier is too computationally demanding in both memory required to store the SVs and in runtime. As shown in our experiments, we can obtain a classification error comparable to the kernel SVM with $M \ll S$, thus much faster.

It is possible to train a linear SVM by learning \vv\ and $b$ directly given fixed basis functions $\bphi(\x)$, but this achieves a worse classification error, and does not do DR, as we seek. Our low-dimensional classifier can be seen as a special regularization structure on the classifier's weights $\vv = \W^T \w$, where \W\ and \w\ are regularized separately. This effect is more pronounced in the multiclass case since each one-vs-all SVM $\w_k$ interacts with the same DR mapping \W. If using a different functional form for \F\ (e.g.\ deep nets), this resemblance with kernel SVMs disappears.

Our model can also be seen as learning a ``low-dimensional kernel'', since we pass a pair of latent vectors $(\F(\x),\F(\x'))$ to the linear SVM kernel, rather than applying a kernel $K(\x,\x')$ directly to the high-dimensional inputs. If $\F(\x) = \A\x$ is a linear DR mapping (no need for bias in latent space) then this becomes a form of metric learning with SVMs, using a low-rank metric $\A^T\A$.

\section{Related work}
\label{s:related}

Filter approaches typically learn a DR mapping \F\ using the inputs and label information first, and then fit a classifier \g\ to the latent projections and labels $(\F(\x_n),y_n)$. They are quite popular due to the ease of optimization, but they rely on the choice of objective function for the filter. Linear discriminant analysis (LDA; \citealp{Belhum_97a}) and its kernel version KLDA \citep{Mika_99b} look for a transformation of the inputs such that, in the latent space, the within-class scatter is minimized while the between-class scatter is maximized. The solution for \F\ can be obtained by solving an eigenvalue problem. These two algorithms can only produce up to $L = K-1$ latent dimensions for a $K$-class problem, due to the singularity of the between-class scatter matrix. Among other variations, \citet{Sugiyam07a} modify LDA to work for manifold data, but the projection mapping is linear, and \citet{UrtasunDarrel07a} derive a prior distribution from LDA in latent space and use it for a GPLVM to achieve DR, and the latent representation is then fed to a Gaussian process classifier. A clear disadvantage of filter approaches is the heuristic nature of the objective for DR, which acts as a proxy for classification, and is therefore not optimal for the classifier learned afterwards. As we showed, a good filter objective would be to collapse all classes and place them in the corners of a simplex.

Metric learning algorithms \citep{Xing_02a,Goldber_05a,GloberRoweis06a,WeinberSaul09a,Davis_07a} are closely related to DR for classification. Their goal is to find a Mahalanobis metric (or equivalently a linear transform) in input space such that samples in the same class are projected nearby in latent space while samples from different classes are pushed far apart. One achieves DR if the metric is low-rank. However, most metric learning algorithms first solve a positive semidefinite program without rank constraints, and then do a low-rank approximation of the learned Mahalanobis matrix to enforce dimensional reduction, thus the optimality of the projection is no longer guaranteed.

There also exist several wrapper approaches that train the DR mapping \F\ jointly with a classifier \g\ in a unified objective function. \citet{PereirGordon06a} (further generalized by \citealp{Rish_08a}) use an objective function that combines the approximation error of the inputs using SVD and the hinge loss from applying a linear SVM to the coordinates, so that the representation extracted this way will be good for classification. This is closely related to supervised dictionary learning \citep{Yang_12b}, only that the bias (approximation error) always exists in the model. Also, in this model the latent projections are an implicit function of the inputs, i.e., to project a new input, one needs to solve a optimization problem using learned basis. In contrast, our \F\ is an explicit mapping, directly applicable to test samples. \citet{JiYe09a} directly minimize the hinge loss of a SVM that operates on linearly transformed inputs (therefore it is a nested error, similar to us). They apply alternating optimization over both mappings. Due to the linearity of \F\ and \g, they are able to solve for \g\ in the dual and solve for \F\ using SVD. This would not be possible in general if \F\ has a different nonlinear form, unlike in our algorithm. Also, the SVD solution of \F\ limits the maximum meaningful latent dimension $L$ to the number of classes. In contrast with these approaches, our algorithm is bias free, works with any $L$, and trains a nonlinear DR mapping fast.

Auxiliary variables were used previously for unsupervised dimensionality reduction \citep{CarreirLu08a,CarreirLu10a} and regression \citep{WangCarreir12a}. But the unconstrained objective function defined there, while jointly optimized over a dimension reduction mapping \F\ and a regressor \g, differs from a true wrapper objective function, and results in optima for the combined mapping $\g \circ \F$ that are biased.

\section{Conclusion}

We have proposed an efficient algorithm to train a nonlinear low-dimensional classifier jointly over the nonlinear DR mapping and the classifier (a wrapper approach). The algorithm is easy to implement, reuses existing regression and SVM procedures, and parallelizes well. The resulting classifier achieves state-of-the-art classification error with a small number of basis functions, which can be tuned by the user. The algorithm can be seen as an iterated filter approach with provable convergence to a local minimum of the joint objective. This justifies filter approaches that use a secondary criterion over the DR mapping such as class separability or intra-class scatter in an effort to construct a good classifier, but also obviates them, since one can ensure to get the best low-dimensional classifier (under the model assumptions) with just a little more computation.

Our experiments illuminate the role of nonlinear DR in linear classification. If we optimize the classification error---the figure of merit one really cares about---\emph{jointly} over the projection mapping and the classifier, the best DR in fact erases all structure (manifold and otherwise) in the input other than class membership, and uses the latent space to place collapsed classes in such a way that they are maximally linearly separable. Future work should analyze the role of DR with nonlinear classifiers.

Our algorithm generalizes beyond the specific forms of DR mapping and classifier used here, and we are exploring other combinations. In particular, one can replace the DR mapping with a complex feature-extraction mapping that can handle invariances, such as convolutional neural nets, and jointly optimize this and the classifier.

\subsubsection*{Acknowledgments}

Work funded in part by NSF CAREER award IIS--0754089.

% \bibliographystyle{abbrvnat}
% \bibliography{macp,macp-xref}

\end{document}

%% file: MACPdef.tex
\newcommand{\A}{\ensuremath{\mathbf{A}}}

\newcommand{\F}{\ensuremath{\mathbf{F}}}

\newcommand{\W}{\ensuremath{\mathbf{W}}}
\newcommand{\X}{\ensuremath{\mathbf{X}}}
\newcommand{\Y}{\ensuremath{\mathbf{Y}}}
\newcommand{\Z}{\ensuremath{\mathbf{Z}}}

\renewcommand{\c}{\ensuremath{\mathbf{c}}}

\newcommand{\g}{\ensuremath{\mathbf{g}}}

\newcommand{\vv}{\ensuremath{\mathbf{v}}}
\newcommand{\w}{\ensuremath{\mathbf{w}}}
\newcommand{\x}{\ensuremath{\mathbf{x}}}

\newcommand{\z}{\ensuremath{\mathbf{z}}}
\newcommand{\0}{\ensuremath{\mathbf{0}}}

\newcommand{\bmu}{\ensuremath{\boldsymbol{\mu}}}

\newcommand{\bphi}{\ensuremath{\boldsymbol{\phi}}}

\newcommand{\bxi}{\ensuremath{\boldsymbol{\xi}}}

\newcommand{\bPhi}{\ensuremath{\boldsymbol{\Phi}}}

\newcommand{\bbR}{\ensuremath{\mathbb{R}}}

\newcommand{\calL}{\ensuremath{\mathcal{L}}}

\newcommand{\calO}{\ensuremath{\mathcal{O}}}

\newcommand{\norm}[1]{\left\lVert#1\right\rVert}

{%
\begin{list}{#1}{
\vspace{-\topsep}
\vspace{-\partopsep}
\setlength{\itemindent}{0cm}
\setlength{\rightmargin}{0cm}
\setlength{\listparindent}{0cm}
\settowidth{\labelwidth}{#1}
\setlength{\leftmargin}{\labelwidth}
\addtolength{\leftmargin}{\labelsep}
\setlength{\itemsep}{0cm}
}%
}%
{%
\end{list}
\vspace{-\topsep}
\vspace{-\partopsep}
}

{\begin{enumerate}%
}%
{\end{enumerate}}

%% file: MACPdef-ams.tex
\theoremstyle{plain}% default

\newtheorem*{lemma*}{Lemma}

\newtheorem*{prop*}{Proposition}

\theoremstyle{definition}

\newtheorem*{defn*}{Definition}

\newtheorem*{exmp*}{Example}

\newtheorem*{conj*}{Conjecture}

\theoremstyle{remark}

\newtheorem*{rmk*}{Remark}